\documentclass[10pt,twocolumn,letterpaper]{article}

\usepackage{cvpr}
\usepackage{microtype}

\usepackage{nth} % For 1st,2nd,3rd...
\usepackage{enumitem} % For custom lists for questions and parameter definitions
\usepackage{float} % for figure placement
\usepackage{siunitx}          % S column type for numeric IDs
\usepackage{pifont}        % provides \ding{…}
\DeclareMathOperator*{\argmin}{arg\,min}

\usepackage[accsupp]{axessibility} 

\newcommand{\cmark}{\ding{51}}   % ✓
\newcommand{\xmark}{\ding{55}}   % ✗

\usepackage{chapterbib}

\definecolor{cvprblue}{rgb}{0.21,0.49,0.74}
\usepackage[pagebackref,breaklinks,colorlinks,allcolors=cvprblue]{hyperref}

\def\paperID{33954}

\def\confName{CVPR}
\def\confYear{2026}

\title{U-SEG: Uncertainty in SEGmentation - A systematic multi-variable exploration}

\author{Michael Smith \quad Frank P. Ferrie\\
Centre for Intelligent Machines, McGill University\\
{\tt\small michael.smith6@mail.mcgill.ca \quad  frank.ferrie@mcgill.ca}
}

\begin{document}
\maketitle
\begin{abstract}
In this study, we explore in depth a few under-studied topics at the intersection of uncertainty estimation and segmentation. Prior work has shown that the quality of uncertainty estimates can be very sensitive to a range of variables. As one of the main uses of uncertainty estimation is to help identify and deal with prediction errors in practical scenarios, any factors that affect this must be clearly identified. For example, do more challenging domains or different datasets and architectures result in worse performance when using uncertainty estimates? Can prior frames in a video sequence in fact provide useful uncertainty estimates comparable to other approaches? Is it possible to combine uncertainty estimation approaches, taking advantage of sample diversity, to get better estimates? Finally, when might it make sense to use an ensemble-based uncertainty estimate over a deterministic network? We address these questions by creating a framework for and executing a large scale study across many variables such as datasets, backbones, and downstream tasks, for both semantic and panoptic segmentation. We find that a) the more challenging task of panoptic segmentation usually results in worse performance while high performance variance between datasets and backbones indicates that generalization is not guaranteed, b) time series samples can be useful for specific configurations, but in many cases are not worth the cost, c) sample diversity shows the most promise in the downstream task of calibration, but otherwise fails to beat simpler alternatives, d) a deterministic approach is adequate for some downstream tasks, but ensembles allow for significant improvements if the right conditions can be achieved in deployment.
\end{abstract}    
\section{Introduction}
\label{sec:intro}

\begin{figure*}[ht]
    \centering
    \includegraphics[width=1\linewidth]{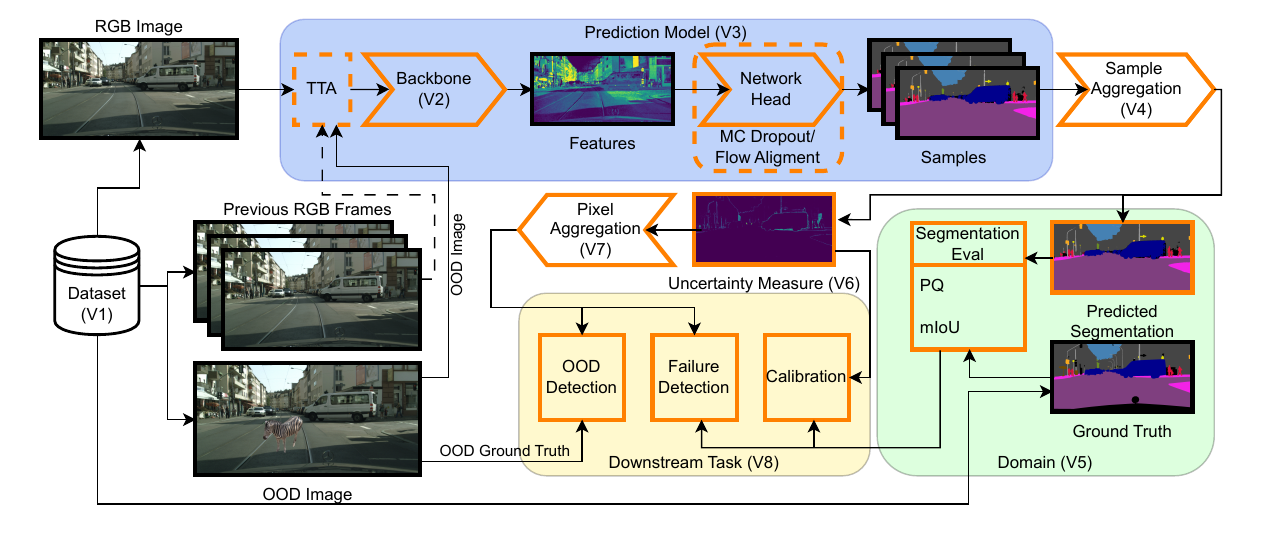}
    \caption{A simplified overview of our framework (\cref{subsec:general_implementation}) and the key variables (\cref{subsec:key_variables_list}) we discuss. Orange outlines depict operations. Black outlines depict inputs and outputs. Dashed lines indicate components or processes that are only used with certain prediction models. For brevity, some steps are shown with pertinent outputs overlaid \eg the uncertainty measure.}
    \label{fig:study_overview}
\end{figure*}

For any neural network model deployed outside the lab, reliability and robustness are critical. Many algorithms continue to be proposed in an effort to handle real-world deployment challenges such as unknown classes or distribution shifts \cite{schmidt_prior2former_2025,benigmim_floss_2025,bui_density-softmax_2024}. However, deficiencies in evaluation continue to be an issue \cite{maier-hein_metrics_2024,valiuddin_review_2024}. With respect to uncertainty estimation, it has been shown that a number of variables can significantly affect it in both image classification \cite{jaeger_call_2023} and semantic segmentation \cite{kahl_values_2024}. While such work has made great strides in assisting practitioners to determine the best uncertainty estimation approach for their problem, it unfortunately fails to address a number of important questions.

We address this deficiency by creating a framework that enables us to study multiple little explored aspects of uncertainty estimation on the task of segmentation. Based on Mask2Former \cite{cheng_masked-attention_2022}, our framework is designed such that we can make changes to one variable (such as the domain) without needing to change others, unlike some prior work \cite{kahl_values_2024}. These allow us to precisely quantify the effect on uncertainty estimates by changes in variables such as datasets, domains or backbones. We also design our framework such that we can explore time series data and combinations of prediction models more thoroughly than previously studied. As diversity in various forms has proven its worth in many other domains, we wish to determine to what extent it may prove useful towards generating uncertainty estimates. Lastly, we put particular emphasis on evaluating deterministic (sampling-free) baselines for uncertainty estimation under the same conditions as more complex approaches. This allows us to determine if and when a baseline may be acceptable to use. Evaluations conducted on a range of downstream tasks allow us to comprehensively answer our questions.

Our experiments span 2 datasets, 2 distribution shifts, 3 backbones, 39 prediction model configurations (including combinations and a baseline), 2 domains (semantic and panoptic segmentation), 3 sample aggregation algorithms, 11 uncertainty measures, 2 pixel aggregation methods, and 3 downstream tasks. We find that even beyond expected changes to segmentation performance (mIoU and PQ), as measured by the AURC, ECE and AUROC metrics, different domains and datasets can significantly affect uncertainty estimates to the point that a baseline may be better in some circumstances than others. Compared to a baseline, we can see performance drops below $100\%$ and gains around $75\%$ across different downstream tasks, with the exact numbers highly dependent on the variables. Furthermore, under certain circumstances, it can be worth it to explore either combinations of prediction models or time series data as a source of uncertainty, but they are not universally useful.

\textbf{Contributions}: Our main contribution is a comprehensive large scale evaluation and discussion of different uncertainty estimation approaches towards segmentation across numerous variables applied in downstream tasks, with a particular focus on several under-explored areas in the literature:
    \begin{itemize}
        \item Time-series data as an ensemble
        \item Combinations of prediction models \eg MC Dropout, Test-time augmentation
        \item Changes in domain, dataset, and backbone
        \item Reasonable baseline comparisons
    \end{itemize}
To this end, we make several secondary contributions:
\begin{itemize}
    \item A framework that allows for independent variable evaluation in the context of sample-based uncertainty estimation in segmentation
    \item Introducing \textit{Mask Distance}, a novel memory-efficient approach for sample aggregation with universal-type networks such as Mask2Former
\end{itemize}
Code and data are available at \url{https://github.com/mdsmith-cim/U-SEG-CVPRF-2026}.
\section{Related Work}
\textbf{Uncertainty}: Given the real world deployment applications, uncertainty estimation continues to be an active area of research \cite{wilson_evaluating_2022,papamarkou_position_2024,smith_rethinking_2025,valiuddin_review_2024} applied across a diverse set of tasks \cite{nie_epistemic_2025,schrufer_are_2024,urbinati_are_2025,liu_uncertainty_2025}. Approaches to estimating uncertainty can roughly be categorized into those that sample from an ensemble \cite{gal_dropout_2016,lakshminarayanan_simple_2017,huang_efficient_2018,laurent_packed_2023,rame_model_2023} and those that do not (typically by fitting a distribution) \cite{blundell_weight_2015,sensoy_evidential_2018,charpentier_natural_2022,bui_density-softmax_2024,osband_epistemic_2023,li_vicinal_2025}. A parallel track focuses on calibration \cite{guo_calibration_2017,cheng_towards_2024,liu_improving_2025,arad_improving_2025}, which is part of uncertainty estimation but often treated as a distinct sub-field. This silo-type approach is widespread, with evaluations thorough in some ways (\eg datasets) but not others (\eg downstream tasks). When different domains are evaluated (\eg image classification and semantic segmentation), the approach may be the same but changes to the experimental procedure (\eg model architecture, weights) prevent the effect of the domain change from being quantified. Several excellent studies have begun to address the problem of focused evaluation as it pertains to uncertainty estimation \cite{jaeger_call_2023,kahl_values_2024,roschewitz_where_2025,minderer_revisiting_2021}, with results complementary to ours. \cite{minderer_revisiting_2021,roschewitz_where_2025} only discuss calibration, \cite{jaeger_call_2023} addresses image classification, while \cite{kahl_values_2024} is perhaps the closest to our work but they address only semantic segmentation. Compared to our work, they do not disentangle dataset, domain or backbone changes, nor do they evaluate time series samples or combinations of approaches.

\textbf{Diversity}: With ensembles for uncertainty estimation, diversity is broadly considered crucial \cite{bursuc_many_2023,zhou_ensemble_2025}. Having samples also makes it possible to combine approaches (unlike alternatives) which begs the question if there is any benefit to doing so. The literature is quite sparse, and somewhat contradictory. \cite{asgharnezhad_uncertainty-aware_2025} combine Deep Ensembles and MC Dropout for image level skin cancer classification, but fail to gain any benefit from it. Meanwhile, \cite{wen_combining_2021} find that using data augmentation and ensembles at the same time harms calibration performance while \cite{roschewitz_where_2025} find that Deep Ensembles helps calibration under many conditions. We thus have claims that leveraging diversity can be useless, harmful, or helpful, making this an open problem.

\textbf{Segmentation}: Segmentation continues to be a well-studied area, with attempts at addressing real-world deployment challenges in semantic segmentation \cite{thisanke_semantic_2023,pan_exploring_2025,wei_stronger_2024,chen_stronger_2025,ge_clip-adapted_2025} and (2D) panoptic segmentation \cite{kirillov_panoptic_2019,chen_generalist_2023,guo_visual_2023,saha_edaps_2023,schmidt_prior2former_2025}. Most however are out-of-scope for our work, with a focus on other areas such as domain adaptation. With respect to network architectures, Mask2Former \cite{cheng_masked-attention_2022} is extremely popular and we use it as part of our framework. Its design as a \textit{universal architecture} allows us to fairly evaluate both semantic and panoptic segmentation, unlike alternatives \eg \cite{cheng_panoptic-deeplab_2020}. Recent work complementary to ours has also evaluated uncertainty estimates as applied towards semantic segmentation \cite{mukhoti_deep_2023,landgraf_uncertainty-aware_2024,thoma_uncertainty_2025} and panoptic segmentation \cite{sirohi_uncertainty-aware_2023,deery_propandl_2023,smith_uncertainty_2024}, with narrow coverage on most potential variables and generally ignoring the questions we prioritize. We also question the utility of various permutations of ECE (Expected Calibration Error) introduced in the literature as it relates to panoptic segmentation. Specifically, in our view \cite{sirohi_uncertainty-aware_2023} needlessly biases calibration error calculations via a reliance on segment matching while \cite{deery_propandl_2023} add further complication by decomposing the calibration error into the constituent parts of panoptic segmentation.

\textbf{Baselines}: A repeated trend in the literature is to compare only against other approaches in the same sub-field without a thorough consideration of what a baseline should be \cite{kahl_values_2024}. As an example, \cite{charpentier_natural_2022,li_vicinal_2025} provide a thorough comparison against other approaches \eg \cite{gal_dropout_2016,sensoy_evidential_2018}. However, minimal consideration is given to evaluating potentially simple to implement alternatives. As previously noted in the literature, a good baseline can sometimes match or exceed the performance of a more complex approach \cite{henderson_deep_2018,gallos_active_2019,jaeger_call_2023,kahl_values_2024}. As such, in our work we evaluate ensembles and baselines under the same conditions.

\textbf{Temporal data}: \cite{huang_efficient_2018} introduced the idea of using time series data in an ensemble to estimate uncertainty in semantic segmentation, albeit with limited and contradictory evaluation results. While taking advantage of temporal data remains a popular topic \eg \cite{xu_dual-temporal_2025}, the literature since \cite{huang_efficient_2018} is limited, with only \cite{maag_improving_2021} who focus on instance segmentation and devise a complex tracking algorithm to match instances between frames followed by \cite{maag_false_2021} who evaluate on a variant of failure detection.

\textbf{Matching}: The problem we address via our \textit{Mask Distance} approach is not new. \cite{smith_uncertainty_2024} address it in the panoptic domain with a variant of the BSAS (Basic Sequential Algorithmic Scheme); they also compare against the Hungarian Method. The problem is also of interest in the instance segmentation literature, where BSAS \cite{miller_evaluating_2019}, GMM \cite{heidecker_towards_2021} and BGM \cite{heidecker_sampling-based_2023} have been used. All are classical approaches. Of note is \cite{morrison_estimating_2019} who uses masks rather than bounding boxes during matching as we do, albeit via IoU matching and BSAS rather than Euclidean distance.
\section{Uncertainty Estimation in Segmentation}
\label{sec:uncert_est}
We focus on two well defined domains: semantic and panoptic segmentation. In semantic segmentation, the goal is to assign a class label to each pixel in an image. In panoptic segmentation \cite{kirillov_panoptic_2019}, we additionally assign an instance label to some pixels where relevant, allowing us to distinguish instances \eg two cars side by side.

\subsection{Ensembles}
\label{subsec:ensembles}
Considering uncertainty estimation, we take a Bayesian viewpoint of ensembles following \cite{malinin_uncertainty_2019}. Unlike other literature \eg \cite{kahl_values_2024}, we do not use the term ensembles to refer solely to Deep Ensembles \cite{lakshminarayanan_simple_2017}. We assume we have an ensemble of $Q$ independent models $\left\{P(y \mid \boldsymbol{x}^*, \mathcal{Q}^{(q)})\right\}^Q_{q=1}$, with the approach used (\eg MC Dropout \cite{gal_dropout_2016}) being the differentiating factor between them. Being independent also makes it possible to combine approaches. We can get the expected prediction via the mean (if we know the correspondence, see \ref{param:sample_agg} and \cref{subsec:mask_distance}):
\begin{equation}
    \label{eq:predictive_mean}
    P(y \mid \boldsymbol{x}^*, \mathcal{D}) = \frac{1}{Q} \sum^Q_{q=1} P(y \mid \boldsymbol{x}^*, \mathcal{Q}^{(q)}),
\end{equation}
where $\mathcal{D}$ is a dataset and $\boldsymbol{x}^*$ is an input image. We can also calculate measures of uncertainty such as predictive entropy \cite{shannon_mathematical_1948}:
\begin{equation}
\label{eq:pred_entropy}
\mathcal{H}\left[P(y \mid \boldsymbol{x}^*, \mathcal{D})\right] \approx \mathcal{H}\left[\frac{1}{Q} \sum^Q_{q=1} P(y \mid \boldsymbol{x}^*, \mathcal{Q}^{(q)})\right].
\end{equation}

\subsection{Key Variables}
\label{subsec:key_variables_list}
Applying ensembles as presented in the general case above to segmentation is not necessarily straightforward, however, with many possible variables. Excluding those tied to human annotation (\eg inter-rater reliability in medical imaging), we define eight of them, shown in \cref{fig:study_overview} and defined below:
\begin{enumerate}[label=\textbf{V\arabic*}]
    \item \textbf{Dataset}: Different datasets can have very different distribution characteristics, and generalization from one to another is usually a key concern. We evaluate on two different datasets and explore the effect of distribution shifts by cross-evaluating models trained on each.\label{param:dataset}
    \item \textbf{Backbone}: Changes in architecture (\eg $\text{CNN} \rightarrow \text{Transformer}$) are known to impact many performance metrics, all else being equal. Large foundation models have also been shown to lead in some benchmarks such as BRAVO \cite{vu_bravo_2025}.\label{param:backbone}
    \item \textbf{Prediction Model}: Numerous approaches (\eg MC Dropout \cite{gal_dropout_2016}, Deep Ensembles \cite{lakshminarayanan_simple_2017}) exist to estimate uncertainty. We treat all evaluated approaches as ensembles as defined in \cref{subsec:ensembles}. Importantly, we also evaluate different numbers of samples and combinations of the different approaches.\label{param:pred_model}
    \item \textbf{Sample Aggregation}: Given a universal network architecture such as Mask2Former \cite{cheng_masked-attention_2022} and the complex problem formulation of panoptic segmentation, correspondence between samples is needed to merge the predicted segmentations.\label{param:sample_agg}
    \item \textbf{Domain}: Multiple formulations of the segmentation task exist in the literature, all of which share the general objective of attempting to segment one or more objects in an image. Popular examples include semantic, instance or panoptic segmentation.\label{param:domain}
    \item \textbf{Uncertainty Measure}: Given an ensemble, multiple uncertainty measures can be calculated, such as predictive entropy or the softmax operator as a simple baseline. Different measures capture varying amounts of aleatoric and epistemic uncertainty, with downstream tasks requiring one, the other, or both.\label{param:uncert_measure}
    \item \textbf{Pixel Aggregation}: In our work, 2 of 3 downstream tasks require pixel-level uncertainty to be converted to image level, \eg by applying the sum of each pixel.\label{param:pixel_agg}
    \item \textbf{Downstream Task}: Uncertainty estimates alongside a segmentation task are only useful if they can be applied towards some downstream task, such as identifying prediction errors in failure detection.\label{param:downstream_task}
\end{enumerate}

\section{Approach}
\label{sec:study_approach}

\subsection{Key Questions}
With the high number of potential variables, a practitioner attempting to deploy a given approach may have to contend with a large number of unknowns that can significantly impact expected performance. Our study attempts to comprehensively address several open questions surrounding segmentation models and uncertainty estimation:

\begin{enumerate}[label=\textbf{Q\arabic*}]
    \item \textbf{Effect of domain (\labelcref{param:domain}), dataset (\labelcref{param:dataset}), backbone (\labelcref{param:backbone})}: Much work in segmentation focuses on a single domain such as semantic segmentation. However, trends in one domain do not necessarily generalize to another (\eg image classification to semantic segmentation) \cite{jaeger_call_2023,kahl_values_2024}. We also know that performance can vary significantly when testing on different datasets or with different network architectures. For instance, foundation models were shown to be much more effective than other architectures on the BRAVO challenge \cite{vu_bravo_2025}. How might changing the domain, dataset or backbone affect our ability to quantify uncertainty? Our framework allows us to evaluate each of these variables and their effect on uncertainty estimates while keeping all others effectively fixed.\label{q:domain_dataset_backbone}
    \item \textbf{Sample Diversity}: Increased sample diversity has been noted to be beneficial in many domains, including segmentation \cite{everingham_pascal_2015} and for ensembles specifically \cite{zhou_ensemble_2025}. To what extent can having more samples and several different sources of samples (\ref{param:pred_model}) help us with uncertainty estimation?\label{q:samplediversity}
    \item \textbf{Time series}: Using time series data (predictions from prior frames in a video sequence) as a prediction model (\labelcref{param:pred_model}) for semantic segmentation was introduced in \cite{huang_efficient_2018}. However, experimental validation is not comprehensive. We revisit this approach and attempt to determine if it can indeed generate viable uncertainty estimates under a wide range of conditions, including the panoptic segmentation domain.\label{q:time}
    \item \textbf{Baseline utility}: Different downstream tasks, variables or combinations thereof can significantly affect performance \cite{kahl_values_2024}. In image classification, it has been shown that a simple softmax-based approach is sufficient in many cases \cite{jaeger_call_2023}. At the same time, a typical evaluation procedure for a new uncertainty estimation approach will omit simple baselines \eg \cite{li_region-based_2023}. Given the high number of variables in play, to what extent is it worth it to use a complex (ensemble-based) uncertainty estimation method?\label{q:baseline_usefulness_uncert}
\end{enumerate}

\subsection{General Implementation}
\label{subsec:general_implementation}
In \cref{fig:study_overview}, we provide an overview of our framework, showing how different components are connected and tie into the variables discussed in \cref{subsec:key_variables_list}. In general, it is possible to change the value of one variable without requiring a change in another, unlike other approaches \eg \cite{kahl_values_2024}. This means we can directly compare uncertainty estimates between say panoptic and semantic segmentation.

We evaluate on two datasets: Cityscapes \cite{cordts_cityscapes_2016} (real-world images) and VIPER \cite{richter_playing_2017} (simulated images). As the datasets are both driving oriented (see \cref{subsec:app:datasets}), we also create variants of each where the labels are mapped as best as possible to those of the other dataset. This allows for distribution shift evaluation with a model trained on one dataset evaluated on the other. When evaluating a time-series prediction model, both datasets provide RGB images of prior video frames preceding each annotated frame we evaluate on. Once processed by the network, the resulting predictions are aligned using ground truth optical flow for VIPER and the RAFT optical flow network \cite{teed_raft_2020} for Cityscapes. For OOD (out-of-distribution) evaluation, we use OOD images provided by \cite{mohan_panoptic_2024} (Cityscapes only).

Moving on to the model, RGB images are fed into a neural network based on the Mask2Former model \cite{cheng_masked-attention_2022}. The network is split into a backbone and head. We evaluate three backbones: \textit{ResNet50} \cite{he_resnet_2016}, \textit{Swin-B} \cite{liu_swin_2021} and \textit{DINOv2} \cite{oquab_dinov2_2024}, with the latter being a foundation model that uses the \textit{ViT-B} architecture \cite{dosovitskiy_image_2021}. For each dataset, we use weights if possible from \cite{cheng_masked-attention_2022}; otherwise, models are trained with relevant hyperparameters mapped from \cite{cheng_masked-attention_2022} as much as possible. The architecture for the head is fixed but retraining is required for any changes to the backbone. Importantly, only changes in training dataset and backbone affect the network weights.

Three prediction models (the process used to generate samples) are evaluated, both individually and combined: MC Dropout \cite{gal_dropout_2016}, Time series \cite{huang_efficient_2018}, and TTA (Test-Time Augmentation) \cite{ayhan_test-time_2018}. Including the baseline and combinations, we evaluate 39 different configurations (\ie different numbers of samples). When combining samples, the net results are multiplicative \eg 3 MC Dropout samples and a horizontal flip will result in 6 total samples, half of which are flipped. Baselines are defined to be the deterministic Mask2Former network used without any sampling; \labelcref{param:sample_agg} thus does not apply. Note that our implementation of \cite{huang_efficient_2018} is not one-to-one and does not correct for cumulative noise with increasing samples; please see \cref{subsec:app:time_series_details}.

Assuming we have samples from one or more prediction models, we need a way to aggregate them into a coherent prediction. We are aware of three ways to do this: \textit{Averaging} (a BSAS variant) \cite{smith_uncertainty_2024}, \textit{Pixel Decoder Averaging}, and \textit{Mask Distance}, the latter two which we introduce in \cref{subsec:mask_distance,subsec:app:pixel_decoder} respectively. All experiments use the \textit{Mask Distance} approach given its performance as shown in \cref{subsec:app:sample_agg_results}.

The next step is the choice of domain: semantic or panoptic segmentation. Ground truth annotations from the dataset are used to calculate standard segmentation metrics as shown in \cref{subsec:metrics}.

The penultimate part of our framework is the calculation of uncertainty. First, we calculate 11 different uncertainty measures (see \cref{sec:app:uncert_measures} for details) from the samples. Note that our ability to disentangle the effect of individual variables is not perfect, as the choice of sample aggregation can restrict the choice of uncertainty measure. For failure and out-of-distribution detection, we calculate uncertainty (\eg mutual information) at the pixel level and convert it to an image-level measure by either Patch Level or Image level aggregation following \cite{kahl_values_2024}. For calibration, we apply the pixel-level uncertainty measure of a mask-weighted softmax vector directly.

The last part of our framework is the practical application of uncertainty on downstream tasks: OOD (out-of-distribution detection), failure detection, and calibration. For out-of-distribution detection, the goal is to identify images via an uncertainty measure where an object from a class outside the training set has been added. Given two identical images other than the added OOD object (\eg as shown in \cref{fig:study_overview}), at the image level, uncertainty should be greater in the OOD case. In failure detection, the goal is to identify images where the network has made mistakes in its predicted segmentation as defined by standard segmentation evaluation metrics. We expect image-level uncertainty to be higher the more prediction errors there are. For calibration, the goal is to have predicted confidence scores match their actual occurrence rate \eg all predictions with a confidence of $70\%$ should be correct $70\%$ of the time. We follow argmax calibration as used in \cite{guo_calibration_2017}; see \cite{minderer_revisiting_2021} for a thorough theoretical definition of calibration. Overall, we expect that better uncertainty estimates will result in higher uncertainty for OOD images and those with prediction errors, as well as reduced calibration error.
For further implementation details, please see \cref{sec:app:implementation_details}.

\subsection{Mask Distance Sample Aggregation}
\label{subsec:mask_distance}
A single sample $S_{i}$ from the network head of a universal architecture such as Mask2Former consists of a fixed number of predictions $p \in P$, each of which represents a potential object in the scene. Each prediction has a vector $L$ representing the class distribution and a mask $M$ representing the location of the object in the scene \cite{cheng_masked-attention_2022,smith_uncertainty_2024}:
\begin{equation}
    \label{eq:model_output_logits_mask}
    S_{i=1\dots n } = \{L_i \in \mathbb{R}^{P \times c}, M_i \in \mathbb{R}^{P \times h \times w}\},
\end{equation}
where $c$ is the number of classes in the dataset, $h \times w$ is the size of the mask, and $n$ the number of samples. Unfortunately, with multiple samples, we have no immediate way to determine the correspondence between each of the outputs \eg is prediction $\#4$ in the first sample the same as prediction $\#21$ in the second?

This leads to our introduction of \textit{Mask Distance}, where we find that, for each prediction, in each sample minimizing the Euclidean distance between the masks (with a sigmoid function) is optimal in terms of performance:
\begin{equation}
    \label{eq:euclidean_mask_distance}
    \argmin_{p \in P} \|\sigma(M_{i})-\sigma(M_{i-1})\|, \quad \forall i = 1 \dots n,
\end{equation}
allowing us to calculate the correspondence between each sample and apply the mean to retrieve $S=\frac{\sum^{n}_{i=1}S_i}{N}$. Note that our implementation is slightly different from the equations as presented due to optimizations; please see \cref{subsec:app:mask_distance_extended} for further details. 

\subsection{Metrics}
\label{subsec:metrics}
All metrics in the study are chosen or designed such that they maintain objective compatibility across domains. Starting with segmentation, we use the common PQ (Panoptic Quality) \cite{kirillov_panoptic_2019} and mIoU (mean Intersection-over-Union) \cite{shelhamer_fully_2017,lin_microsoft_2014} metrics for panoptic and semantic segmentation respectively. These measure the pixel-level correctness of predicted segments, with PQ additionally using a segment-level overlap threshold to determine if a given segment is correct. Note that mIoU, Dice and PQ can all be related to each other \cite{maier-hein_metrics_2022-arxiv}.

For the downstream task of failure detection, we rely on the AURC (Area under the Risk-Coverage-Curve) metric used in \cite{jaeger_call_2023} and applied to semantic segmentation in \cite{kahl_values_2024}. It encapsulates expected criteria for failure detection, taking into account both uncertainty estimates and the fact that an outright better classifier can do better by simply making fewer errors. We follow \cite{kahl_values_2024}, except we substitute the Dice score with mIoU and PQ for semantic and panoptic segmentation respectively. We define the confidence scoring function as either the negative uncertainty score or the confidence score where appropriate \eg for a simple uncertainty metric like softmax confidence.

For out-of-distribution detection, following standard practice, we use the AUROC (Area under the Receiver Operating Characteristic). Uncertainty estimates from the model for images with and without OOD objects are collected and the problem is treated as binary image classification.

For calibration, we forgo existing metrics such as pECE \cite{sirohi_uncertainty-aware_2023} or the image-level Platt scaled Average Calibration Error used by \cite{kahl_values_2024} in favour of more naturally extending the extremely widely used ECE \cite{pakdaman_naeini_obtaining_2015,guo_calibration_2017}, despite its flaws \cite{minderer_revisiting_2021}, to segmentation. We use a mask-weighted softmax as we expect the confidence score to be calibrated as it encapsulates all relevant information the model has for each pixel. This score is then compared against the correctness of the prediction at each pixel. For semantic segmentation, this resolves to the class label being correct; for panoptic segmentation, we use the criteria established by the PQ metric which requires the class label and instance label to both be correct and to have a minimum degree of overlap with a relevant ground truth segment.

For further details on metrics, see \cref{sec:app:metrics_extended}.
\section{Results}
\label{sec:results}
Using our framework, we evaluate many prediction model configurations (\labelcref{param:pred_model}) on various datasets (\labelcref{param:dataset}), backbones (\labelcref{param:backbone}), domains (\labelcref{param:domain}) and downstream tasks (\labelcref{param:downstream_task}), across 3 seeds. We evaluate different uncertainty measures (\labelcref{param:uncert_measure}) and pixel aggregation (\labelcref{param:pixel_agg}) methods as well but marginalize them out for clarity as we feel that the impact of those variables have been sufficiently explored in \cite{kahl_values_2024}. We individually plot downstream tasks, and outside of \cref{fig:fd_calib_by_dataset_backbone_raw} we present all results normalized to a baseline. This baseline is defined as the best possible performance on a given downstream task for a particular dataset and backbone combination without using a prediction model to generate samples - in other words, a deterministic off-the-shelf Mask2Former network with the same weights but significantly less overhead with no samples to collect or aggregate. This allows us to focus on the contribution of variables of interest, such as the prediction model, without interference from known quantities such as differences in segmentation performance between datasets.
\begin{figure*}[htpb]
    \centering
    \begin{subfigure}{1\linewidth}
        \includegraphics[width=1\linewidth]{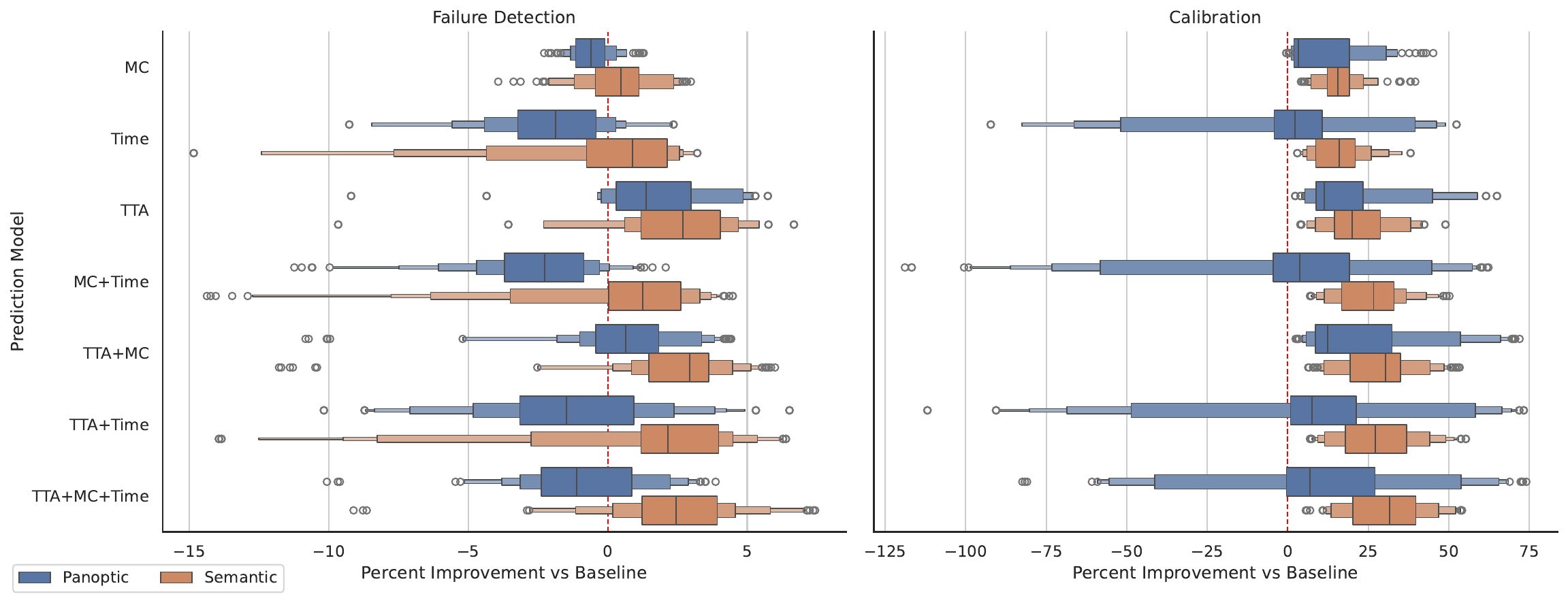}
        \caption{}
        \label{fig:iid_fd_calib_results_predmodel}
    \end{subfigure}
    \begin{subfigure}{1\linewidth}
        \includegraphics[width=1\linewidth]{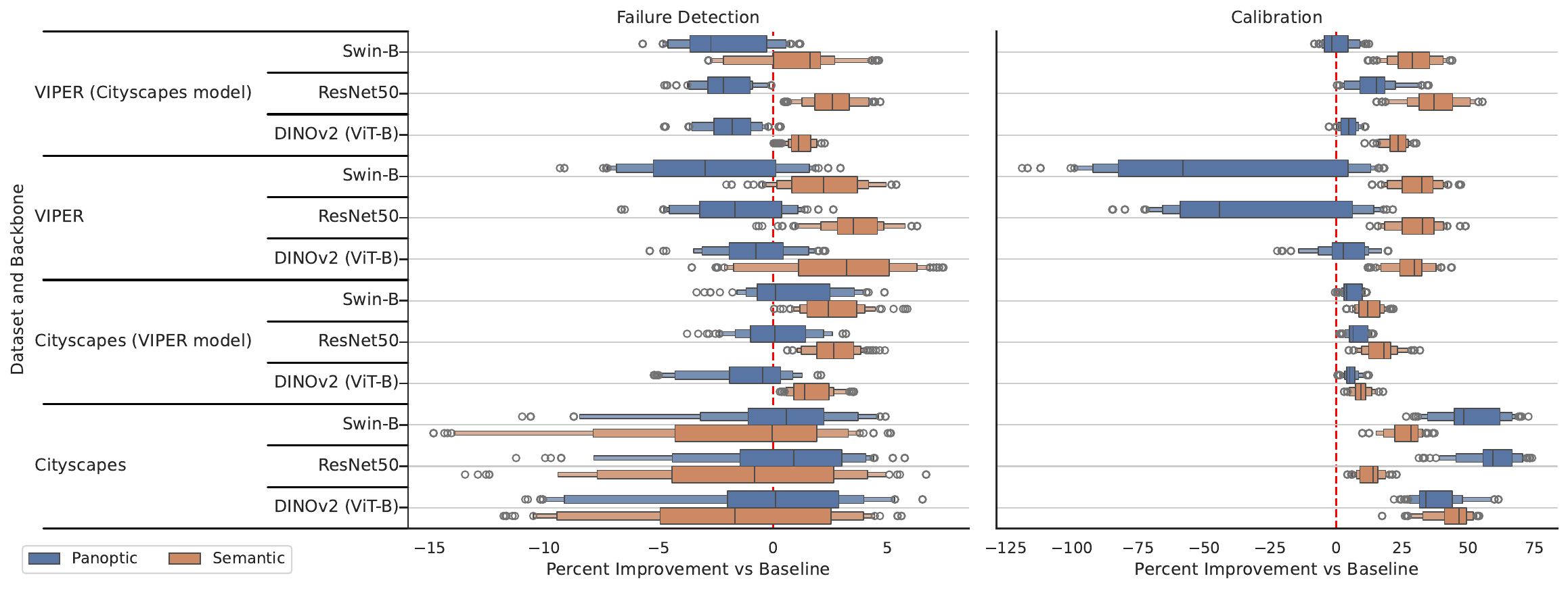}
        \caption{}
        \label{fig:iid_fd_calib_dsback}
    \end{subfigure}
    \caption{Results on the downstream tasks (\labelcref{param:downstream_task}) of failure detection and calibration, plotted by (a) prediction model \labelcref{param:pred_model} and by (b) dataset (\labelcref{param:dataset}) and backbone (\labelcref{param:backbone}). The letter-value plot \cite{hofmann_letter-value_2017} is an extension of the box plot, with the median shown by the middle line, the largest box representing 50\% of the data, and each subsequent smaller box representing an additional half ($75\%$, $87.5\%$, \etc). Circles represent outliers. Note that all results are \textbf{relative} to a baseline defined as the best possible result achievable with a given dataset, model and no sampling (\ie without MC Dropout, TTA, etc.). Variables \labelcref{param:dataset,param:backbone,param:pred_model,param:domain} are swept through and shown, \labelcref{param:sample_agg} is fixed to the best performer \textit{Mask Distance}, and \labelcref{param:uncert_measure,param:pixel_agg} are optimized over for the failure detection case where they apply.}
    \label{fig:iid_fd_calib_results_combined}
    \end{figure*}

\begin{figure}[htpb]
    \centering
    \begin{subfigure}{1\linewidth}
        \includegraphics[width=1\linewidth]{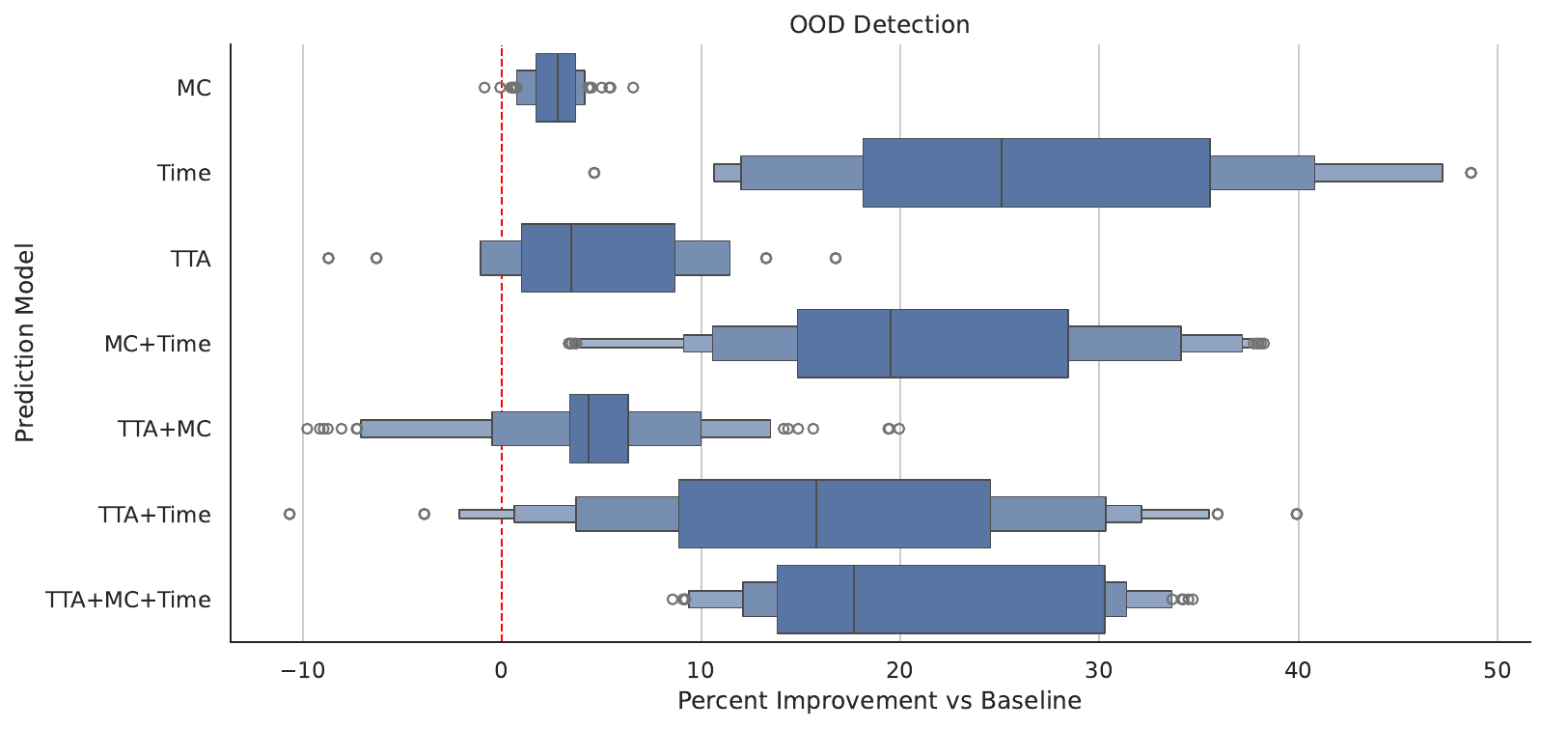}
        \caption{}
        \label{fig:ood_results_predmodel}
    \end{subfigure}
    \begin{subfigure}{1\linewidth}
        \includegraphics[width=1\linewidth]{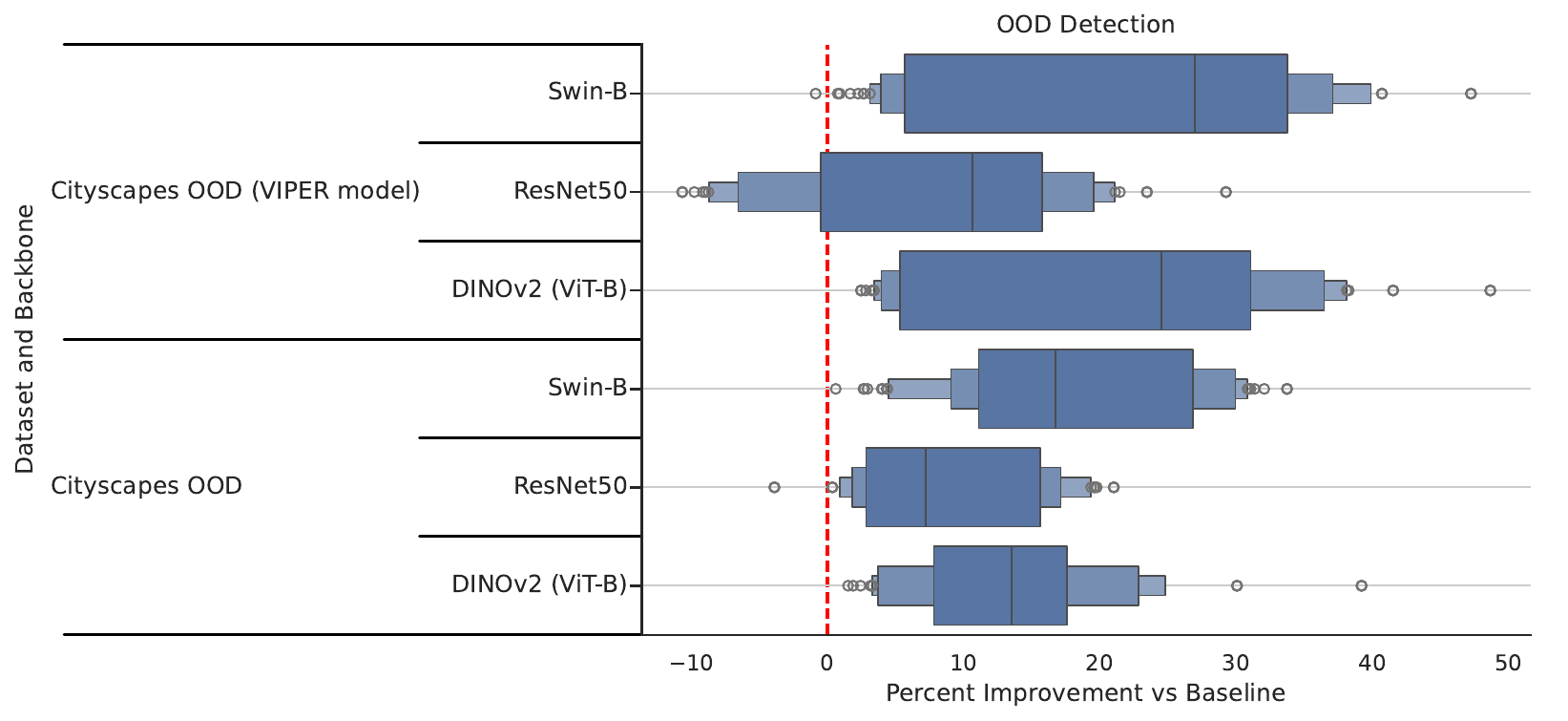}
        \caption{}
        \label{fig:ood_results_dsback}
    \end{subfigure}
    \caption{Results on the out-of-distribution detection downstream task (\labelcref{param:downstream_task}), by (a) prediction model and (b) dataset and backbone. Plot choice and experiment configuration follow \cref{fig:iid_fd_calib_results_combined} except \labelcref{param:pixel_agg,param:uncert_measure} which are applicable while \labelcref{param:domain} is not.}
    \label{fig:ood_results_combined}
\end{figure}

\textbf{\ref{q:domain_dataset_backbone}}: As seen in \cref{fig:iid_fd_calib_results_combined,fig:fd_calib_by_dataset_backbone_raw}, under most conditions relative and absolute performance for the panoptic version of a given metric is worse than the semantic equivalent (even including segmentation performance shown in \cref{fig:app:seg_perf_combined,fig:seg_perf_raw_dataset_backbone}). Thus, \textit{changing domains can harm even the relative benefit of uncertainty estimation}. When it comes to changes in dataset and backbone, shown in \cref{fig:iid_fd_calib_dsback,fig:ood_results_dsback} there is no particular combination that allows for repeatable improvement over a baseline under the same conditions. Relative performance is generally consistent albeit noisy, with a few notable outliers \eg panoptic calibration on VIPER vs. Cityscapes. These outliers mean that \textit{even small changes such as a different dataset or backbone can have cascading effects on uncertainty performance} on top of segmentation quality. Distribution shifts do not appear to cause any more variance than dataset or backbone changes when normalized; of course absolute performance as seen in \cref{fig:fd_calib_by_dataset_backbone_raw} is a different story with massive performance losses. \textit{It thus calls into question the real-world generalization of any uncertainty estimation approach without explicit validation on relevant parameters}. In terms of absolute performance, there is also no clear winner as shown in \cref{fig:fd_calib_by_dataset_backbone_raw}.

\textbf{\ref{q:samplediversity}}: In \cref{fig:iid_fd_calib_results_predmodel,fig:ood_results_predmodel} we see that \textit{exploiting sample diversity} via combinations of prediction models, such as TTA and MC Dropout, \textit{can lead to improved performance on select scenarios such as calibration under semantic segmentation}. In most other cases, however, we do not see the same benefits, with combination performance functionally a biased rough average of the constituent components with the possibility to sometimes go a little beyond. We do not see a clear improvement in performance between individual prediction models and combinations thereof, such as we see when we compare MC Dropout and TTA for failure detection. Under most conditions, the best performance can be achieved with either a single approach such as TTA or, in panoptic failure detection, a simple deterministic baseline.

\textbf{\ref{q:time}}: The idea of using prior \textit{time series frames appears to be useful in several scenarios, but not all}. We qualify the claim of \cite{huang_efficient_2018} that time series can be substituted for MC Dropout \cite{gal_dropout_2016} with the observation that their performance can diverge under some conditions such as panoptic failure detection (\cref{fig:iid_fd_calib_results_predmodel}), making a direct replacement unwise. We note that the approach has significant potential in any deployment where the sudden appearance of unexpected objects is undesirable (\cref{fig:ood_results_predmodel}), although this exceptional performance (including relevant combinations) is somewhat of an artifact of the experimental setup. This is because out-of-distribution objects do not exist in prior frames, making it relatively easy to detect the aberration. Beyond this, however, \emph{we feel that time series imposes overly burdensome requirements (existence of prior frame data, optical flow correspondence) that limit its utility.}

\textbf{\ref{q:baseline_usefulness_uncert}}: Across \cref{fig:iid_fd_calib_results_combined,fig:ood_results_combined}, we see a high degree of variation for results normalized to a baseline as a result of changes across numerous variables such as dataset, domain, prediction model or downstream task. Prediction models can do \emph{worse} than baseline in panoptic failure detection but then bring double-digit percentage improvements in calibration in \cref{fig:iid_fd_calib_results_predmodel} and OOD detection in \cref{fig:ood_results_predmodel}. Under a distribution shift (likely in some real-world scenarios), we see absolute performance collapse in \cref{fig:fd_calib_by_dataset_backbone_raw}. Yet, at the same time relative performance in \cref{fig:ood_results_dsback,fig:iid_fd_calib_dsback} swings from an outlier low of $-100\%$ in panoptic calibration to somewhat consistent double digit percentage improvements in OOD detection over a deterministic model. For a practitioner attempting to deploy a model, it is thus \emph{very difficult to know what configuration is best and how it might perform.}. Deploying a baseline first (in a non-critical scenario) is likely a decent starting point. \emph{Clearly, the general objective of any uncertainty estimation approach - to assist with real-world deployment - seems far from solved. If a real world deployment allows setting the right conditions, however, double-digit percentage improvements can make implementing uncertainty estimation worth the cost.} A corollary to this is that research on uncertainty estimation should be thoroughly evaluated under realistic scenarios if possible so as to avoid overpromising potential benefits.

\textbf{Prior Work Comparison}: \cite{roschewitz_where_2025} show that ensembling improves calibration, which we also see in \cref{fig:iid_fd_calib_results_combined}, if we ignore the poor panoptic calibration under certain conditions such as the VIPER dataset. \cite{kahl_values_2024} find that TTA makes for a simple but effective standard approach; given our results in \cref{fig:iid_fd_calib_results_predmodel,fig:ood_results_predmodel} we concur. We also agree that joint consideration of all variables is essential, although the number of variables makes it challenging. \cite{wen_combining_2021} find a pathology between data augmentation (TTA) and ensembles (MC Dropout), but this does not appear in our results. We hypothesize that the pathology they see is a result of Mixup \cite{zhang_mixup_2018} specifically and not data augmentation in general. In the BRAVO challenge \cite{vu_bravo_2025}, foundation models dominated the rankings, including those based on DINOv2. While we do not wish to speculate on the causes, given the differences in experimental procedure between us and the challenge, we do not see the same dominance or robustness under distribution shift in \cref{fig:fd_calib_by_dataset_backbone_raw,fig:app:ood_auroc_datasec_backbone_raw,fig:seg_perf_raw_dataset_backbone}. Further exploration is warranted, although our results are also limited by the difficulty we encountered training DINOv2 as discussed in \cref{subsec:app:foundation_model}. \cite{jaeger_call_2023} find that a baseline softmax outperforms more complex alternatives in image classification, but we do not see the same pattern in our baselines except for panoptic failure detection. 

For additional results, please see \cref{sec:app:addtional_results}.

\begin{figure}[htpb]
    \centering
    \begin{subfigure}{1\linewidth}
        \includegraphics[width=1\linewidth]{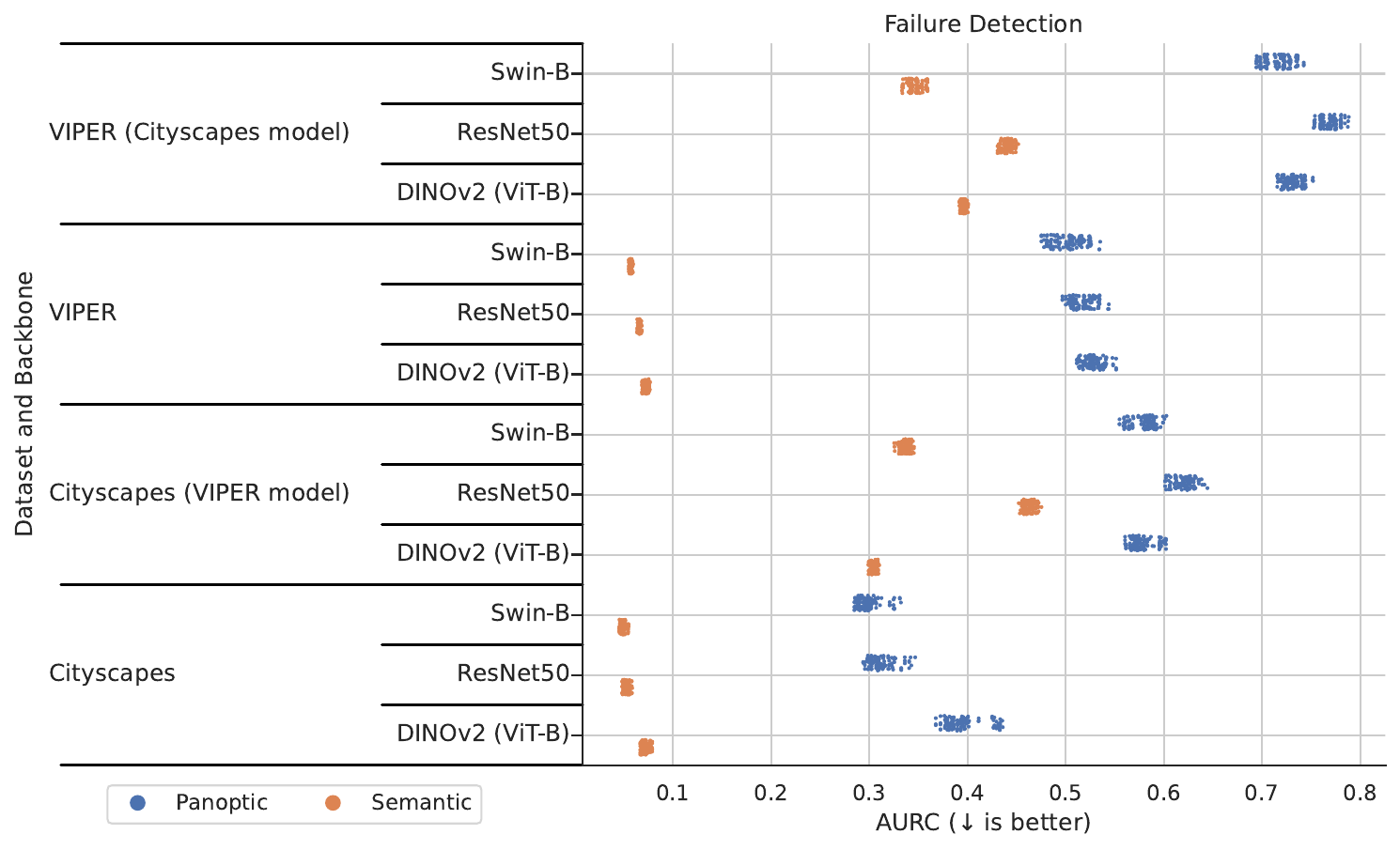}
        \caption{}
        \label{fig:fd_by_dataset_backbone_raw}
    \end{subfigure}
    \begin{subfigure}{1\linewidth}
        \includegraphics[width=1\linewidth]{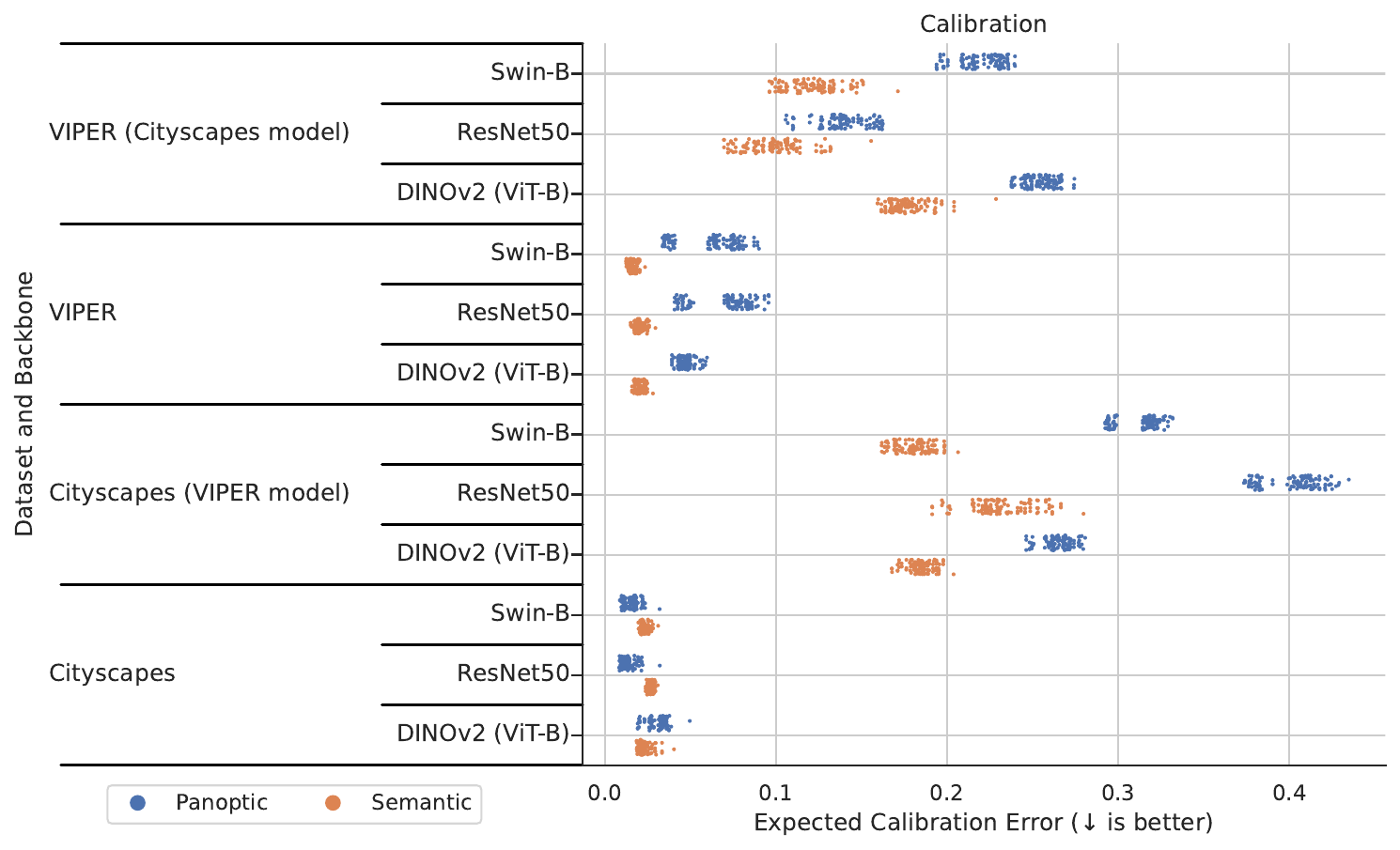}
        \caption{}
        \label{fig:calib_by_dataset_backbone_raw}
    \end{subfigure}
    \caption{Failure detection (a) and calibration (b) results following the experimental procedure of \cref{fig:iid_fd_calib_results_combined} without any normalization to a baseline. In this scatter plot each dot represents the results from a particular prediction model configuration and seed.}
    \label{fig:fd_calib_by_dataset_backbone_raw}
\end{figure}

\begin{figure}
    \centering
    \includegraphics[width=1\linewidth]{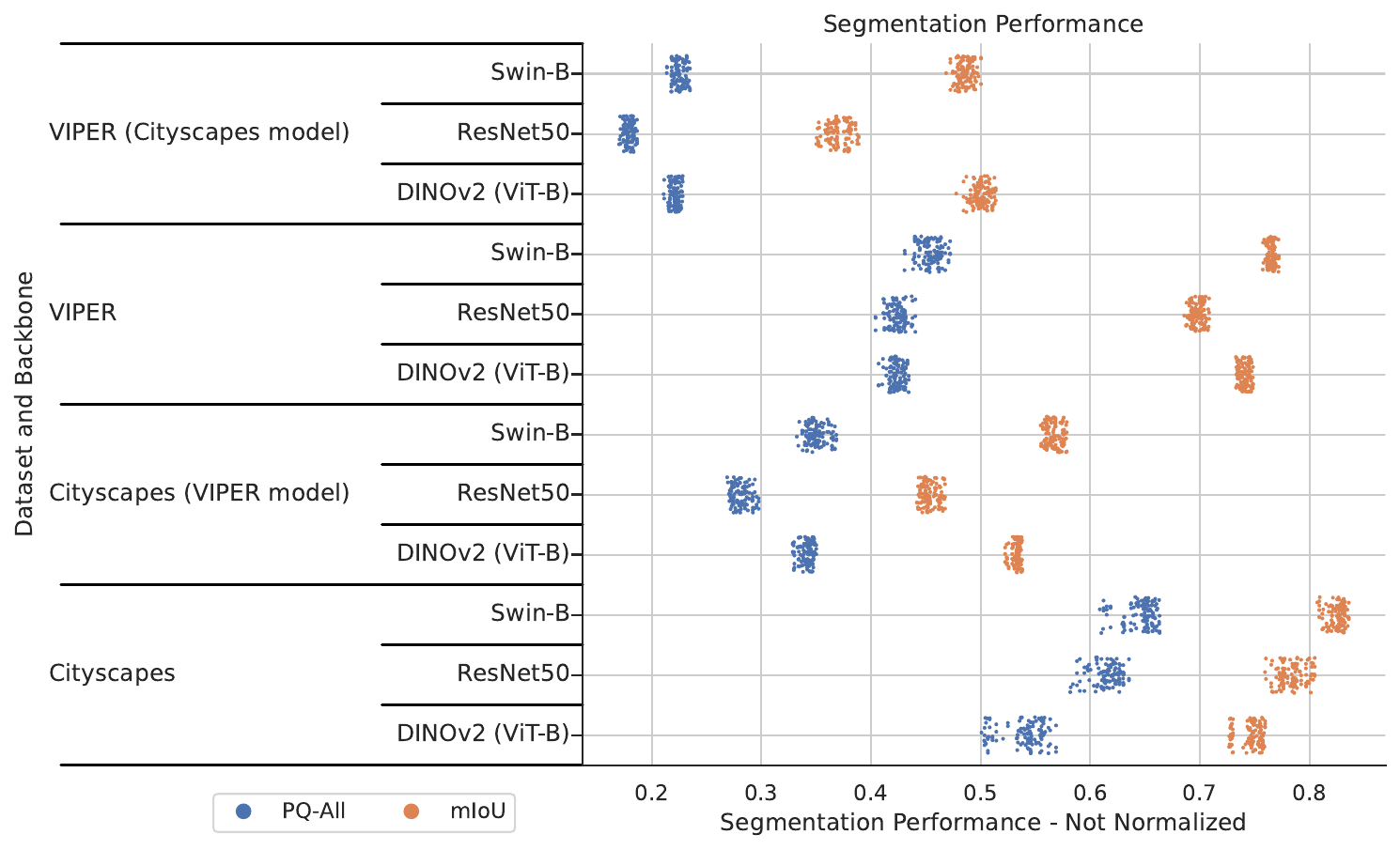}
    \caption{Segmentation performance by dataset and backbone with the PQ (panoptic) and mIoU (semantic) metrics. Note that the PQ metric is calculated across all classes, including both \textit{things} and \textit{stuff}. We follow the experimental procedure of \cref{fig:iid_fd_calib_results_combined}, except \labelcref{param:uncert_measure,param:pixel_agg,param:downstream_task} do not apply.}
    \label{fig:seg_perf_raw_dataset_backbone}
\end{figure}
\section{Conclusion}
\label{sec:conclusion}
We introduce a framework for evaluating uncertainty estimation on the semantic and panoptic segmentation tasks as well as \textit{Mask Distance}, a novel sample aggregation algorithm that enables it. This framework allows for the individual evaluation of many variables, including domains, datasets, backbones and prediction models. Using this framework, we conduct a large-scale study and answer a number of questions not thoroughly studied in the literature but relevant to practitioners. \emph{We find that individual and combinations of ensemble-based prediction models such as MC Dropout, Time-series or TTA can provide useful uncertainty estimates under many conditions. However, to successfully leverage such models in real-world deployments, care must be taken to ensure that the right conditions hold.}

\section{Acknowledgements}
NVIDIA provided one of the GPUs used during development. Most experiments were made possible by support provided by Calcul Québec, Compute Ontario, the BC DRI Group, and the Digital Research Alliance of Canada.

% CVPR template reference section
{
    \small
    \bibliographystyle{ieeenat_fullname}
    \bibliography{main}
}

\clearpage
\appendix
\setcounter{page}{1}
\renewcommand\thesection{A\arabic{section}}
\renewcommand\thefigure{A\arabic{figure}}
\renewcommand\thetable{A\arabic{table}}
\renewcommand\thepage{A-\arabic{page}}
\setcounter{figure}{0}
\setcounter{table}{0}
\maketitlesupplementary

\section{Implementation Details}
\label{sec:app:implementation_details}

\subsection{Datasets}
\label{subsec:app:datasets}
As discussed in \cref{sec:study_approach}, we evaluate over the VIPER \cite{richter_playing_2017} and Cityscapes \cite{cordts_cityscapes_2016} datasets, both of which are driving-focused. This somewhat limited selection is due to the requirements we have for our datasets. First, they must have a minimum level of complexity, as we are evaluating uncertainty estimation approaches whose primary goal is to assist with real world issues. Second, we wish to be able to evaluate distribution shifts, which requires distinct datasets that have significant annotation overlap. Third, in order to evaluate out-of-distribution detection we need to be able to acquire appropriate out-of-distribution images. Fourth, the time series prediction model requires prior frames to exist and to be close enough (in time/space) to the annotated frames such that we can align them via optical flow or some other method. Lastly, the dataset needs to have panoptic annotations or, as in our case, semantic and instance annotations that can be converted to panoptic.

Both datasets are high resolution, with images being a minimum of 1024 pixels in the smallest dimension before any processing. The Cityscapes and VIPER training set include 2975 and 13367 images respectively, with their validation sets including 500 and 4959. When it comes to time series data, Cityscapes provides un-annotated frames for an interval before and after the annotated frames we evaluate on, while in VIPER all frames are annotated but we choose to only use primary set consisting of an annotation for every 10 frames so as to effectively match the Cityscapes configuration. During evaluation we only use at most 5 prior frames.

We note that one of the limitations of our study is the exclusive use of driving related datasets. We emphasize that this is not a requirement of our proposed approach or our study. We believe that our findings generalize to other domains, but to our knowledge there are no non-driving datasets that match our requirements for us to evaluate on.

With respect to the distribution shift variants of each dataset, please see \cref{subsec:app:dist_shift_details,tab:app:viper_to_cityscapes_mapping,tab:app:cityscapes_to_viper_mapping}.

\subsection{Optical Flow}
\label{subsec:app:optical_flow}
For the VIPER dataset, ground truth optical flow provided by the authors is used. This data is high quality as the VIPER dataset is artificially generated, but there are some gaps in the data where an ``interruption'' occurs (consecutive frames in a sequence have a much larger gap than normal and lack the relevant flow data). There are also parts of any given image where there is simply no flow data \eg any pixels with a pedestrian.

For the Cityscapes dataset, we use the popular RAFT \cite{teed_raft_2020} to generate the optical flow. We use the \texttt{torchvision} implementation and the \texttt{raft\_large} variant with weights \texttt{DEFAULT} as we found them to work best. Default model parameters are used, with images resized to $\frac{3}{4}$ normal size (\ie $1536 \times 768$) due to computational constraints.

In both datasets, there are many instances of gaps where there are no prior time series frames. Beyond some missing frames in the VIPER dataset, these are primarily due to the start of a new driving sequence. In such cases, we do not use any prior time series frames, meaning that no prior frames can be sampled if the time series prediction model is in use for those frames. Such occurrences are rare however and should not meaningfully affect our results.

\subsection{Distribution Shifts}
\label{subsec:app:dist_shift_details}

\begin{table*}[t]         
    \centering
        \begin{tabular}{S[table-format=2]lccS[table-format=2]lc}
        \toprule
        \multicolumn{3}{c}{\textbf{VIPER}} & → & \multicolumn{3}{c}{\textbf{Cityscapes}} \\
        \cmidrule(lr){1-3} \cmidrule(lr){5-7}
        \textbf{ID} & \textbf{Class Name} & \textbf{Instances} & & \textbf{ID} & \textbf{Class Name} & \textbf{Instances} \\
        \midrule
        13  & \texttt{trafficlight}   & \cmark  & & 19 & \texttt{traffic light}   & \xmark \\
        16  & \texttt{firehydrant}    & \cmark  & & 0  & \texttt{VOID}            & \xmark \\
        17  & \texttt{chair}          & \cmark  & & 0  & \texttt{VOID}            & \xmark \\
        19  & \texttt{trashcan}       & \cmark  & & 0  & \texttt{VOID}            & \xmark \\
        20  & \texttt{person}         & \cmark  & & 24 & \texttt{person}          & \cmark  \\
        23  & \texttt{motorcycle}     & \cmark  & & 32 & \texttt{motorcycle}      & \cmark  \\
        24  & \texttt{car}            & \cmark  & & 26 & \texttt{car}             & \cmark  \\
        25  & \texttt{van}            & \cmark  & & 26 & \texttt{car}             & \cmark  \\
        26  & \texttt{bus}            & \cmark  & & 28 & \texttt{bus}             & \cmark  \\
        27  & \texttt{truck}          & \cmark  & & 27 & \texttt{truck}           & \cmark  \\
        2   & \texttt{sky}            & \xmark & & 23 & \texttt{sky}            & \xmark \\
        3   & \texttt{road}           & \xmark & & 7  & \texttt{road}           & \xmark \\
        4   & \texttt{sidewalk}       & \xmark & & 8  & \texttt{sidewalk}       & \xmark \\
        6   & \texttt{terrain}        & \xmark & & 22 & \texttt{terrain}        & \xmark \\
        7   & \texttt{tree}           & \xmark & & 21 & \texttt{vegetation}    & \xmark \\
        8   & \texttt{vegetation}     & \xmark & & 21 & \texttt{vegetation}    & \xmark \\
        9   & \texttt{building}       & \xmark & & 11 & \texttt{building}       & \xmark \\
        10  & \texttt{infrastructure} & \xmark & & 0  & \texttt{VOID}           & \xmark \\
        11  & \texttt{fence}          & \xmark & & 13 & \texttt{fence}          & \xmark \\
        12  & \texttt{billboard}      & \xmark & & 0  & \texttt{VOID}           & \xmark \\
        14  & \texttt{trafficsign}    & \xmark & & 20 & \texttt{traffic sign}   & \xmark \\
        15  & \texttt{mobilebarrier}  & \xmark & & 0  & \texttt{VOID}           & \xmark \\
        18  & \texttt{trash}          & \xmark & & 0  & \texttt{VOID}           & \xmark \\
        0   & \texttt{VOID}           & \xmark & & 0  & \texttt{VOID}           & \xmark \\
        \bottomrule
        \end{tabular}
    \caption{Mapping of labels from VIPER to Cityscapes.  \texttt{VOID} denotes the background/ignore class; the Instances column indicates whether instance annotations are provided in the corresponding dataset. ID numbers follow the respective dataset definitions as used in ground truth. Following standard practice only classes that are normally evaluated (\ie with a valid contiguous training ID) are considered.}
    \label{tab:app:viper_to_cityscapes_mapping}
\end{table*}

\begin{table*}
    \centering
    \begin{tabular}{S[table-format=2]lccS[table-format=2]lc}
        \toprule
        \multicolumn{3}{c}{\textbf{Cityscapes}} & → & \multicolumn{3}{c}{\textbf{VIPER}}\\
        \cmidrule(lr){1-3} \cmidrule(lr){5-7}
        \textbf{ID} & \textbf{Class Name} & \textbf{Instances} & & \textbf{ID} & \textbf{Class Name} & \textbf{Instances} \\
        \midrule
        7  & \texttt{road}          & \xmark & & 3  & \texttt{road}          & \xmark \\
        8  & \texttt{sidewalk}      & \xmark & & 4  & \texttt{sidewalk}      & \xmark \\
        11 & \texttt{building}      & \xmark & & 9  & \texttt{building}      & \xmark \\
        12 & \texttt{wall}          & \xmark & & 10 & \texttt{infrastructure} & \xmark \\
        13 & \texttt{fence}         & \xmark & & 11 & \texttt{fence}         & \xmark \\
        17 & \texttt{pole}          & \xmark & & 10 & \texttt{infrastructure} & \xmark \\
        19 & \texttt{traffic light}& \xmark & & 13 & \texttt{trafficlight}  & \cmark \\
        20 & \texttt{traffic sign} & \xmark & & 14 & \texttt{trafficsign}   & \xmark \\
        21 & \texttt{vegetation}    & \xmark & & 8  & \texttt{vegetation}    & \xmark \\
        22 & \texttt{terrain}       & \xmark & & 6  & \texttt{terrain}       & \xmark \\
        23 & \texttt{sky}           & \xmark & & 2  & \texttt{sky}           & \xmark \\
        24 & \texttt{person}        & \cmark & & 20 & \texttt{person}        & \cmark \\
        25 & \texttt{rider}         & \cmark & & 20 & \texttt{person}        & \cmark \\
        26 & \texttt{car}           & \cmark & & 24 & \texttt{car}           & \cmark \\
        27 & \texttt{truck}         & \cmark & & 27 & \texttt{truck}         & \cmark \\
        28 & \texttt{bus}           & \cmark & & 26 & \texttt{bus}           & \cmark \\
        31 & \texttt{train}         & \cmark & & 0  & \texttt{VOID}          & \xmark \\
        32 & \texttt{motorcycle}    & \cmark & & 23 & \texttt{motorcycle}    & \cmark \\
        33 & \texttt{bicycle}       & \cmark & & 0  & \texttt{VOID}          & \xmark \\
        0  & \texttt{VOID}          & \xmark & & 0  & \texttt{VOID}          & \xmark \\
        \bottomrule
    \end{tabular}
    \caption{Mapping of labels from Cityscapes to VIPER. \texttt{VOID} denotes the background/ignore class; the Instances column indicates whether instance annotations are provided in the corresponding dataset. ID numbers follow the respective dataset definitions as used in ground truth. Following standard practice only classes that are normally evaluated (\ie with a valid contiguous training ID) are considered.}
    \label{tab:app:cityscapes_to_viper_mapping}
\end{table*}

In \cref{tab:app:viper_to_cityscapes_mapping,tab:app:cityscapes_to_viper_mapping} we provide the mapping of classes from VIPER \cite{richter_playing_2017} to Cityscapes \cite{cordts_cityscapes_2016} and vice-versa. We use these in the distribution shift scenario where we evaluate a model trained on one dataset on the other. While both datasets are focused on driving, and thus have significant overlap in definitions, it is not always one-to-one. For example, the VIPER \texttt{infrastructure} class represents a number of objects such as walls, overpasses and street lights. However, we are forced to map it to Cityscapes \texttt{VOID} as there is no equivalent and no way to split out the matching objects such as walls. Going the other way, however, we can map \texttt{wall} in Cityscapes to \texttt{infrastructure} in VIPER. We further note that while most classes are treated the same in panoptic segmentation, being semantic only labels or including instance annotations, there are some differences such as \texttt{traffic light}. Going from a model trained to expect that class as one requiring individual instance annotation to evaluation that does not should not cause any issues. The opposite direction, however, would be more problematic and prone to errors where the instance IDs are not properly assigned. The general design of the network where potential objects are predicted (\cref{eq:model_output_logits_mask}) means that an error is not guaranteed, though, and correct instance labels are still possible.

\subsection{Time series}
\label{subsec:app:time_series_details}
Our implementation of using time series frames in an ensemble is primarily limited by errors in the alignment of prior frames to the current one. In the case of Cityscapes, we have any errors made by RAFT, while for VIPER even the ground truth remains limited by standard optical flow constraints such as occlusions. The effect of these errors can be seen in \cref{subsec:app:downsteram_failure_detect_calib_ood_results} and most notably in \cref{fig:app:iid_failure_detect_detailed_predmodel} where panoptic segmentation, which is more sensitive to any errors given the problem definition, is hampered more than semantic segmentation as the number of prior frames increases. \cite{huang_efficient_2018} propose a second method that relies on reconstruction error to address this, but we elected to not implement it as we did not wish to add further (potentially confounding) variables to our study.

\subsection{Foundation Model Performance}
\label{subsec:app:foundation_model}
As mentioned in \cref{subsec:key_variables_list,sec:study_approach}, we were inspired by the performance of foundation models like DINOv2 \cite{oquab_dinov2_2024} on the BRAVO challenge, where they demonstrated excellent performance on a range of challenging tasks including out-of-distribution detection, calibration or simply handling adverse weather conditions. As we explain in \cref{sec:results}, however, we do not see the same dominance with another backbone performing the same or better in many cases. For failure detection in particular, which is very sensitive to segmentation performance, our inability to train a DINOv2 backbone to match or exceed a Swin-B or ResNet50 alternative is likely a notable contributing factor.

In \cref{tab:app:dinov2_vs_other_backbones}, we compare our trained DINOv2 models against a ResNet50 on panoptic and semantic segmentation performance. We see that for the Cityscapes and VIPER datasets, DINOv2 performance lags a basic ResNet50 (with the exception of mIoU for VIPER). However, on the more diverse ADE20K dataset, a smaller DINOv2 variant with the ViT-S architecture is easily able to beat a ResNet50. We made many attempts to improve DINOv2 performance, including different transfer learning sources, longer training, and freezing the DINOv2 model except the \textit{qkv} and/or projections layers. All were unsuccessful, and we ultimately conclude that the lack of diversity in Citycapes and VIPER is the primary cause given the ADE20K performance.
\begin{table}[t]
    \centering
    \begin{tabular}{lccc}
    \toprule
    \textbf{Backbone} & \textbf{Dataset} & \textbf{PQ} & \textbf{mIoU} \\
    \midrule
    DINOv2 (ViT‑B) & Cityscapes & 55.73 & 75.18 \\
    ResNet50$^{\text{\textdagger}}$ & Cityscapes & \textbf{62.09} & \textbf{77.41} \\
    \addlinespace
    DINOv2 (ViT‑S) & ADE20K    & \textbf{42.99} & \textbf{50.06} \\
    ResNet50$^{\text{\textdagger}}$      & ADE20K    & 39.51 & 45.81 \\
    \addlinespace
    DINOv2 (ViT‑B) & VIPER     & 42.72 & \textbf{73.76} \\
    ResNet50      & VIPER     & \textbf{43.39} & 69.07 \\
    \bottomrule
    \end{tabular}
    \caption{Segmentation performance comparison of different backbones across datasets using a baseline (deterministic) Mask2Former model. All runs use a fixed seed of 1. $^{\text{\textdagger}}$ indicates model provided by \cite{cheng_masked-attention_2022}; training is done by us otherwise.}
    \label{tab:app:dinov2_vs_other_backbones}
\end{table}

\subsection{Training parameters}
Where possible, trained models from \cite{cheng_masked-attention_2022} were used, covering ResNet50 and Swin-B on Cityscapes. All other models were trained by us. We attempted to follow as best as possible the training parameters used by \cite{cheng_masked-attention_2022}, including the number of epochs (steps), preprocessing settings, data augmentations, optimizer settings, learning rates, and other model parameters. For ResNet50 and Swin-B on VIPER, transfer learning is used from corresponding COCO \cite{lin_microsoft_2014} trained models. For DINOv2 on VIPER, the Mask2Former head is initialized with weights from a COCO-trained Swin-B model, while on Cityscapes it is randomly initialized. Transfer learned models are sourced from \cite{cheng_masked-attention_2022}. The DINOv2 backbone itself is initialized with weights from \cite{oquab_dinov2_2024} from the \texttt{ViT-B/14 distilled with registers} model for both datasets.

\subsection{Calibration}
For calibration, we use 15 bins following \cite{guo_calibration_2017,vu_bravo_2025}; most literature appears to use a value between 10 and 20. As well, following \cite{minderer_revisiting_2021} we implement argmax calibration, where only the calibration of the highest likelihood score is evaluated.

\section{Prediction Models}

\begin{table*}[t]
    \centering
    \begin{tabular}{
        S[table-format=2]      % MC Samples
        S[table-format=1]      % #Prev. time frames
        l                      % TTA transform
        S[table-format=2]      % Total # samples
        c                      % Mask distance
        c                      % Pixel decoder
        c                      % Averaging (Smith et al., 2024)
    }
        \toprule
        \multicolumn{3}{c}{\textbf{Prediction Model Configuration}} & & \multicolumn{3}{c}{\textbf{Applicable Sample Aggregation (\labelcref{param:sample_agg})}} \\
        \cmidrule(lr){1-3} \cmidrule(lr){5-7} 
        \textbf{MC Drop.} & \textbf{Prev. Time Frames} & \textbf{TTA Transf.} & \textbf{Total \# Samples} & \textbf{Mask Dist.} & \textbf{Pixel Dec.} & \textbf{Averaging}\cite{smith_uncertainty_2024} \\
        \midrule
        0 & 0 & None                         & 1  & - & - & - \\
        0 & 0 & Horizontal Flip             & 2  & \cmark & \cmark & \cmark \\
        0 & 0 & Scale + Horizontal Flip     & 6  & \cmark & \xmark & \cmark \\
        0 & 0 & Scale                       & 3  & \cmark & \xmark & \cmark \\
        0 & 1 & None                         & 2  & \cmark & \cmark & \cmark \\
        0 & 1 & Horizontal Flip             & 4  & \cmark & \cmark & \cmark \\
        0 & 1 & Scale + Horizontal Flip     & 12 & \cmark & \xmark & \cmark \\
        0 & 1 & Scale                       & 6  & \cmark & \xmark & \cmark \\
        0 & 2 & None                         & 3  & \cmark & \cmark & \cmark \\
        0 & 2 & Horizontal Flip             & 6  & \cmark & \cmark & \cmark \\
        0 & 2 & Scale + Horizontal Flip     & 18 & \cmark & \xmark & \cmark \\
        0 & 2 & Scale                       & 9  & \cmark & \xmark & \cmark \\
        0 & 3 & None                         & 4  & \cmark & \xmark & \xmark \\
        0 & 3 & Horizontal Flip             & 8  & \cmark & \xmark & \xmark \\
        0 & 3 & Scale                       &12  & \cmark & \xmark & \xmark \\
        0 & 5 & None                         & 6  & \cmark & \xmark & \xmark \\
        0 & 5 & Horizontal Flip             &12  & \cmark & \xmark & \xmark \\
        0 & 5 & Scale                       &18  & \cmark & \xmark & \xmark \\
        3 & 0 & None                         & 3  & \cmark & \cmark & \cmark \\
        3 & 0 & Horizontal Flip             & 6  & \cmark & \xmark & \xmark \\
        3 & 0 & Scale + Horizontal Flip     &18  & \cmark & \xmark & \xmark \\
        3 & 0 & Scale                       & 9  & \cmark & \xmark & \xmark \\
        3 & 1 & None                         & 6  & \cmark & \cmark & \cmark \\
        3 & 1 & Horizontal Flip             &12  & \cmark & \cmark & \cmark \\
        3 & 1 & Scale                       &18  & \cmark & \xmark & \cmark \\
        3 & 2 & None                         & 9  & \cmark & \xmark & \xmark \\
        3 & 2 & Horizontal Flip             &18  & \cmark & \xmark & \xmark \\
        3 & 3 & None                         &12  & \cmark & \xmark & \xmark \\
        3 & 5 & None                         &18  & \cmark & \xmark & \xmark \\
        5 & 0 & None                         & 5  & \cmark & \xmark & \xmark \\
        5 & 0 & Horizontal Flip             &10  & \cmark & \xmark & \xmark \\
        5 & 0 & Scale                       &15  & \cmark & \xmark & \xmark \\
        5 & 1 & None                         &10  & \cmark & \xmark & \xmark \\
        5 & 1 & Horizontal Flip             &20  & \cmark & \xmark & \xmark \\
        5 & 2 & None                         &15  & \cmark & \xmark & \xmark \\
        5 & 3 & None                         &20  & \cmark & \xmark & \xmark \\
        10& 0 & None                         &10  & \cmark & \xmark & \xmark \\
        10& 0 & Horizontal Flip             &20  & \cmark & \xmark & \xmark \\
        10& 1 & None                         &20  & \cmark & \xmark & \xmark \\
        \bottomrule
    \end{tabular}
    \caption{List of all prediction model configurations tested, alongside the applicable Sample Aggregation approach under which they were tested. All 39 configurations were tested with Mask Distance for most experiments. The first row is the baseline approach. Approaches other than Mask Distance were only used for the comparison of sample aggregation approaches in \cref{subsec:app:sample_agg_results}.}
    \label{tab:app:pred_model_config_list}
\end{table*}

In \cref{tab:app:pred_model_config_list}, we show all the prediction model configurations we test. Most experiments were evaluated with all of those shown with the Mask Distance sample aggregation (\labelcref{param:sample_agg}) method. The other two approaches are used only in \cref{subsec:app:sample_agg_results} for the comparison of different approaches. Fewer configurations were evaluated for the aggregation comparisons compared to the main experiments due to implementation limits (\eg the pixel decoder is incompatible with the scale TTA transform), resource constraints or computational limits. For example, the Averaging \cite{smith_uncertainty_2024} approach has GPU memory requirements linear with the number of samples; at higher levels, this requires very large GPUs. The Mask Distance approach in contrast has constant GPU memory requirements with respect to the number of samples.

Other prediction model parameters are as follows:
\begin{itemize}
    \item \textbf{MC Dropout}: 
    \begin{itemize}
        \item Dropout rate fixed at $p=0.1$.
        \item Dropout layers are only activated when MC Dropout is enabled. Thus, all non-MC Dropout configurations are deterministic.
        \item The only Dropout layers activated in the network are those in the head; specifically, in the pixel decoder. Following \cite{cheng_masked-attention_2022}, we use the multi-scale deformable attention Transformer (MSDeformAttn) \cite{zhu_deformable_2021} as the pixel decoder.
    \end{itemize}
    \item \textbf{TTA}: The following transforms are used in TTA:
    \begin{itemize}
        \item Horizontal flips, where the image is flipped horizontally. Flips were shown to perform well in \cite{kahl_values_2024}.
        \item Scale transformations: These result in two distinct samples from rescaling operations; the first at 0.8x and the second at 1.25x. Scale values were chosen following \cite{ghiasi_simple_2021}.
    \end{itemize}
\end{itemize}

Of note, unlike \cite{kahl_values_2024} we do not use Gaussian noise as we observed significant performance degradation. We believe this to be a result of the fact that such noise is not part of the normal augmentations applied during training.

\section{Sample Aggregation Details}
\label{sec:app:sample_agg_extended}
In this study, we introduce Mask Distance to address the problem of sample aggregation. This problem exists because of our choice of a universal architecture model (Mask2Former), which was in turn chosen as a result of our desire to cover both the panoptic and semantic segmentation domains. Complementary work such as \cite{kahl_values_2024} does not require such a step as the task of semantic segmentation produces a class distribution at each pixel, and aggregating them is as straightforward as taking the mean. With instances, however, we must first determine the correspondences.

Other work on uncertainty and panoptic segmentation such as \cite{sirohi_uncertainty-aware_2023} do not use ensembles and thus sample aggregation is not a problem. Were ensembles to be validated on such network designs, some form of sample aggregation would be needed; given the design of the panoptic fusion model in \cite{sirohi_uncertainty-aware_2023} for instance this is likely to be a complex process.

For a comparison of the various sample aggregation approaches, please see \cref{subsec:app:sample_agg_results,fig:app:agg_compare}.

\subsection{Mask Distance}
\label{subsec:app:mask_distance_extended}
As presented in \cref{subsec:mask_distance}, the key detail of the Mask Distance approach is that the relatively straightforward computation of the Euclidean distance between all the masks representing potential objects in each sample is sufficient to determine the correspondence without any consideration of classes. As presented, however, \cref{eq:euclidean_mask_distance} obfuscates a notable implementation detail: we calculate the Euclidean distance between the mask of one sample against the next, but only for the first two samples. Subsequent samples are then matched against a running average of the samples accumulated up until that point. This eliminates the need to keep more samples in limited GPU memory, as well as the potential for one or more poor quality samples to lead to bad or missing matches. The memory requirements for Mask Distance are constant with respect to the number of samples. This allows us to evaluate prediction models combinations with up to 20 samples on a single 12GB GPU.

\subsection{Pixel Decoder}
\label{subsec:app:pixel_decoder}
With the pixel decoder, we average the samples from the ensemble inside the segmentation head. Specifically, we calculate the average at the four outputs of the pixel decoder in Fig. 2 of \cite{cheng_masked-attention_2022}. We then execute the remainder of the head as normal, giving us a single prediction. Not having access to samples at the output of the network however limits us in terms of the variety of uncertainty measures we can calculate, as shown in \cref{tab:app:uncertainty_measures}. Attempting to merge intermediate network features also places some restrictions on applicable prediction models such as any TTA configuration with multiple scales, shown in \cref{tab:app:pred_model_config_list}. As this approach performs poorly in terms of uncertainty and segmentation performance, shown in \cref{sec:app:sample_agg_extended,fig:app:agg_compare}, we did not investigate this approach further in favour of Mask Distance. It does, however, do well in terms of the time needed to aggregate samples, as shown in \cref{sec:app:computational_cost}.

\section{Uncertainty Measures}
\label{sec:app:uncert_measures}

\begin{table*}
  \centering
    \begin{tabular}{lcccc}
        \toprule
        & \multicolumn{4}{c}{\textbf{Applicable Sample Aggregation}} \\ 
        \cmidrule(lr){2-5}
        \textbf{Uncertainty Measure} & \textbf{Baseline (None)} &
        \textbf{Mask Dist.} &
        \textbf{Pixel Dec.} &
        \textbf{Averaging \cite{smith_uncertainty_2024}} \\
        \midrule
        Predictive Entropy (Class \& Mask)                & \cmark & \cmark & \cmark & \cmark \\
        Predictive Entropy (Mask)                         & \cmark & \cmark & \cmark & \xmark \\
        Expected Entropy (Mask)                           & \xmark & \cmark & \xmark & \cmark \\
        Expected Entropy (Class \& Mask)                  & \xmark & \cmark & \xmark & \cmark \\
        Mutual Information (Mask)                         & \xmark & \cmark & \xmark & \xmark \\
        Mutual Information (Class \& Mask)                & \xmark & \cmark & \xmark & \cmark \\
        Expected Mask Variance                            & \xmark & \cmark & \xmark & \cmark \\
        Predictive Mask Variance                          & \cmark & \cmark & \cmark & \xmark \\
        Maximum Softmax Score (Class \& Mask)$^{\text{\textdagger}}$ & \cmark & \cmark & \cmark & \cmark \\
        Maximum Normalized Sigmoid Mask Score            & \cmark & \cmark & \cmark & \xmark \\
        Combined Maximum Softmax \& Normalized Sigmoid   & \cmark & \cmark & \cmark & \xmark \\
        \bottomrule
    \end{tabular}
    \caption{Uncertainty measures evaluated in the paper. A checkmark indicates that the measure can be calculated with the corresponding sample aggregation approach (\labelcref{param:sample_agg}), while an X indicates that it cannot be calculated. Baseline represents no sample aggregation \ie a deterministic network, where measures are mapped to the nearest approximation. For example, rather than predictive entropy following \cref{eq:predictive_mean} we use the softmax entropy \cite{mukhoti_deep_2021}. $^{\text{\textdagger}}$ indicates the measure is used for calibration experiments.}
    \label{tab:app:uncertainty_measures}
\end{table*}

In \cref{tab:app:uncertainty_measures}, we list all the uncertainty measures we calculate and indicate which sample aggregation methods (\labelcref{param:sample_agg}) they can be calculated with. We include both typical well-defined measures of uncertainty such as the Predictive Entropy (\cref{eq:pred_entropy}) as well as more ad-hoc versions such as the variance. As the network produces two outputs (\cref{eq:model_output_logits_mask}), we also ensure that we apply a variety of measures that are able to fully capture all information contained in said outputs. 

This results in two different general approaches: \textit{Class \& Mask} and \textit{Mask}. The former is applied by using the mask assignments as weights for a weighted sum of the softmax output following \cite{cheng_masked-attention_2022}. The result at each pixel is thus a class distribution whose composition has been influenced by potential assignments to one or more objects. For the latter approach, we ignore the class output and use only the mask output with measures such as the Expected Mask Variance or Mutual Information (Mask). This is inspired from prior work that such as \cite{morrison_estimating_2019,deery_propandl_2023} that explicitly calculate a measure of spatial uncertainty. Towards this end, we further introduce the Maximum Normalized Sigmoid Mask Score, which is simply the maximum of the mask output of the network passed through a sigmoid function and normalized to add to 1. This represents the maximum uncertainty the network has with respect to its instance assignments at each pixel. We further introduce the Combined Maximum Softmax \& Normalized Sigmoid, which is defined to be the Maximum Softmax Score at all pixels except those determined by the network to require instance annotations (\textit{things} in panoptic segmentation), where it is instead the mean of the Maximum Softmax Score and the Maximum Normalized Sigmoid Mask Score. These are very ad-hoc measures, but we wish to give downstream tasks the most choice possible as we optimize over the uncertainty measure in most experiments.

\section{Optimization}
In this work, in many experiments we chose to optimize over variables such as the uncertainty measure (\labelcref{param:uncert_measure}) or pixel aggregation approach (\labelcref{param:pixel_agg}). This allows the best performance possible for the variables under study, without introducing any bias as a result of fixing the choices as shown by \cite{kahl_values_2024}. As an example, assume the best possible result achievable with a deterministic baseline relies on Predictive Entropy, but with MC Dropout where we can calculate Expected Entropy (as we have $>1$ sample), we can do better. As shown in \cref{tab:app:uncertainty_measures}, there are a number of incompatible sample aggregation and uncertainty measure combinations. Going back to our example, if we fix the uncertainty measure to Expected Entropy, how could we fairly compare with a baseline? Going the other direction, if we fix the uncertainty measure to Predictive Entropy, we then ignore any performance gains achievable with other uncertainty measures that are not universally compatible, effectively holding them back. As we show in the main paper, in deployment a practitioner will need to make a choice, jointly considering all variables.

\section{Metrics}
\label{sec:app:metrics_extended}
For out-of-distribution detection, when calculating the AUROC metric we use the \texttt{roc\_auc\_score} function from scikit-learn \cite{pedregosa_scikit-learn_2011}. The uncertainty measures are treated as the target score, while the classification of the image as out-of-distribution or not serves as the label. For failure detection, we calculate per-image IoU and PQ metrics for semantic and panoptic segmentation. These per-image metrics are calculated across all classes; we cannot calculate standard metrics where an average is taken across all classes (\eg Sec. 4.2 in \cite{kirillov_panoptic_2019}) as unlike with an entire dataset there may only be a few classes present in each image. These are then used to calculate the risk and subsequent AURC metric following \cite{kahl_values_2024,jaeger_call_2023}.

For distribution shifted versions of both datasets, when evaluating segmentation metrics we ignore any classes for which there is no ground truth to evaluate. When evaluating panoptic segmentation, we always report the PQ metric for all categories, outside of per-image evaluation used in failure detection where it is not applicable.

In \cref{sec:study_approach}, we state that we attempt to preserve ``objective compatibility'' during evaluation. By this, we mean that we choose to use metrics in such a way that we can fairly compare them across semantic and panoptic segmentation. Obviously, direct comparisons are impossible, but relative comparisons as we show in many plots are possible given the right conditions. Specifically, the metrics need to be measuring a quantity directly tied to the domain, such that a change in domain results in an easy to understand change to the metric as a direct result of the domain change and nothing else. For semantic segmentation, we have the mIoU which represents the average across classes of per-pixel label assignment. When we go to panoptic segmentation, this becomes PQ, which also represents the average across classes but now of the correct segment assignment. This extends to failure detection (AURC-PQ vs AURC-IoU) and calibration (ECE). As we state in \cref{subsec:metrics}, for calibration we forgo existing proposed metrics such as pECE \cite{sirohi_uncertainty-aware_2023} as a reliance on segment matching in panoptic segmentation has no direct correlation in semantic segmentation. With our implementation, we evaluate the ECE at the pixel level for both domains; the difference lies in what is used to determine if a pixel is correct. In semantic segmentation, it is the class label; in panoptic segmentation, it is both the class label and the instance label as used in the PQ metric calculation.

As shown in \cref{tab:app:uncertainty_measures}, for all calibration experiments we use the Maximum Softmax Score (Class \& Mask). We chose this approach due to its simplicity and no potential for confounding variables. \cite{kahl_values_2024} by contrast use Platt scaling to be able to evaluate multiple sample aggregation measures, but this requires a held out dataset (which we do not have) and has the potential to add bias from the learned Platt scaling parameters. They also use ACE (Average Calibration Error) as their metric of choice due to the predominance of background annotated pixels in their experiments. As our datasets are not from the medical imaging domain, we do not experience this issue and thus the ECE is appropriate.

\section{Additional Results}
\label{sec:app:addtional_results}
\subsection{Segmentation Performance}
\label{subsec:app:seg_perf}

\begin{figure}
    \centering
    \begin{subfigure}{1\linewidth}
        \includegraphics[width=1\linewidth]{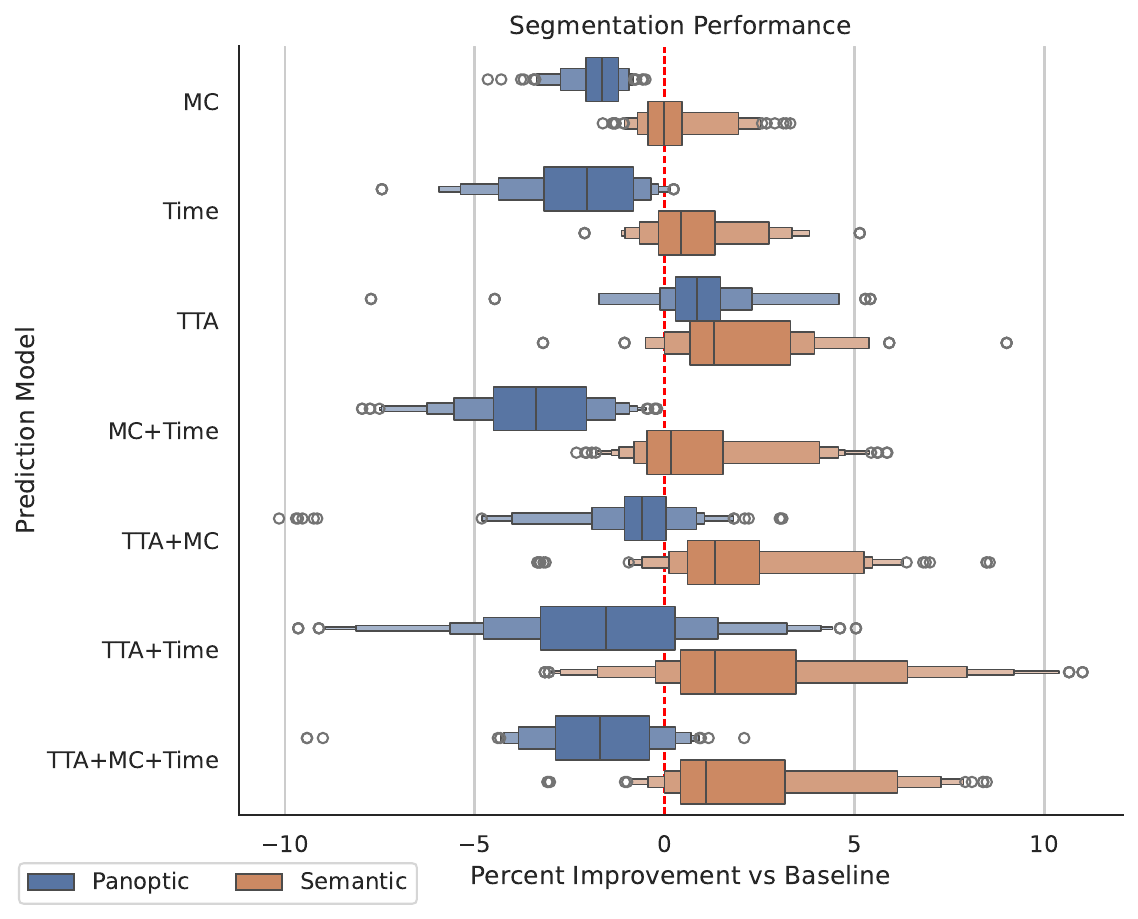}
        \caption{}
        \label{fig:app:seg_perf_predmodel}
    \end{subfigure}
    \begin{subfigure}{1\linewidth}
        \includegraphics[width=1\linewidth]{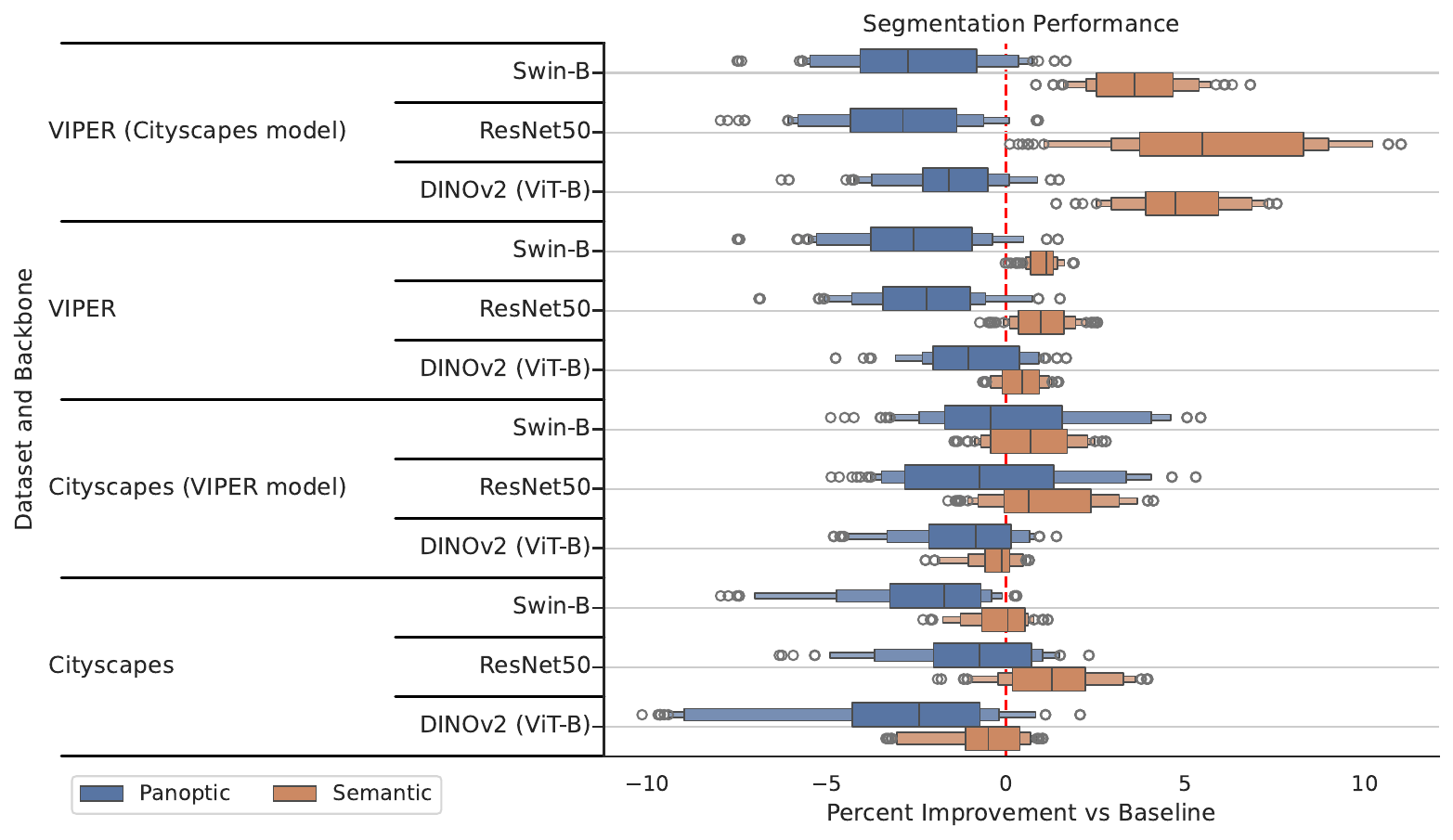}
        \caption{}
        \label{fig:app:seg_perf_dataset_backbone}
    \end{subfigure}
    \caption{Normalized segmentation performance, by (a) prediction model and (b) dataset and backbone. Segmentation performance is not a downstream task but nonetheless is important being a key part of the problem definition. Use of the letter-value plot and baseline definition follow \cref{fig:iid_fd_calib_results_combined}. Variables \labelcref{param:dataset,param:backbone,param:pred_model,param:domain} are swept through and shown, with \labelcref{param:sample_agg} fixed to the best performer \textit{Mask Distance}. Variables \labelcref{param:uncert_measure,param:pixel_agg} are not applicable for this task.}
    \label{fig:app:seg_perf_combined}
\end{figure}

In \cref{fig:app:seg_perf_combined}, we compare the normalized segmentation performance across various prediction models, datasets, and backbones. We see that the relative performance between the various categories roughly match that seen in the failure detection plots (\cref{fig:iid_fd_calib_results_combined}). Thus, while the objective of any uncertainty estimation approach is to generate a high quality uncertainty estimate, any unintended effects on segmentation performance can be a significant contributor to (potentially poor) performance on some downstream tasks for which uncertainty estimates are used such as failure detection. This calls into question the claims made by some that ``not overly compromising performance on the base task'' \cite{deery_propandl_2023} in exchange for better uncertainty estimates is acceptable, at least for downstream tasks such as failure detection.

\subsection{Out-of-distribution performance}
\label{subsec:app:ood_perf}
\begin{figure}
    \centering
    \includegraphics[width=1\linewidth]{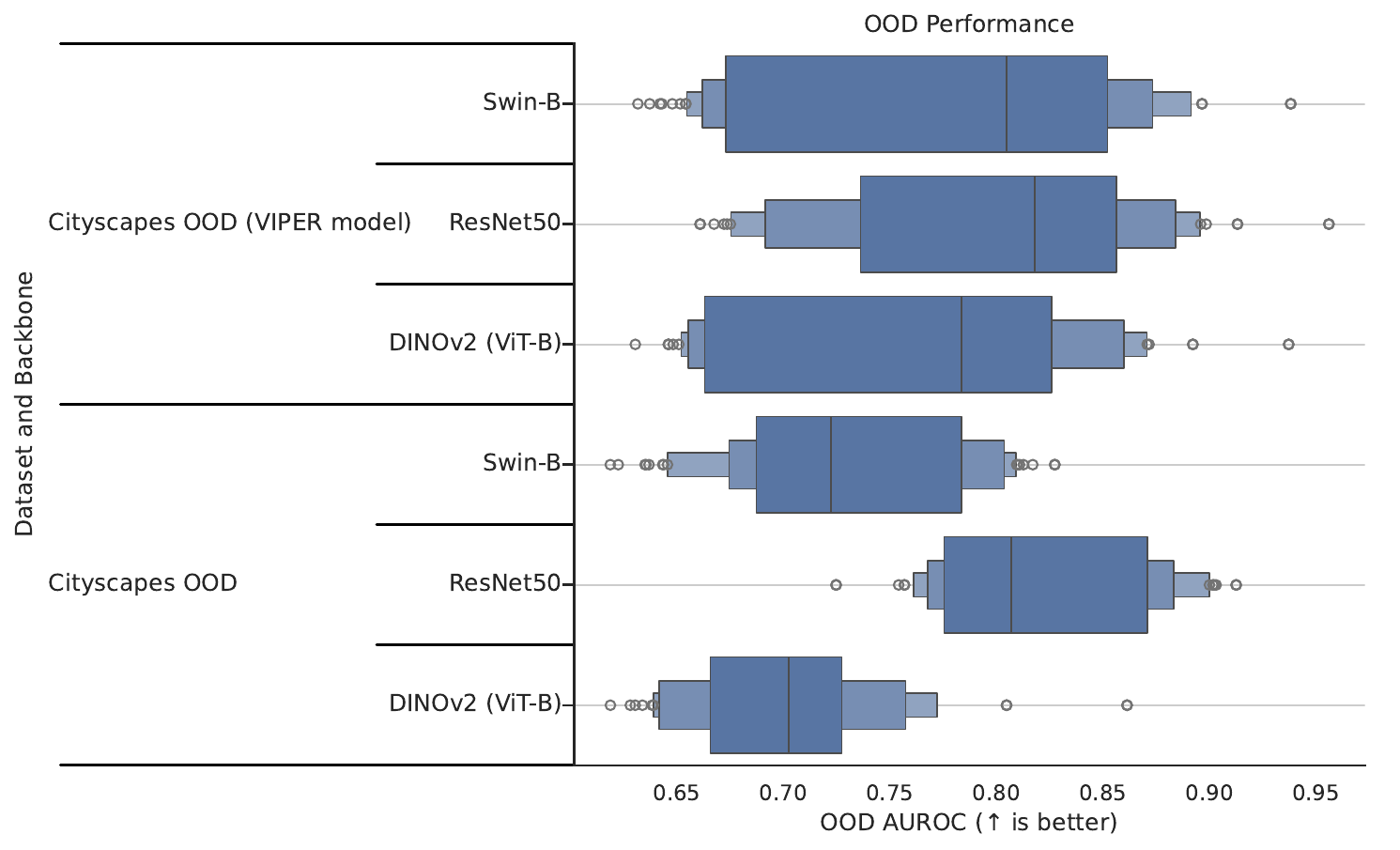}
    \caption{Out-of-distribution performance by dataset (\labelcref{param:dataset}) and backbone (\labelcref{param:backbone}) without normalization using the Area Under the Receiver Operating Characteristic metric. Experimental procedure follows \cref{fig:ood_results_combined}.}
    \label{fig:app:ood_auroc_datasec_backbone_raw}
\end{figure}

In \cref{fig:app:ood_auroc_datasec_backbone_raw}, we show the OOD performance without normalization on the Cityscapes dataset and with a distribution shift where we use the VIPER model instead. We see that un-normalized results are broadly in line with \cref{fig:ood_results_combined,fig:fd_calib_by_dataset_backbone_raw}, with a fairly broad spread of results and no clear performance leader, including for the foundation DINOv2 model which tends to trail other backbones in absolute terms.

\subsection{Downstream Tasks}
\label{subsec:app:downsteram_failure_detect_calib_ood_results}

\begin{figure*}
    \centering
    \includegraphics[width=1\linewidth]{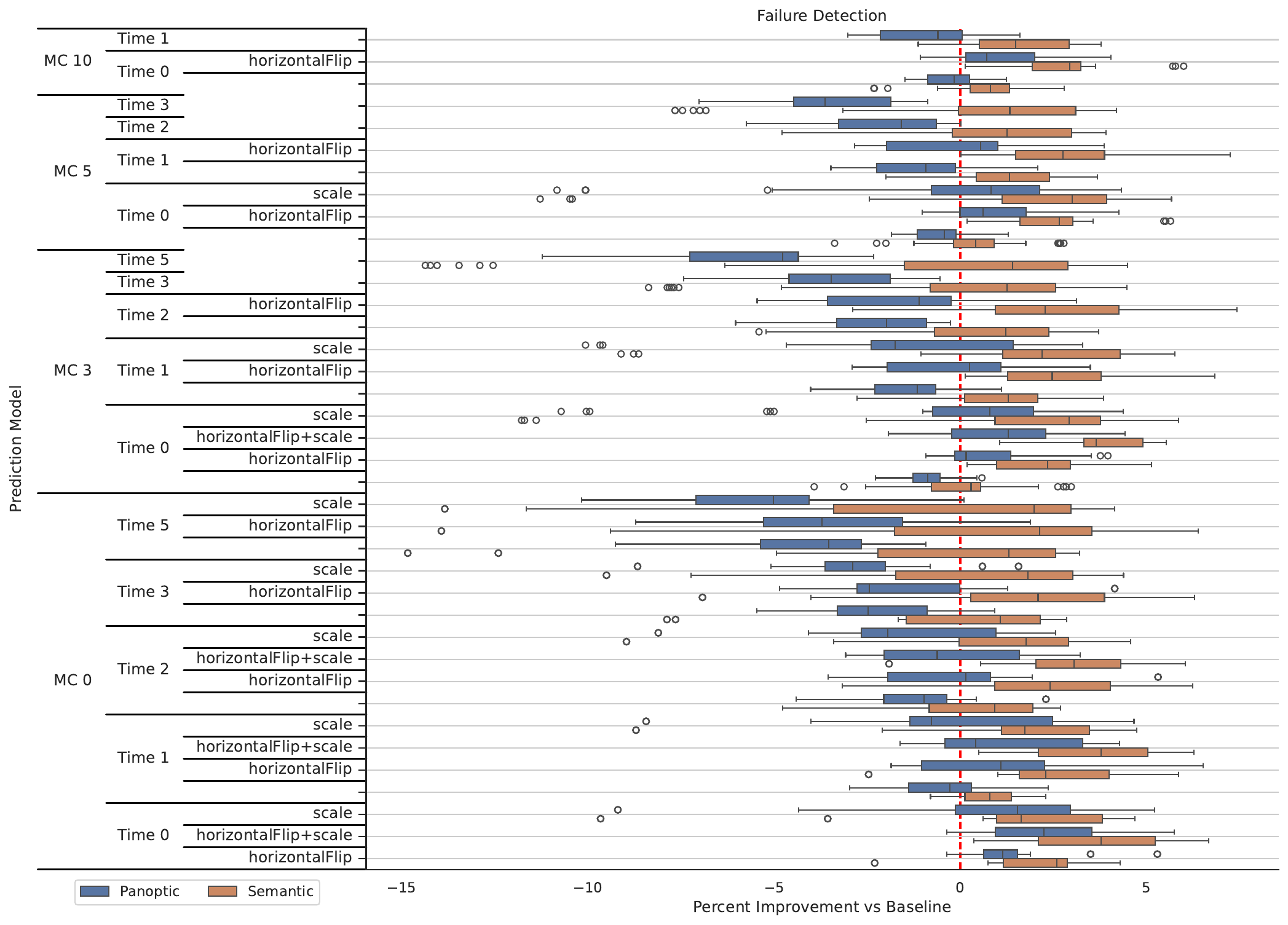}
    \caption{Results on the downstream task (\labelcref{param:downstream_task}) of failure detection, plotted by prediction model (\labelcref{param:pred_model}) and broken down into each individual tested configuration. Experiment parameters follow \cref{fig:iid_fd_calib_results_predmodel}, except a traditional box plot is used.}
    \label{fig:app:iid_failure_detect_detailed_predmodel}
\end{figure*}

\begin{figure*}
    \centering
    \includegraphics[width=1\linewidth]{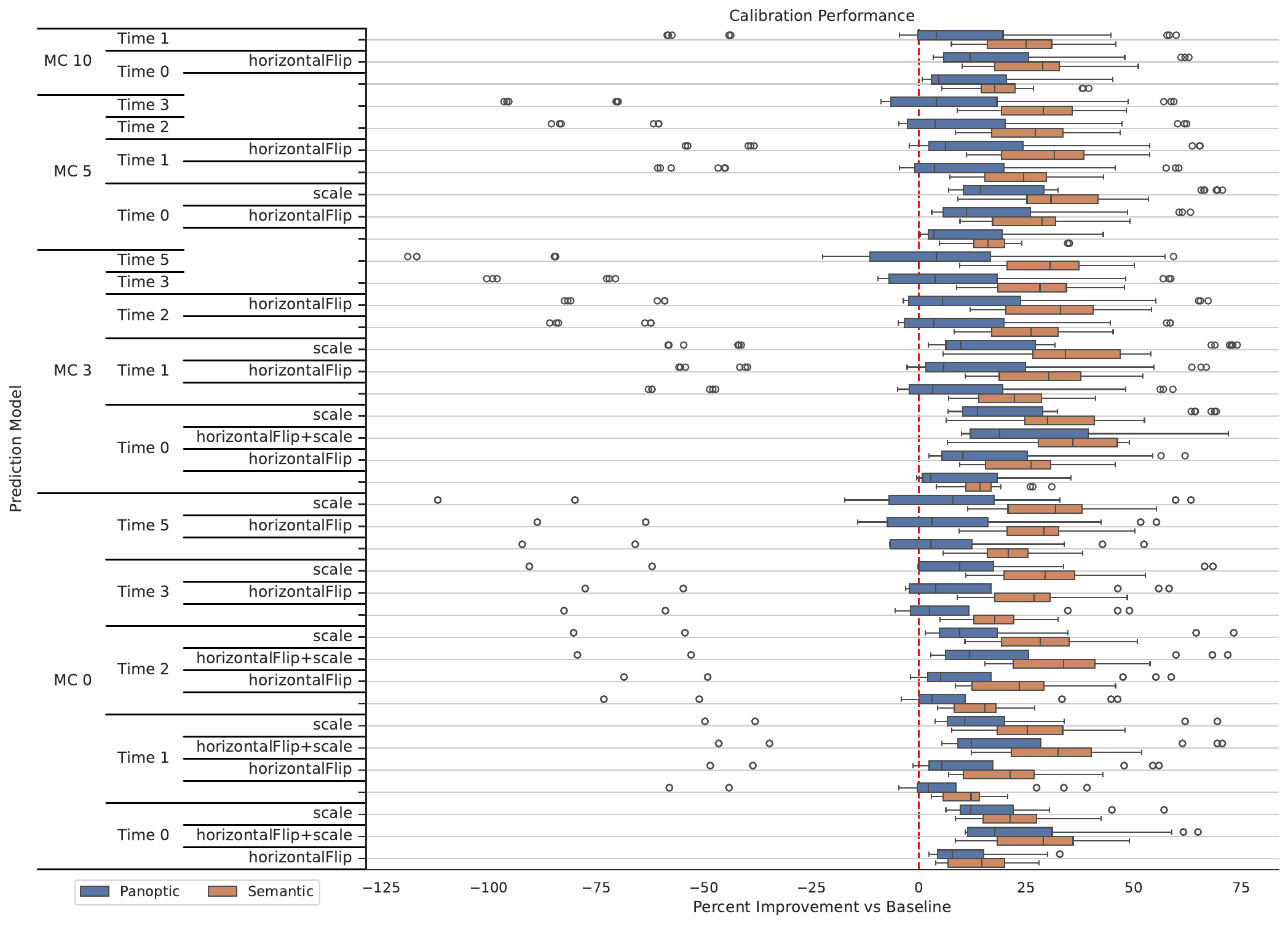}
    \caption{Results on the downstream task (\labelcref{param:downstream_task}) of calibration, plotted by prediction model (\labelcref{param:pred_model}) and broken down into each individual tested configuration. Experiment parameters follow \cref{fig:iid_fd_calib_results_predmodel}, except a traditional box plot is used.}
    \label{fig:app:iid_calib_detailed_predmodel}
\end{figure*}

In \cref{fig:app:iid_failure_detect_detailed_predmodel,fig:app:iid_calib_detailed_predmodel}, we show failure detection and calibration results respectively from the same experiments as \cref{fig:iid_fd_calib_results_predmodel} but broken down further into the individual prediction model configurations we evaluate. Most of the relevant observations are discussed in \cref{sec:results}, but we do note the fact that time series frames perform worse the more frames are added in the failure detection case. This applies to both semantic and panoptic segmentation, but is more pronounced in the latter case. Meanwhile, for panoptic calibration we do see some degradation with increased number of time series frames but it is fairly minimal.

\begin{figure*}
    \centering
    \includegraphics[width=1\linewidth]{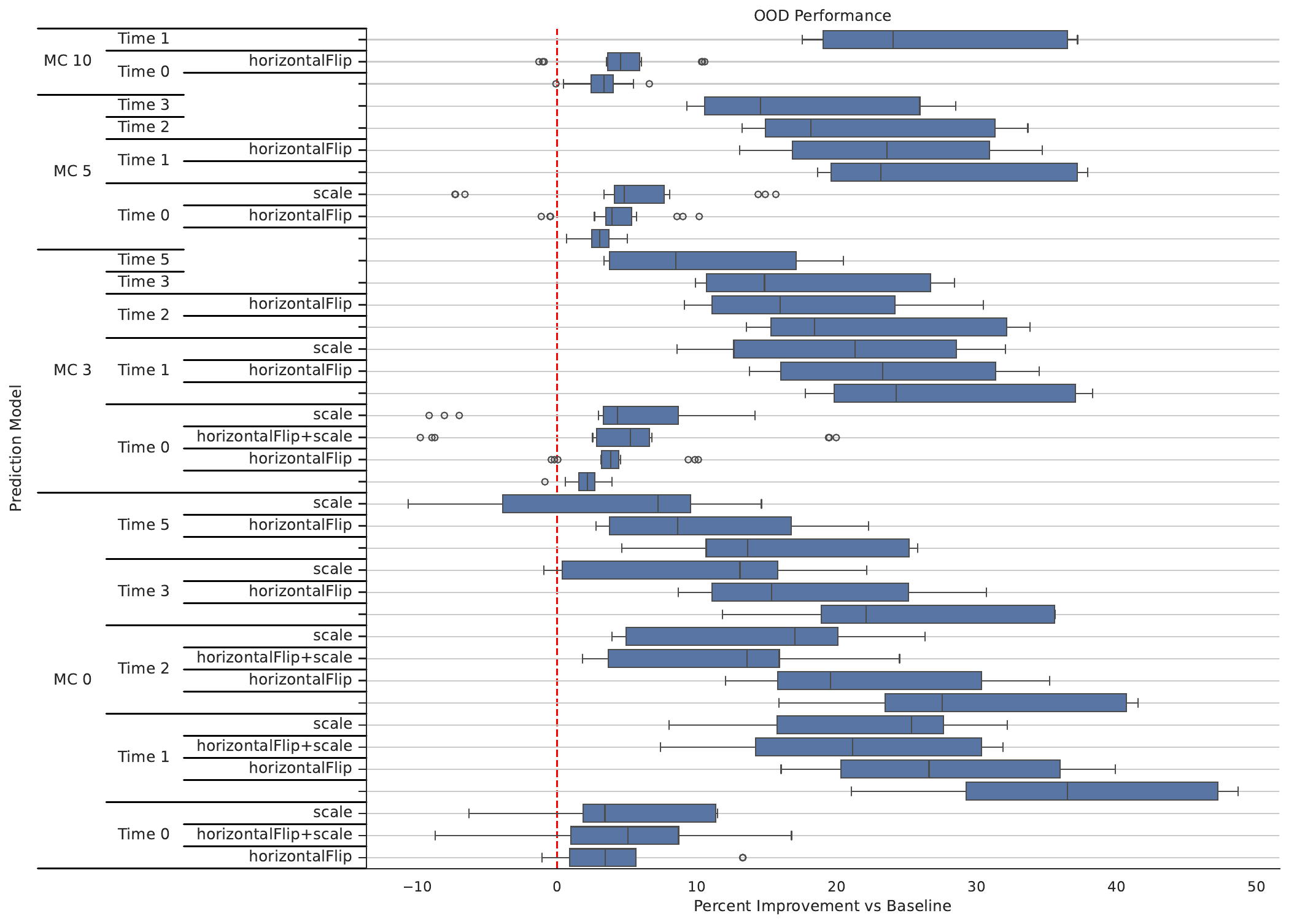}
    \caption{Results on the downstream task (\labelcref{param:downstream_task}) of out-of-distribution detection, plotted by prediction model (\labelcref{param:pred_model}) and broken down into each individual tested configuration. Experiment parameters follow \cref{fig:ood_results_predmodel}, except a traditional box plot is used.}
    \label{fig:app:ood_detailed_predmodel}
\end{figure*}

In \cref{fig:app:ood_detailed_predmodel}, we break down the results shown in \cref{fig:ood_results_predmodel} into the individual prediction model configurations. We see that despite the high performance of time series data as explained in \cref{sec:results}, performance degrades with more samples. Part of this may be due to noise from the frame alignment process (optical flow errors, occlusions) but it is also likely an artifact of the implementation. The out-of-distribution task expect higher uncertainty of OOD scenarios, but using more images without the OOD object from prior frames will decrease overall uncertainty explaining the worse results.

\begin{figure*}
    \centering
    \includegraphics[height=0.95\textheight]{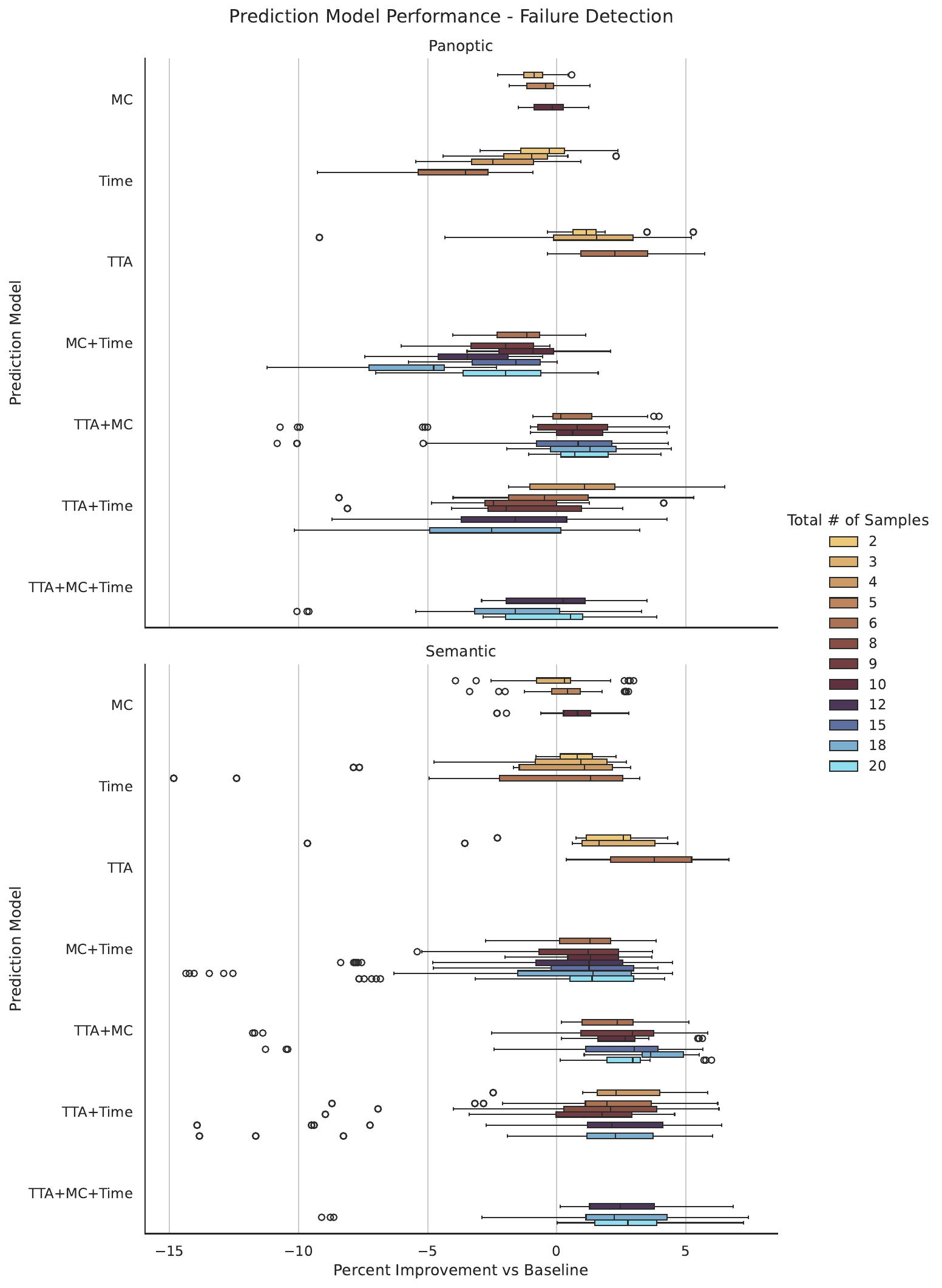}
    \caption{Results on the downstream tasks (\labelcref{param:downstream_task}) of failure detection, following the parameters of \cref{fig:iid_fd_calib_results_predmodel} except we break out the configurations by the number of samples and use a traditional box plot.}
    \label{fig:app:pred_model_comparison_withnumbersamples_fd}
\end{figure*}

\begin{figure*}
\centering
    \includegraphics[height=0.95\textheight]{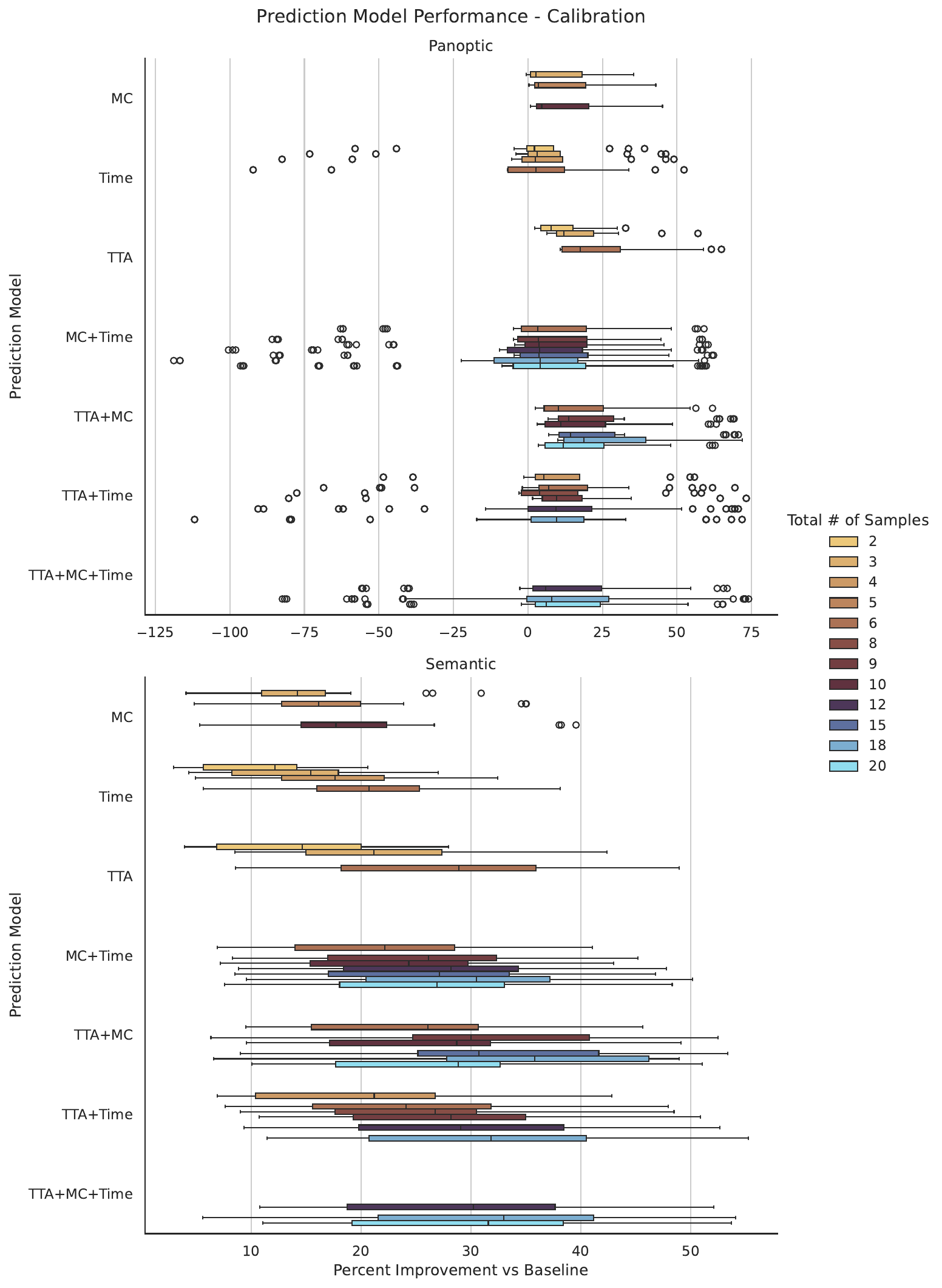}
    \caption{Results on the downstream tasks (\labelcref{param:downstream_task}) of calibration, following the parameters of \cref{fig:iid_fd_calib_results_predmodel} except we break out the configurations by the number of samples and use a traditional box plot.}
    \label{fig:app:pred_model_comparison_withnumbersamples_calib}
\end{figure*}

\begin{figure*}
    \centering
    \includegraphics[width=0.95\linewidth]{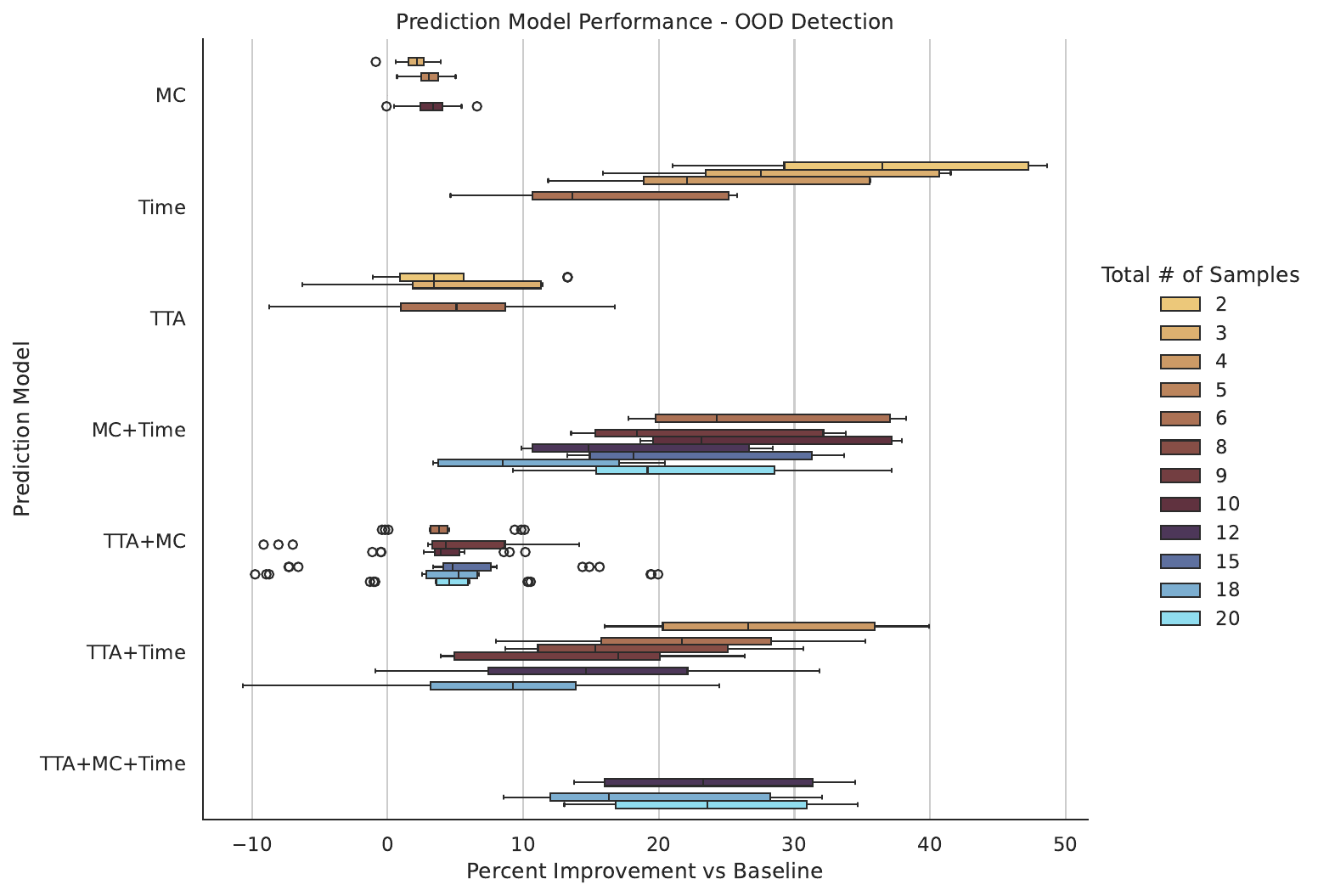}
    \caption{Results on the downstream tasks (\labelcref{param:downstream_task}) of out-of-distribution detection, following the parameters of \cref{fig:ood_results_predmodel} except we break out the configurations by the number of samples and use a traditional box plot.}
    \label{fig:app:pred_model_comparison_withnumbersamples_ood}
\end{figure*}
    
In \cref{fig:app:pred_model_comparison_withnumbersamples_fd,fig:app:pred_model_comparison_withnumbersamples_calib,fig:app:pred_model_comparison_withnumbersamples_ood} we extend the results shown in \cref{fig:iid_fd_calib_results_predmodel,fig:ood_results_predmodel} by additionally breaking the results down by the number of samples (see \cref{tab:app:pred_model_config_list}). We see the same trends previously discussed in this section, where using more time series frames is detrimental in some cases due to noise but less so in others.

Comparing across downstream tasks in \cref{fig:app:iid_failure_detect_detailed_predmodel,fig:app:iid_calib_detailed_predmodel,fig:app:ood_detailed_predmodel}, the relative performance differences are a result of the task specification and the underlying properties of the evaluation metrics. For instance, in \cref{fig:app:iid_failure_detect_detailed_predmodel} we see that adding more time series frames hurts performance, but in \cref{fig:app:iid_calib_detailed_predmodel} the effect exists but is minimal in comparison. This is not surprising, as the increased noise (previously discussed in this section) from adding more frames harms segmentation performance more than any increase in failure detection from a better uncertainty estimate. Meanwhile, calibration error is independent of segmentation error, although the noise from alignment errors as more frames are added still causes a (minimal) decrease.

\subsection{Sample Aggregation}
\label{subsec:app:sample_agg_results}
\begin{figure*}
    \centering
    \includegraphics[width=1\linewidth]{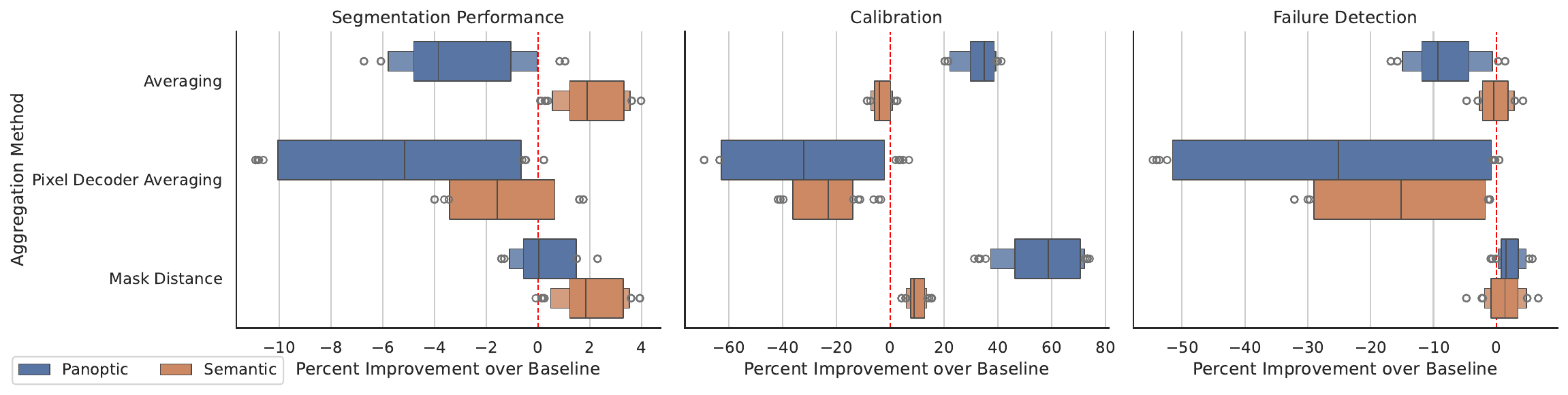}
    \caption{Comparison of different approaches for \textbf{Sample Aggregation} (\labelcref{param:sample_agg}) in terms of performance on segmentation and the downstream tasks (\labelcref{param:downstream_task}) of calibration and failure detection. As in \cref{fig:iid_fd_calib_results_combined}, letter-value plots are used. Dataset and backbone (\labelcref{param:dataset,param:backbone}) are fixed to \textit{Cityscapes} and \textit{ResNet50} respectively. The baseline is a deterministic model with no sampling and thus sample aggregation (\labelcref{param:sample_agg}) is not needed. Prediction model combinations (\labelcref{param:pred_model}) are swept through, while \labelcref{param:uncert_measure,param:pixel_agg} are optimized over.}
    \label{fig:app:agg_compare}
\end{figure*}
In \cref{fig:app:agg_compare}, we see that our choice to use \textit{Mask Distance} in all other experiments is justified, as it leads across the board on segmentation performance and the downstream tasks of calibration and failure detection. This applies for both panoptic and semantic segmentation. Please see \cref{sec:study_approach,sec:app:sample_agg_extended} for details.

\subsection{Uncertainty Measures}
\label{subsec:app:uncert_measure_results}

\begin{figure}
    \centering
    \begin{subfigure}{1\linewidth}
        \centering
        \includegraphics[width=\linewidth]{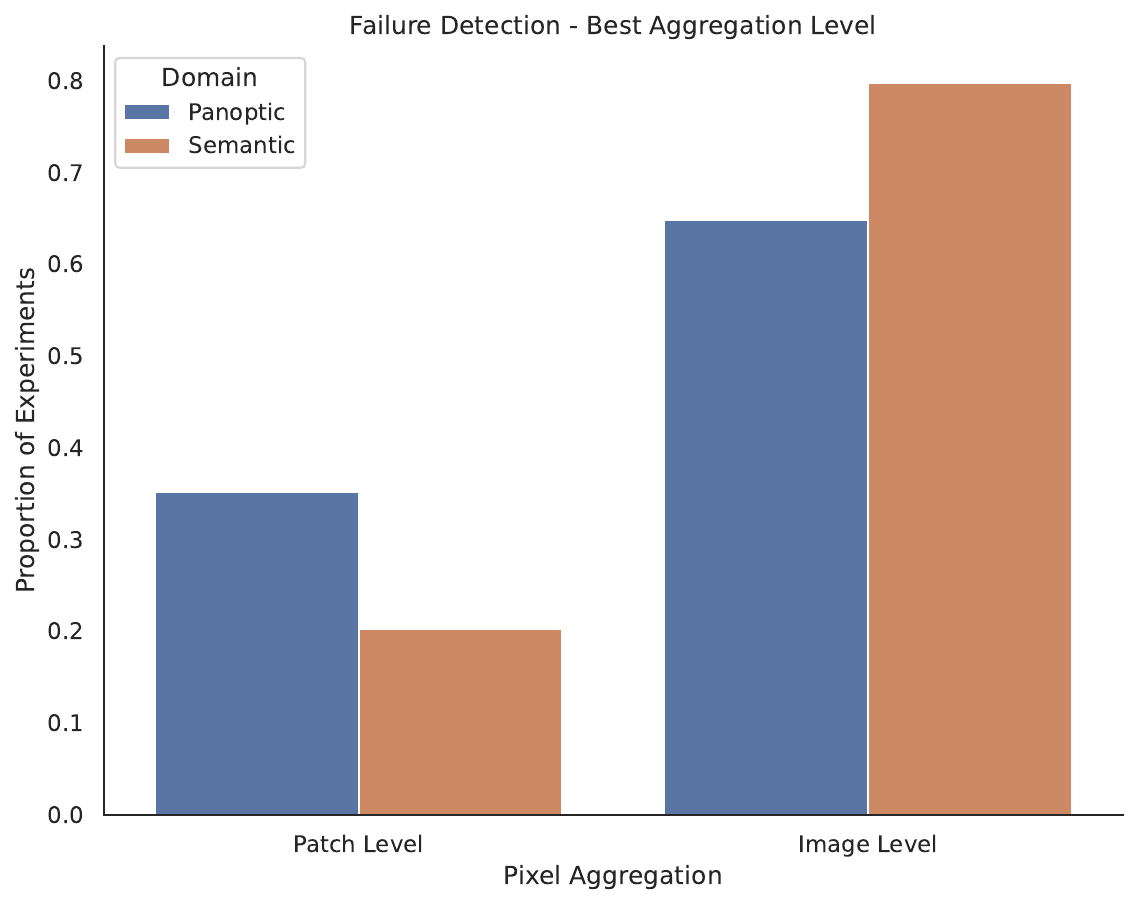}
        \caption{}
        \label{fig:app:fd_agglevel_byuncert}
    \end{subfigure}
    \begin{subfigure}{1\linewidth}
        \centering
        \includegraphics[width=\linewidth]{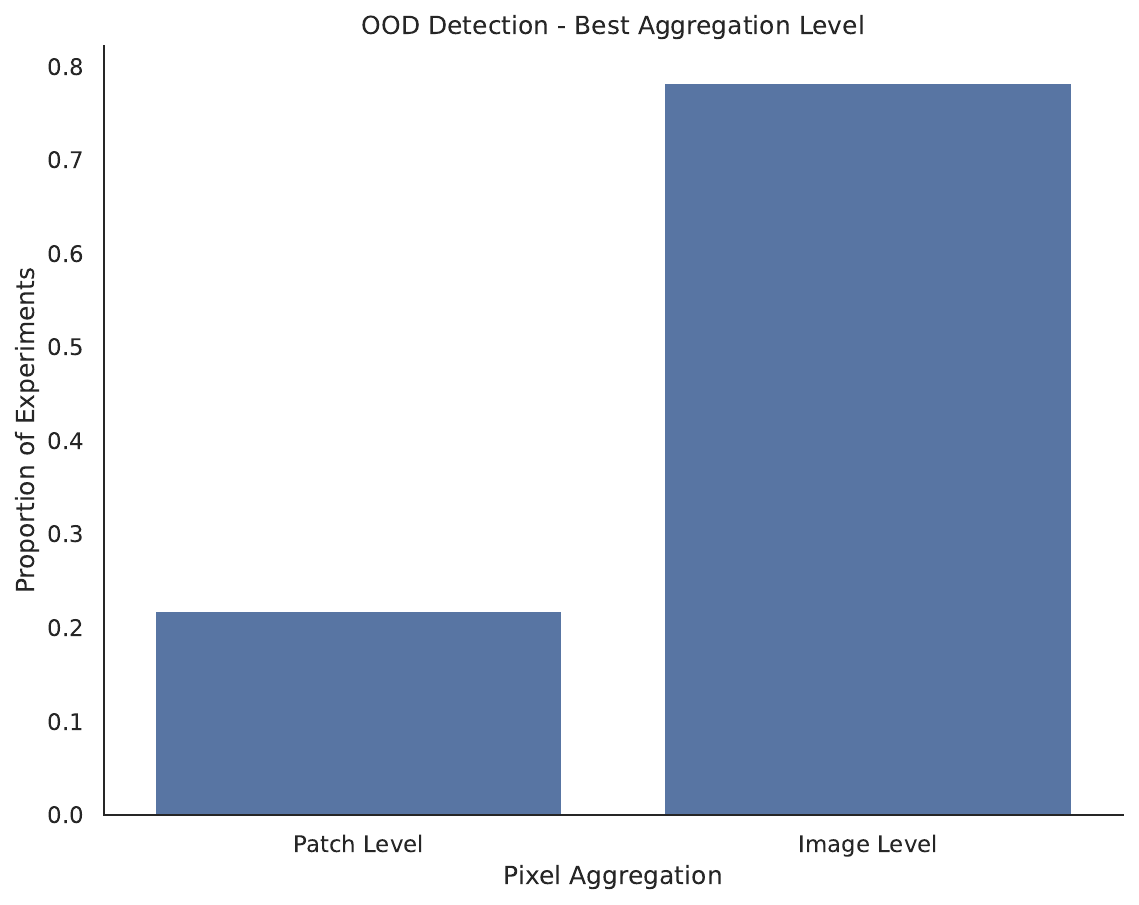}
        \caption{}
        \label{fig:app:ood_agglevel_byuncert}
    \end{subfigure}
    \caption{Pixel Aggregation (\labelcref{param:pixel_agg}) comparison, presented as a proportion of experiments where one approach was better than the other. Experiments used are the same as \cref{fig:iid_fd_calib_results_combined,fig:ood_results_combined} for (a) failure detection and (b) out-of-distribution detection respectively. Comparison is made after fixing dataset, backbone, prediction model configuration and uncertainty measure. The seed variable is removed via averaging.}
    \label{fig:app:agg_level_comparison}
\end{figure}

\begin{figure}
    \centering
    \begin{subfigure}{0.85\linewidth}
        \centering
        \includegraphics[width=\linewidth]{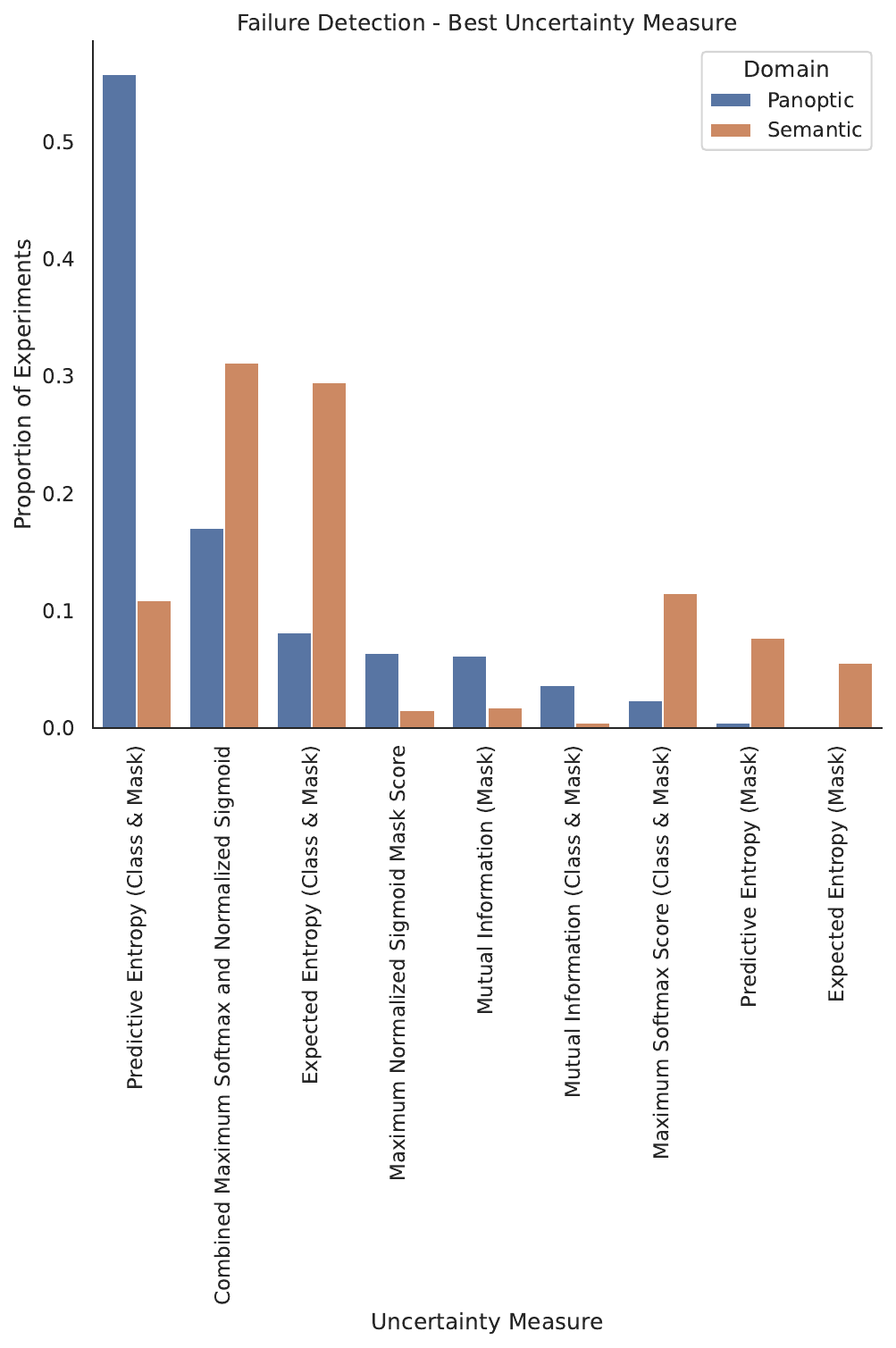}
        \caption{}
        \label{fig:app:fd_uncerttype_bypredmodelcombo}
    \end{subfigure}
    \begin{subfigure}{0.85\linewidth}
        \centering
        \includegraphics[width=\linewidth]{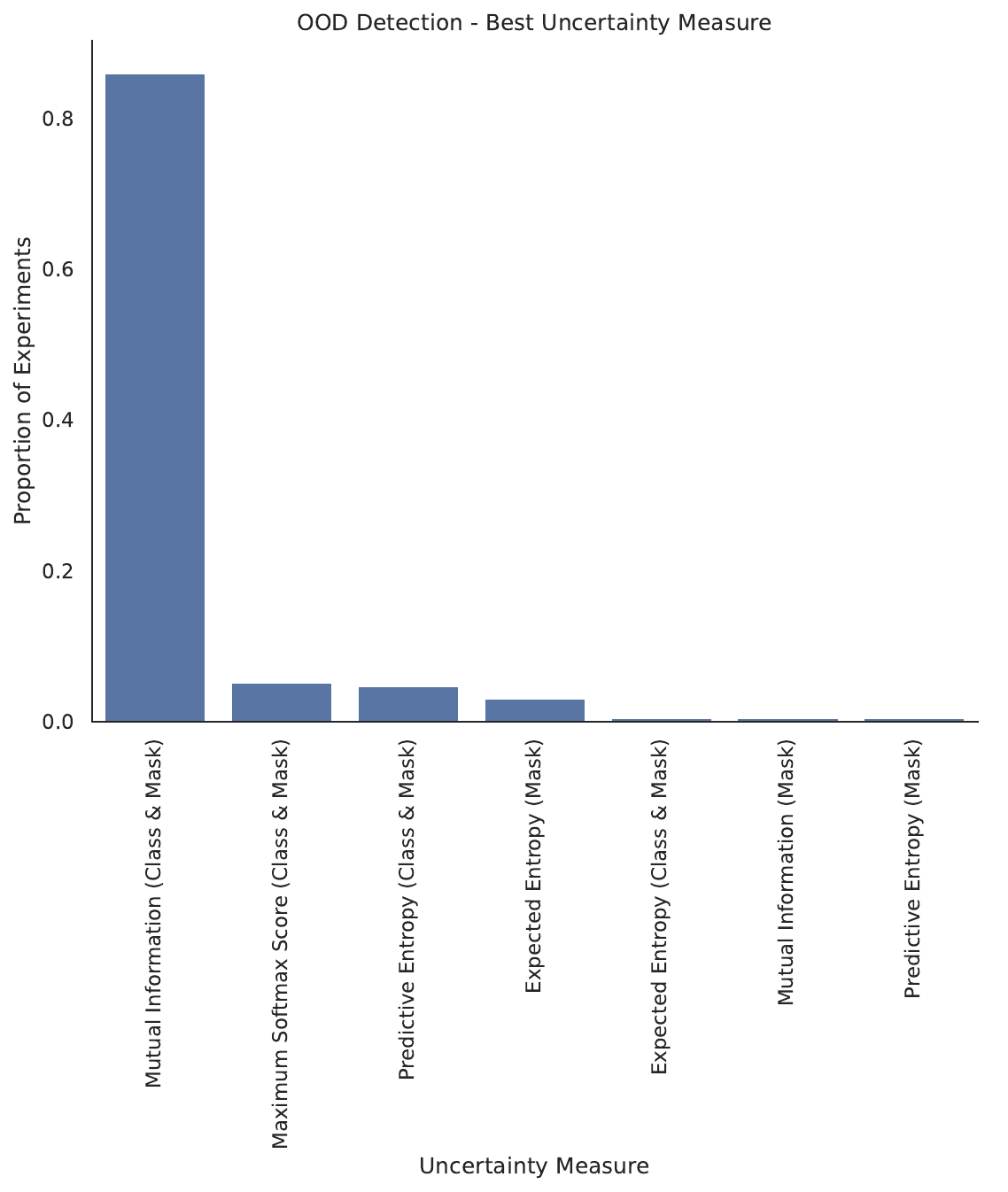}
        \caption{}
        \label{fig:app:ood_uncerttype_bypredmodelcombo}
    \end{subfigure}
    \caption{Uncertainty measure (\labelcref{param:uncert_measure}) comparison, presented as a proportion of experiments where one approach was better than the other. Experiments used are the same as \cref{fig:iid_fd_calib_results_combined,fig:ood_results_combined} for (a) failure detection and (b) out-of-distribution detection respectively. Comparison is made after fixing dataset, backbone, and prediction model configuration. The seed variable is removed via averaging.}
    \label{fig:app:uncert_level_comparison}
\end{figure}

In \cref{fig:app:agg_level_comparison,fig:app:uncert_level_comparison}, using the data from our main experiments as presented in \cref{fig:iid_fd_calib_results_combined,fig:ood_results_combined} we compare different uncertainty measures and pixel aggregation approaches. Note that this is a specific case where we wished to examine if one uncertainty measure or pixel aggregation was dominant; in all other experiments, we optimize over uncertainty measures and pixel aggregation approaches so as to avoid introducing any bias as discussed in \cref{sec:app:uncert_measures}.

What the results show is broadly in line with \cite{kahl_values_2024} in that the best possible approach to take is very conditional, with a joint optimization of all variables being necessary when choosing the best setting for each variable. Were we forced to make an overall choice, image level aggregation seems the best, but there is no clear winner for uncertainty measures. In failure detection, predictive entropy leads in the panoptic domain, while the ad-hoc measure that combines the softmax and normalized mask score is roughly tied with expected entropy in the semantic domain. Meanwhile, in out-of-distribution detection mutual information is a clear leader.

\subsection{Domain Changes}
In \cref{fig:iid_fd_calib_results_combined,fig:app:iid_failure_detect_detailed_predmodel,fig:app:iid_calib_detailed_predmodel}, we compare results across both the panoptic and semantic segmentation domains. In the main paper, we note that panoptic performance is usually worse than semantic on a given metric. We believe the primary reason for this is the problem specification, where semantic segmentation requires a class label and panoptic segmentation requires the both a class and instance label for each pixel. The metrics used for each (discussed in \cref{subsec:metrics}) are also different, with panoptic segmentation requiring individual objects to achieve a minimum overlap with ground truth. Meanwhile, semantic segmentation only evaluates pixel-level correctness for each class. This makes the panoptic domain more sensitive to pixel level errors than semantic segmentation, such as around object boundaries. We see this especially when we add time series frames (discussed in \cref{subsec:app:time_series_details}), with the relative performance dropping as more frames are added. This is expected, as alignment errors will accumulate, especially at object boundaries. This is also the case with ground truth optical flow data due to its deficiencies, as discussed in \cref{subsec:app:optical_flow}.

\section{Computational Cost}
\label{sec:app:computational_cost}
Training models is fairly resource intensive, and our requirements follow those of \cite{cheng_masked-attention_2022} where a multi-GPU machine with sufficiently large GPUs to fit larger batch sizes is ideal. In our work, we primarily used nodes with 4 Nvidia V100-SMX2-32GB.

As our work focuses on uncertainty estimation, inference performance is an issue, particularly with ensembles requiring multiple samples. Our implementation of ensembles is straightforward but restricted to a single GPU. Different experiments, however, such as with a different seed or dataset can of course run concurrently on other GPUs. Depending on the size of the dataset validation set and prediction model, a pass through can take anywhere from a few minutes for a deterministic baseline to several hours for prediction model configurations with many samples. At inference time, with Mask Distance sample aggregation all experiments in the configurations presented can run on a single 12GB GPU. To evaluate all the approaches we show in this study, however, multiple GPUs are required to run experiments in parallel so as to have them finish in a reasonable amount of time. As implemented, we also save the network output to disk (segmentation results, uncertainty estimates) for later processing which does not require a GPU (\eg calculating the AURC metrics for failure detection) so as to allow the GPUs to continue with other experiments but this requires multiple terabytes of storage space, ideally over 100 TB.

One of the key limitations of any ensemble approach is the time required to compute samples, and our approach is no different. To demonstrate, we evaluate different prediction model combinations on the Cityscapes dataset with the ResNet50 backone and record the inference time required per image to generate the samples, aggregate them and calculate uncertainty measures. Time to load data from disk is excluded. Evaluation is done on a single machine with a Nvidia RTX 4090 with 24B of VRAM. All prediction model combinations applicable in \cref{tab:app:pred_model_config_list} are tested, with the exception of the following with the Averaging sample aggregation approach due to VRAM limits:
\begin{itemize}
    \item 3 MC samples, 1 previous time frame, horizontal flips for TTA
    \item 3 MC samples, 1 previous frame, scale transforms for TTA
    \item No MC samples, 1 previous frame, horizontal flips and scale transforms for TTA
    \item No MC samples, 2 previous frames, horizontal flips and scale transforms for TTA
\end{itemize}

\begin{figure}
    \centering
    \includegraphics[width=1\linewidth]{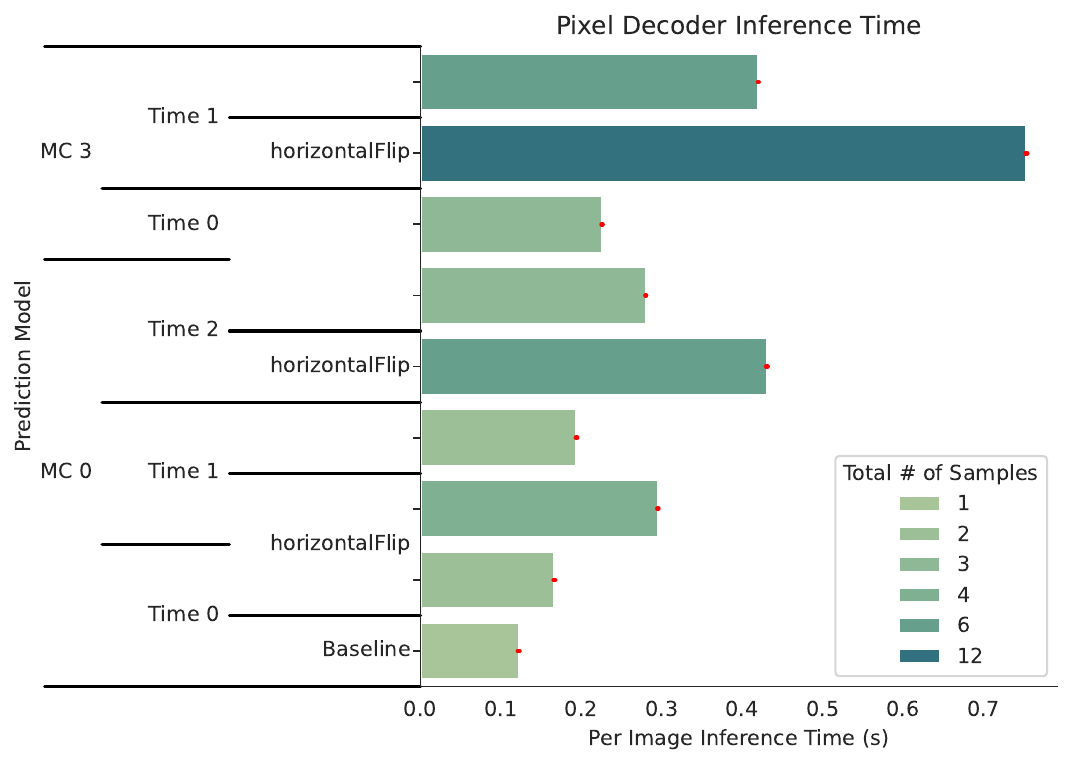}
    \caption{Inference time comparison for the Pixel Decoder approach to sample aggregation (\labelcref{param:sample_agg}), executed on a Nvidia RTX 4090. Results are averaged over 3 runs; error bars in red show the $95\%$ confidence interval, generated via bootstrapping.}
    \label{fig:app:pixel_dec_time_comparison}
\end{figure}

\begin{figure}
    \centering
    \includegraphics[width=1\linewidth]{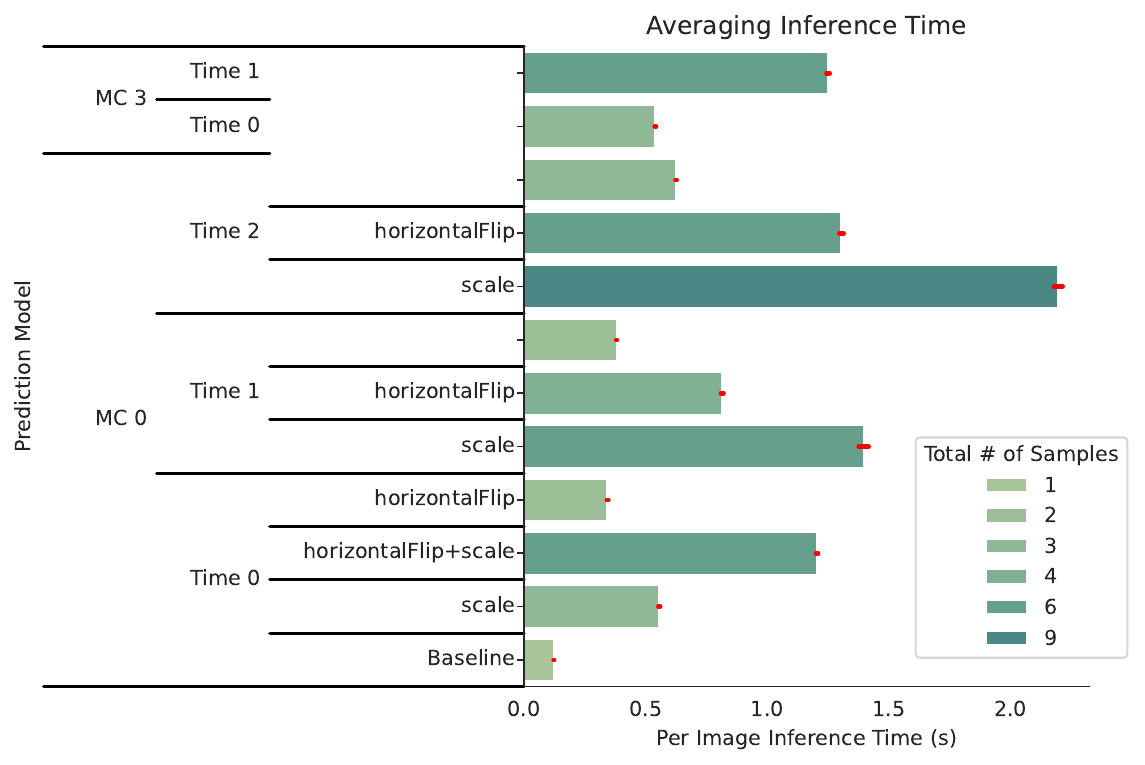}
    \caption{Inference time comparison for the Averaging approach \cite{smith_uncertainty_2024} to sample aggregation (\labelcref{param:sample_agg}), executed on a Nvidia RTX 4090. Results are averaged over 3 runs; error bars in red show the $95\%$ confidence interval, generated via bootstrapping.}
    \label{fig:app:avg_time_comparison}
\end{figure}

\begin{figure*}[p]
    \centering
    \includegraphics[width=1\linewidth]{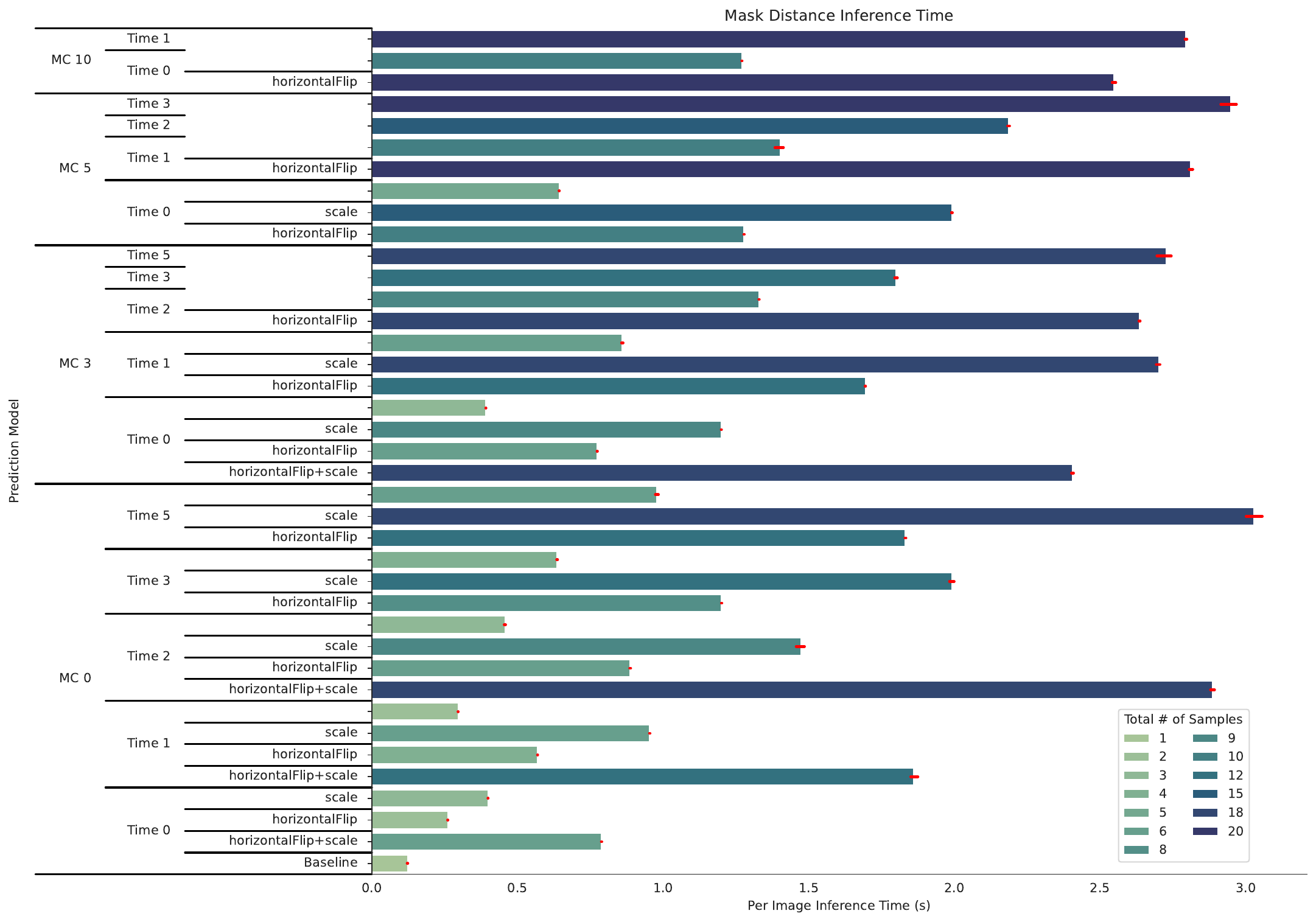}
    \caption{Inference time comparison for the Mask Distance approach to sample aggregation (\labelcref{param:sample_agg}), executed on a Nvidia RTX 4090. Results are averaged over 3 runs; error bars in red show the $95\%$ confidence interval, generated via bootstrapping.}
    \label{fig:app:mask_dist_time_comparison}
\end{figure*}

In \cref{fig:app:mask_dist_time_comparison,fig:app:pixel_dec_time_comparison,fig:app:avg_time_comparison} we compare the inference time required for the Mask Distance, Pixel Decoder and Averaging approaches respectively. We see that the number of samples is the most significant factor, with the time required being roughly proportional to the number of samples. We also see that the Averaging approach from \cite{smith_uncertainty_2024} is the slowest, followed by our Mask Distance approach. The fastest approach is the Pixel Decoder which we introduce in \cref{subsec:app:pixel_decoder}, with the required time for similar prediction model configurations being a fraction of the two other approaches. As with all other variables we discuss, a practitioner deploying a model in the real world will need to make a choice as to which sample aggregation may suit them best in the trade-off between uncertainty estimation quality and speed.

{
    \small
    \renewcommand{\refname}{Supplementary Material References}
    \bibliographystyle{ieeenat_fullname}
    \bibliography{main}

\begin{thebibliography}{80}
\providecommand{\natexlab}[1]{#1}
\providecommand{\url}[1]{\texttt{#1}}
\expandafter\ifx\csname urlstyle\endcsname\relax
  \providecommand{\doi}[1]{doi: #1}\else
  \providecommand{\doi}{doi: \begingroup \urlstyle{rm}\Url}\fi

\bibitem[Arad and Rosset(2025)]{arad_improving_2025}
Alon Arad and Saharon Rosset.
\newblock Improving {Multi}-{Class} {Calibration} through
  {Normalization}-{Aware} {Isotonic} {Techniques}.
\newblock In \emph{Forty-second {International} {Conference} on {Machine}
  {Learning}}, 2025.

\bibitem[Asgharnezhad et~al.(2025)Asgharnezhad, Tabarisaadi, Khosravi,
  Alizadehsani, and Acharya]{asgharnezhad_uncertainty-aware_2025}
Hamzeh Asgharnezhad, Pegah Tabarisaadi, Abbas Khosravi, Roohallah Alizadehsani,
  and U~Rajendra Acharya.
\newblock Uncertainty-{Aware} {Deep} {Learning} for {Automated} {Skin} {Cancer}
  {Classification}: {A} {Comprehensive} {Evaluation}.
\newblock \emph{arXiv preprint arXiv:2506.10302}, 2025.

\bibitem[Ayhan and Berens(2018)]{ayhan_test-time_2018}
Murat~Seckin Ayhan and Philipp Berens.
\newblock Test-time {Data} {Augmentation} for {Estimation} of {Heteroscedastic}
  {Aleatoric} {Uncertainty} in {Deep} {Neural} {Networks}.
\newblock In \emph{Medical {Imaging} with {Deep} {Learning}}, 2018.

\bibitem[Benigmim et~al.(2025)Benigmim, Fahes, Vu, Bursuc, and
  de~Charette]{benigmim_floss_2025}
Yasser Benigmim, Mohammad Fahes, Tuan-Hung Vu, Andrei Bursuc, and Raoul de
  Charette.
\newblock {FLOSS}: {Free} {Lunch} in {Open}-vocabulary {Semantic}
  {Segmentation}.
\newblock In \emph{Proceedings of the {IEEE}/{CVF} {International} {Conference}
  on {Computer} {Vision} ({ICCV})}, pages 21471--21481, 2025.

\bibitem[Blundell et~al.(2015)Blundell, Cornebise, Kavukcuoglu, and
  Wierstra]{blundell_weight_2015}
Charles Blundell, Julien Cornebise, Koray Kavukcuoglu, and Daan Wierstra.
\newblock Weight {Uncertainty} in {Neural} {Network}.
\newblock In \emph{Proceedings of the 32nd {International} {Conference} on
  {Machine} {Learning}}, pages 1613--1622, Lille, France, 2015. PMLR.

\bibitem[Bui and Liu(2024)]{bui_density-softmax_2024}
Ha~Manh Bui and Anqi Liu.
\newblock Density-{Softmax}: {Efficient} {Test}-time {Model} for {Uncertainty}
  {Estimation} and {Robustness} under {Distribution} {Shifts}.
\newblock In \emph{Proceedings of the 41st {International} {Conference} on
  {Machine} {Learning}}, pages 4822--4853. PMLR, 2024.

\bibitem[Bursuc(2023)]{bursuc_many_2023}
Andrei Bursuc.
\newblock The {Many} {Faces} of {Reliability}: {Uncertainy} {Estimation} and
  {Ensemble} {Approaches}, 2023.

\bibitem[Charpentier et~al.(2022)Charpentier, Borchert, Zügner, Geisler, and
  Günnemann]{charpentier_natural_2022}
Bertrand Charpentier, Oliver Borchert, Daniel Zügner, Simon Geisler, and
  Stephan Günnemann.
\newblock Natural {Posterior} {Network}: {Deep} {Bayesian} {Predictive}
  {Uncertainty} for {Exponential} {Family} {Distributions}.
\newblock In \emph{International {Conference} on {Learning} {Representations}},
  2022.

\bibitem[Chen et~al.(2025)Chen, Han, Zhang, Luo, Wu, Cai, and
  Su]{chen_stronger_2025}
Siyu Chen, Ting Han, Changshe Zhang, Xin Luo, Meiliu Wu, Guorong Cai, and Jinhe
  Su.
\newblock Stronger, {Steadier} \& {Superior}: {Geometric} {Consistency} in
  {Depth} {VFM} {Forges} {Domain} {Generalized} {Semantic} {Segmentation}.
\newblock In \emph{Proceedings of the {IEEE}/{CVF} {International} {Conference}
  on {Computer} {Vision} ({ICCV})}, pages 8285--8295, 2025.

\bibitem[Chen et~al.(2023)Chen, Li, Saxena, Hinton, and
  Fleet]{chen_generalist_2023}
Ting Chen, Lala Li, Saurabh Saxena, Geoffrey Hinton, and David~J. Fleet.
\newblock A {Generalist} {Framework} for {Panoptic} {Segmentation} of {Images}
  and {Videos}.
\newblock In \emph{Proceedings of the {IEEE}/{CVF} {International} {Conference}
  on {Computer} {Vision} ({ICCV})}, pages 909--919, 2023.

\bibitem[Cheng et~al.(2020)Cheng, Collins, Zhu, Liu, Huang, Adam, and
  Chen]{cheng_panoptic-deeplab_2020}
Bowen Cheng, Maxwell~D. Collins, Yukun Zhu, Ting Liu, Thomas~S. Huang, Hartwig
  Adam, and Liang-Chieh Chen.
\newblock Panoptic-{DeepLab}: {A} {Simple}, {Strong}, and {Fast} {Baseline} for
  {Bottom}-{Up} {Panoptic} {Segmentation}.
\newblock In \emph{Proceedings of the {IEEE}/{CVF} {Conference} on {Computer}
  {Vision} and {Pattern} {Recognition} ({CVPR})}, 2020.

\bibitem[Cheng et~al.(2022)Cheng, Misra, Schwing, Kirillov, and
  Girdhar]{cheng_masked-attention_2022}
Bowen Cheng, Ishan Misra, Alexander~G. Schwing, Alexander Kirillov, and Rohit
  Girdhar.
\newblock Masked-attention {Mask} {Transformer} for {Universal} {Image}
  {Segmentation}.
\newblock In \emph{2022 {IEEE}/{CVF} {Conference} on {Computer} {Vision} and
  {Pattern} {Recognition} ({CVPR})}, pages 1280--1289, 2022.

\bibitem[Cheng and Vasconcelos(2024)]{cheng_towards_2024}
Jiacheng Cheng and Nuno Vasconcelos.
\newblock Towards {Calibrated} {Multi}-label {Deep} {Neural} {Networks}.
\newblock In \emph{Proceedings of the {IEEE}/{CVF} {Conference} on {Computer}
  {Vision} and {Pattern} {Recognition} ({CVPR})}, pages 27589--27599, 2024.

\bibitem[Cordts et~al.(2016)Cordts, Omran, Ramos, Rehfeld, Enzweiler, Benenson,
  Franke, Roth, and Schiele]{cordts_cityscapes_2016}
Marius Cordts, Mohamed Omran, Sebastian Ramos, Timo Rehfeld, Markus Enzweiler,
  Rodrigo Benenson, Uwe Franke, Stefan Roth, and Bernt Schiele.
\newblock The {Cityscapes} {Dataset} for {Semantic} {Urban} {Scene}
  {Understanding}.
\newblock In \emph{Proc. of the {IEEE} {Conference} on {Computer} {Vision} and
  {Pattern} {Recognition} ({CVPR})}, 2016.

\bibitem[Deery et~al.(2023)Deery, Lee, and Waslander]{deery_propandl_2023}
Jacob Deery, Chang~Won Lee, and Steven~L. Waslander.
\newblock {ProPanDL}: {A} {Modular} {Architecture} for {Uncertainty}-{Aware}
  {Panoptic} {Segmentation}.
\newblock In \emph{2023 20th {Conference} on {Robots} and {Vision} ({CRV})},
  pages 137--144, 2023.

\bibitem[Dosovitskiy et~al.(2021)Dosovitskiy, Beyer, Kolesnikov, Weissenborn,
  Zhai, Unterthiner, Dehghani, Minderer, Heigold, Gelly, Uszkoreit, and
  Houlsby]{dosovitskiy_image_2021}
Alexey Dosovitskiy, Lucas Beyer, Alexander Kolesnikov, Dirk Weissenborn,
  Xiaohua Zhai, Thomas Unterthiner, Mostafa Dehghani, Matthias Minderer, Georg
  Heigold, Sylvain Gelly, Jakob Uszkoreit, and Neil Houlsby.
\newblock An {Image} is {Worth} 16x16 {Words}: {Transformers} for {Image}
  {Recognition} at {Scale}.
\newblock In \emph{International {Conference} on {Learning} {Representations}},
  2021.

\bibitem[Everingham et~al.(2015)Everingham, Eslami, Van~Gool, Williams, Winn,
  and Zisserman]{everingham_pascal_2015}
Mark Everingham, S.~M.~Ali Eslami, Luc Van~Gool, Christopher K.~I. Williams,
  John Winn, and Andrew Zisserman.
\newblock The {Pascal} {Visual} {Object} {Classes} {Challenge}: {A}
  {Retrospective}.
\newblock \emph{International Journal of Computer Vision}, 111\penalty0
  (1):\penalty0 98--136, 2015.

\bibitem[Gal and Ghahramani(2016)]{gal_dropout_2016}
Yarin Gal and Zoubin Ghahramani.
\newblock Dropout as a {Bayesian} {Approximation}: {Representing} {Model}
  {Uncertainty} in {Deep} {Learning}.
\newblock In \emph{Proceedings of {The} 33rd {International} {Conference} on
  {Machine} {Learning}}, pages 1050--1059, New York, New York, USA, 2016. PMLR.

\bibitem[Gallos and Ferrie(2019)]{gallos_active_2019}
Dimitrios Gallos and Frank Ferrie.
\newblock Active {Vision} in the {Era} of {Convolutional} {Neural} {Networks}.
\newblock In \emph{2019 16th {Conference} on {Computer} and {Robot} {Vision}
  ({CRV})}, pages 81--88, 2019.

\bibitem[Ge et~al.(2025)Ge, Xie, Xie, Li, Liu, Zhang, Tian, and
  Zhang]{ge_clip-adapted_2025}
Jiannan Ge, Lingxi Xie, Hongtao Xie, Pandeng Li, Sun-Ao Liu, Xiaopeng Zhang, Qi
  Tian, and Yongdong Zhang.
\newblock {CLIP}-{Adapted} {Region}-to-{Text} {Learning} for {Generative}
  {Open}-{Vocabulary} {Semantic} {Segmentation}.
\newblock In \emph{Proceedings of the {IEEE}/{CVF} {International} {Conference}
  on {Computer} {Vision} ({ICCV})}, pages 24034--24044, 2025.

\bibitem[Guo et~al.(2017)Guo, Pleiss, Sun, and
  Weinberger]{guo_calibration_2017}
Chuan Guo, Geoff Pleiss, Yu Sun, and Kilian~Q. Weinberger.
\newblock On {Calibration} of {Modern} {Neural} {Networks}.
\newblock In \emph{Proceedings of the 34th {International} {Conference} on
  {Machine} {Learning}}, pages 1321--1330. PMLR, 2017.

\bibitem[Guo et~al.(2023)Guo, Lu, Liu, Cheng, and Hu]{guo_visual_2023}
Meng-Hao Guo, Cheng-Ze Lu, Zheng-Ning Liu, Ming-Ming Cheng, and Shi-Min Hu.
\newblock Visual attention network.
\newblock \emph{Computational Visual Media}, 9\penalty0 (4):\penalty0 733--752,
  2023.

\bibitem[He et~al.(2016)He, Zhang, Ren, and Sun]{he_resnet_2016}
Kaiming He, Xiangyu Zhang, Shaoqing Ren, and Jian Sun.
\newblock Deep {Residual} {Learning} for {Image} {Recognition}.
\newblock In \emph{Proceedings of the {IEEE} {Conference} on {Computer}
  {Vision} and {Pattern} {Recognition} ({CVPR})}, 2016.

\bibitem[Heidecker et~al.(2021)Heidecker, Hannan, Bieshaar, and
  Sick]{heidecker_towards_2021}
Florian Heidecker, Abdul Hannan, Maarten Bieshaar, and Bernhard Sick.
\newblock Towards {Corner} {Case} {Detection} by {Modeling} the {Uncertainty}
  of {Instance} {Segmentation} {Networks}.
\newblock In \emph{Pattern {Recognition}. {ICPR} {International} {Workshops}
  and {Challenges}}, pages 361--374, Cham, 2021. Springer International
  Publishing.

\bibitem[Heidecker et~al.(2023)Heidecker, El-Khateeb, and
  Sick]{heidecker_sampling-based_2023}
Florian Heidecker, Ahmad El-Khateeb, and Bernhard Sick.
\newblock Sampling-based {Uncertainty} {Estimation} for an {Instance}
  {Segmentation} {Network}.
\newblock \emph{arXiv preprint arXiv:2305.14977}, 2023.

\bibitem[Henderson et~al.(2018)Henderson, Islam, Bachman, Pineau, Precup, and
  Meger]{henderson_deep_2018}
Peter Henderson, Riashat Islam, Philip Bachman, Joelle Pineau, Doina Precup,
  and David Meger.
\newblock Deep {Reinforcement} {Learning} {That} {Matters}.
\newblock \emph{Proceedings of the AAAI Conference on Artificial Intelligence},
  32\penalty0 (1), 2018.
\newblock Section: AAAI Technical Track: Machine Learning.

\bibitem[Hofmann et~al.(2017)Hofmann, Wickham, and
  Kafadar]{hofmann_letter-value_2017}
Heike Hofmann, Hadley Wickham, and Karen Kafadar.
\newblock Letter-{Value} {Plots}: {Boxplots} for {Large} {Data}.
\newblock \emph{Journal of Computational and Graphical Statistics}, 26\penalty0
  (3):\penalty0 469--477, 2017.
\newblock Publisher: ASA Website \_eprint:
  https://doi.org/10.1080/10618600.2017.1305277.

\bibitem[Huang et~al.(2018)Huang, Hsu, Chiu, Wu, and Sun]{huang_efficient_2018}
Po-Yu Huang, Wan-Ting Hsu, Chun-Yueh Chiu, Ting-Fan Wu, and Min Sun.
\newblock Efficient {Uncertainty} {Estimation} for {Semantic} {Segmentation} in
  {Videos}.
\newblock In \emph{Proceedings of the {European} {Conference} on {Computer}
  {Vision} ({ECCV})}, 2018.

\bibitem[Jaeger et~al.(2023)Jaeger, Lüth, Klein, and
  Bungert]{jaeger_call_2023}
Paul~F. Jaeger, Carsten~Tim Lüth, Lukas Klein, and Till~J. Bungert.
\newblock A {Call} to {Reflect} on {Evaluation} {Practices} for {Failure}
  {Detection} in {Image} {Classification}.
\newblock In \emph{The {Eleventh} {International} {Conference} on {Learning}
  {Representations}}, 2023.

\bibitem[Kahl et~al.(2024)Kahl, Lüth, Zenk, Maier-Hein, and
  Jaeger]{kahl_values_2024}
Kim-Celine Kahl, Carsten~T. Lüth, Maximilian Zenk, Klaus Maier-Hein, and
  Paul~F. Jaeger.
\newblock {ValUES}: {A} {Framework} for {Systematic} {Validation} of
  {Uncertainty} {Estimation} in {Semantic} {Segmentation}.
\newblock In \emph{The {Twelfth} {International} {Conference} on {Learning}
  {Representations}}, 2024.

\bibitem[Kirillov et~al.(2019)Kirillov, He, Girshick, Rother, and
  Dollar]{kirillov_panoptic_2019}
Alexander Kirillov, Kaiming He, Ross Girshick, Carsten Rother, and Piotr
  Dollar.
\newblock Panoptic {Segmentation}.
\newblock In \emph{Proceedings of the {IEEE}/{CVF} {Conference} on {Computer}
  {Vision} and {Pattern} {Recognition} ({CVPR})}, 2019.

\bibitem[Lakshminarayanan et~al.(2017)Lakshminarayanan, Pritzel, and
  Blundell]{lakshminarayanan_simple_2017}
Balaji Lakshminarayanan, Alexander Pritzel, and Charles Blundell.
\newblock Simple and {Scalable} {Predictive} {Uncertainty} {Estimation} using
  {Deep} {Ensembles}.
\newblock In \emph{Advances in {Neural} {Information} {Processing} {Systems}}.
  Curran Associates, Inc., 2017.

\bibitem[Landgraf et~al.(2024)Landgraf, Hillemann, Wursthorn, and
  Ulrich]{landgraf_uncertainty-aware_2024}
S. Landgraf, M. Hillemann, K. Wursthorn, and M. Ulrich.
\newblock Uncertainty-aware {Cross}-{Entropy} for {Semantic} {Segmentation}.
\newblock \emph{ISPRS Annals of the Photogrammetry, Remote Sensing and Spatial
  Information Sciences}, X-2-2024:\penalty0 129--136, 2024.

\bibitem[Laurent et~al.(2023)Laurent, Lafage, Tartaglione, Daniel, Martinez,
  Bursuc, and Franchi]{laurent_packed_2023}
Olivier Laurent, Adrien Lafage, Enzo Tartaglione, Geoffrey Daniel, Jean-marc
  Martinez, Andrei Bursuc, and Gianni Franchi.
\newblock Packed {Ensembles} for efficient uncertainty estimation.
\newblock In \emph{The {Eleventh} {International} {Conference} on {Learning}
  {Representations}}, 2023.

\bibitem[Li et~al.(2023)Li, Nan, Del~Ser, and Yang]{li_region-based_2023}
Hao Li, Yang Nan, Javier Del~Ser, and Guang Yang.
\newblock Region-based evidential deep learning to quantify uncertainty and
  improve robustness of brain tumor segmentation.
\newblock \emph{Neural Computing and Applications}, 35\penalty0 (30):\penalty0
  22071--22085, 2023.

\bibitem[Li et~al.(2025)Li, Chen, and Yue]{li_vicinal_2025}
Linye Li, Yufei Chen, and Xiaodong Yue.
\newblock Vicinal {Label} {Supervision} for {Reliable} {Aleatoric} and
  {Epistemic} {Uncertainty} {Estimation}.
\newblock In \emph{The {Thirty}-ninth {Annual} {Conference} on {Neural}
  {Information} {Processing} {Systems}}, 2025.

\bibitem[Lin et~al.(2014)Lin, Maire, Belongie, Hays, Perona, Ramanan, Dollár,
  and Zitnick]{lin_microsoft_2014}
Tsung-Yi Lin, Michael Maire, Serge Belongie, James Hays, Pietro Perona, Deva
  Ramanan, Piotr Dollár, and C.~Lawrence Zitnick.
\newblock Microsoft {COCO}: {Common} {Objects} in {Context}.
\newblock In \emph{Computer {Vision} – {ECCV} 2014}, pages 740--755, Cham,
  2014. Springer International Publishing.

\bibitem[Liu et~al.(2025{\natexlab{a}})Liu, Cui, Wang, Chen, Zhang, Zhu, and
  Hu]{liu_improving_2025}
Han Liu, Peng Cui, Bingning Wang, Weipeng Chen, Yupeng Zhang, Jun Zhu, and
  Xiaolin Hu.
\newblock Improving {Accuracy} and {Calibration} via {Differentiated} {Deep}
  {Mutual} {Learning}.
\newblock In \emph{Proceedings of the {IEEE}/{CVF} {Conference} on {Computer}
  {Vision} and {Pattern} {Recognition} ({CVPR})}, pages 25812--25821,
  2025{\natexlab{a}}.

\bibitem[Liu et~al.(2025{\natexlab{b}})Liu, Chen, Da, Chen, Lin, and
  Wei]{liu_uncertainty_2025}
Xiaoou Liu, Tiejin Chen, Longchao Da, Chacha Chen, Zhen Lin, and Hua Wei.
\newblock Uncertainty {Quantification} and {Confidence} {Calibration} in
  {Large} {Language} {Models}: {A} {Survey}.
\newblock In \emph{Proceedings of the 31st {ACM} {SIGKDD} {Conference} on
  {Knowledge} {Discovery} and {Data} {Mining} {V}.2}, pages 6107--6117, New
  York, NY, USA, 2025{\natexlab{b}}. Association for Computing Machinery.
\newblock event-place: Toronto ON, Canada.

\bibitem[Liu et~al.(2021)Liu, Lin, Cao, Hu, Wei, Zhang, Lin, and
  Guo]{liu_swin_2021}
Ze Liu, Yutong Lin, Yue Cao, Han Hu, Yixuan Wei, Zheng Zhang, Stephen Lin, and
  Baining Guo.
\newblock Swin {Transformer}: {Hierarchical} {Vision} {Transformer} {Using}
  {Shifted} {Windows}.
\newblock In \emph{Proceedings of the {IEEE}/{CVF} {International} {Conference}
  on {Computer} {Vision} ({ICCV})}, pages 10012--10022, 2021.

\bibitem[Maag(2021)]{maag_false_2021}
Kira Maag.
\newblock False {Negative} {Reduction} in {Video} {Instance} {Segmentation}
  using {Uncertainty} {Estimates}.
\newblock In \emph{2021 {IEEE} 33rd {International} {Conference} on {Tools}
  with {Artificial} {Intelligence} ({ICTAI})}, pages 1279--1286, 2021.

\bibitem[Maag et~al.(2021)Maag, Rottmann, Varghese, Hüger, Schlicht, and
  Gottschalk]{maag_improving_2021}
Kira Maag, Matthias Rottmann, Serin Varghese, Fabian Hüger, Peter Schlicht,
  and Hanno Gottschalk.
\newblock Improving {Video} {Instance} {Segmentation} by {Light}-weight
  {Temporal} {Uncertainty} {Estimates}.
\newblock In \emph{2021 {International} {Joint} {Conference} on {Neural}
  {Networks} ({IJCNN})}, pages 1--8, 2021.
\newblock ISSN: 2161-4407.

\bibitem[Maier-Hein et~al.(2022)Maier-Hein, Reinke, Godau, Tizabi, Buettner,
  Christodoulou, Glocker, Isensee, Kleesiek, Kozubek, and
  {others}]{maier-hein_metrics_2022-arxiv}
Lena Maier-Hein, Annika Reinke, Patrick Godau, Minu~D Tizabi, Florian Buettner,
  Evangelia Christodoulou, Ben Glocker, Fabian Isensee, Jens Kleesiek, Michal
  Kozubek, and {others}.
\newblock Metrics reloaded: {Recommendations} for image analysis validation.
\newblock \emph{arXiv preprint arXiv:2206.01653}, 2022.

\bibitem[Maier-Hein et~al.(2024)Maier-Hein, Reinke, Godau, Tizabi, Buettner,
  Christodoulou, Glocker, Isensee, Kleesiek, Kozubek, Reyes, Riegler,
  Wiesenfarth, Kavur, Sudre, Baumgartner, Eisenmann, Heckmann-Nötzel, Rädsch,
  Acion, Antonelli, Arbel, Bakas, Benis, Blaschko, Cardoso, Cheplygina, Cimini,
  Collins, Farahani, Ferrer, Galdran, van Ginneken, Haase, Hashimoto, Hoffman,
  Huisman, Jannin, Kahn, Kainmueller, Kainz, Karargyris, Karthikesalingam,
  Kofler, Kopp-Schneider, Kreshuk, Kurc, Landman, Litjens, Madani, Maier-Hein,
  Martel, Mattson, Meijering, Menze, Moons, Müller, Nichyporuk, Nickel,
  Petersen, Rajpoot, Rieke, Saez-Rodriguez, Sánchez, Shetty, van Smeden,
  Summers, Taha, Tiulpin, Tsaftaris, Van~Calster, Varoquaux, and
  Jäger]{maier-hein_metrics_2024}
Lena Maier-Hein, Annika Reinke, Patrick Godau, Minu~D. Tizabi, Florian
  Buettner, Evangelia Christodoulou, Ben Glocker, Fabian Isensee, Jens
  Kleesiek, Michal Kozubek, Mauricio Reyes, Michael~A. Riegler, Manuel
  Wiesenfarth, A.~Emre Kavur, Carole~H. Sudre, Michael Baumgartner, Matthias
  Eisenmann, Doreen Heckmann-Nötzel, Tim Rädsch, Laura Acion, Michela
  Antonelli, Tal Arbel, Spyridon Bakas, Arriel Benis, Matthew~B. Blaschko,
  M.~Jorge Cardoso, Veronika Cheplygina, Beth~A. Cimini, Gary~S. Collins,
  Keyvan Farahani, Luciana Ferrer, Adrian Galdran, Bram van Ginneken, Robert
  Haase, Daniel~A. Hashimoto, Michael~M. Hoffman, Merel Huisman, Pierre Jannin,
  Charles~E. Kahn, Dagmar Kainmueller, Bernhard Kainz, Alexandros Karargyris,
  Alan Karthikesalingam, Florian Kofler, Annette Kopp-Schneider, Anna Kreshuk,
  Tahsin Kurc, Bennett~A. Landman, Geert Litjens, Amin Madani, Klaus
  Maier-Hein, Anne~L. Martel, Peter Mattson, Erik Meijering, Bjoern Menze,
  Karel G.~M. Moons, Henning Müller, Brennan Nichyporuk, Felix Nickel, Jens
  Petersen, Nasir Rajpoot, Nicola Rieke, Julio Saez-Rodriguez, Clara~I.
  Sánchez, Shravya Shetty, Maarten van Smeden, Ronald~M. Summers, Abdel~A.
  Taha, Aleksei Tiulpin, Sotirios~A. Tsaftaris, Ben Van~Calster, Gaël
  Varoquaux, and Paul~F. Jäger.
\newblock Metrics reloaded: recommendations for image analysis validation.
\newblock \emph{Nature Methods}, 21\penalty0 (2):\penalty0 195--212, 2024.

\bibitem[Malinin(2019)]{malinin_uncertainty_2019}
Andrey Malinin.
\newblock \emph{Uncertainty estimation in deep learning with application to
  spoken language assessment}.
\newblock {PhD} {Thesis}, University of Cambridge, 2019.

\bibitem[Miller et~al.(2019)Miller, Dayoub, Milford, and
  Sünderhauf]{miller_evaluating_2019}
Dimity Miller, Feras Dayoub, Michael Milford, and Niko Sünderhauf.
\newblock Evaluating {Merging} {Strategies} for {Sampling}-based {Uncertainty}
  {Techniques} in {Object} {Detection}.
\newblock In \emph{2019 {International} {Conference} on {Robotics} and
  {Automation} ({ICRA})}, pages 2348--2354, 2019.

\bibitem[Minderer et~al.(2021)Minderer, Djolonga, Romijnders, Hubis, Zhai,
  Houlsby, Tran, and Lucic]{minderer_revisiting_2021}
Matthias Minderer, Josip Djolonga, Rob Romijnders, Frances~Ann Hubis, Xiaohua
  Zhai, Neil Houlsby, Dustin Tran, and Mario Lucic.
\newblock Revisiting the {Calibration} of {Modern} {Neural} {Networks}.
\newblock In \emph{Advances in {Neural} {Information} {Processing} {Systems}},
  2021.

\bibitem[Mohan et~al.(2024)Mohan, Kumaraswamy, Hurtado, Petek, and
  Valada]{mohan_panoptic_2024}
Rohit Mohan, Kiran Kumaraswamy, Juana~Valeria Hurtado, Kürsat Petek, and
  Abhinav Valada.
\newblock Panoptic {Out}-of-{Distribution} {Segmentation}.
\newblock \emph{IEEE Robotics and Automation Letters}, 9\penalty0 (5):\penalty0
  4075--4082, 2024.
\newblock Conference Name: IEEE Robotics and Automation Letters.

\bibitem[Morrison et~al.(2019)Morrison, Milan, and
  Antonakos]{morrison_estimating_2019}
Doug Morrison, Anton Milan, and Nontas Antonakos.
\newblock Estimating uncertainty in instance segmentation using dropout
  sampling.
\newblock In \emph{{CVPR} 2019 {Robotic} {Vision} {Probabilistic} {Object}
  {Detection} {Challenge}}, 2019.

\bibitem[Mukhoti et~al.(2023)Mukhoti, Kirsch, van Amersfoort, Torr, and
  Gal]{mukhoti_deep_2023}
Jishnu Mukhoti, Andreas Kirsch, Joost van Amersfoort, Philip~H.S. Torr, and
  Yarin Gal.
\newblock Deep {Deterministic} {Uncertainty}: {A} {New} {Simple} {Baseline}.
\newblock In \emph{Proceedings of the {IEEE}/{CVF} {Conference} on {Computer}
  {Vision} and {Pattern} {Recognition} ({CVPR})}, pages 24384--24394, 2023.

\bibitem[Nie et~al.(2025)Nie, Zhang, Liu, Cheung, Han, and
  Tian]{nie_epistemic_2025}
Jun Nie, Yonggang Zhang, Tongliang Liu, Yiu-ming Cheung, Bo Han, and Xinmei
  Tian.
\newblock Epistemic {Uncertainty} for {Generated} {Image} {Detection}.
\newblock In \emph{The {Thirty}-ninth {Annual} {Conference} on {Neural}
  {Information} {Processing} {Systems}}, 2025.

\bibitem[Oquab et~al.(2024)Oquab, Darcet, Moutakanni, Vo, Szafraniec, Khalidov,
  Fernandez, HAZIZA, Massa, El-Nouby, Assran, Ballas, Galuba, Howes, Huang, Li,
  Misra, Rabbat, Sharma, Synnaeve, Xu, Jegou, Mairal, Labatut, Joulin, and
  Bojanowski]{oquab_dinov2_2024}
Maxime Oquab, Timothée Darcet, Théo Moutakanni, Huy~V. Vo, Marc Szafraniec,
  Vasil Khalidov, Pierre Fernandez, Daniel HAZIZA, Francisco Massa, Alaaeldin
  El-Nouby, Mido Assran, Nicolas Ballas, Wojciech Galuba, Russell Howes, Po-Yao
  Huang, Shang-Wen Li, Ishan Misra, Michael Rabbat, Vasu Sharma, Gabriel
  Synnaeve, Hu Xu, Herve Jegou, Julien Mairal, Patrick Labatut, Armand Joulin,
  and Piotr Bojanowski.
\newblock {DINOv2}: {Learning} {Robust} {Visual} {Features} without
  {Supervision}.
\newblock \emph{Transactions on Machine Learning Research}, 2024.

\bibitem[Osband et~al.(2023)Osband, Wen, Asghari, Dwaracherla, IBRAHIMI, Lu,
  and Van~Roy]{osband_epistemic_2023}
Ian Osband, Zheng Wen, Seyed~Mohammad Asghari, Vikranth Dwaracherla, MORTEZA
  IBRAHIMI, Xiuyuan Lu, and Benjamin Van~Roy.
\newblock Epistemic {Neural} {Networks}.
\newblock In \emph{Advances in {Neural} {Information} {Processing} {Systems}},
  pages 2795--2823. Curran Associates, Inc., 2023.

\bibitem[Pakdaman~Naeini et~al.(2015)Pakdaman~Naeini, Cooper, and
  Hauskrecht]{pakdaman_naeini_obtaining_2015}
Mahdi Pakdaman~Naeini, Gregory Cooper, and Milos Hauskrecht.
\newblock Obtaining {Well} {Calibrated} {Probabilities} {Using} {Bayesian}
  {Binning}.
\newblock \emph{Proceedings of the AAAI Conference on Artificial Intelligence},
  29\penalty0 (1), 2015.
\newblock Section: Main Track: Novel Machine Learning Algorithms.

\bibitem[Pan et~al.(2025)Pan, Sun, Li, and Zhang]{pan_exploring_2025}
Yuwen Pan, Rui Sun, Wangkai Li, and Tianzhu Zhang.
\newblock Exploring {Weather}-aware {Aggregation} and {Adaptation} for
  {Semantic} {Segmentation} under {Adverse} {Conditions}.
\newblock In \emph{Proceedings of the {IEEE}/{CVF} {International} {Conference}
  on {Computer} {Vision} ({ICCV})}, pages 13952--13962, 2025.

\bibitem[Papamarkou et~al.(2024)Papamarkou, Skoularidou, Palla, Aitchison,
  Arbel, Dunson, Filippone, Fortuin, Hennig, Hernández-Lobato, Hubin, Immer,
  Karaletsos, Khan, Kristiadi, Li, Mandt, Nemeth, Osborne, Rudner, Rügamer,
  Teh, Welling, Wilson, and Zhang]{papamarkou_position_2024}
Theodore Papamarkou, Maria Skoularidou, Konstantina Palla, Laurence Aitchison,
  Julyan Arbel, David Dunson, Maurizio Filippone, Vincent Fortuin, Philipp
  Hennig, José~Miguel Hernández-Lobato, Aliaksandr Hubin, Alexander Immer,
  Theofanis Karaletsos, Mohammad~Emtiyaz Khan, Agustinus Kristiadi, Yingzhen
  Li, Stephan Mandt, Christopher Nemeth, Michael~A Osborne, Tim G.~J. Rudner,
  David Rügamer, Yee~Whye Teh, Max Welling, Andrew~Gordon Wilson, and Ruqi
  Zhang.
\newblock Position: {Bayesian} {Deep} {Learning} is {Needed} in the {Age} of
  {Large}-{Scale} {AI}.
\newblock In \emph{Proceedings of the 41st {International} {Conference} on
  {Machine} {Learning}}, pages 39556--39586. PMLR, 2024.

\bibitem[Ramé et~al.(2023)Ramé, Ahuja, Zhang, Cord, Bottou, and
  Lopez-Paz]{rame_model_2023}
Alexandre Ramé, Kartik Ahuja, Jianyu Zhang, Matthieu Cord, Léon Bottou, and
  David Lopez-Paz.
\newblock Model ratatouille: recycling diverse models for out-of-distribution
  generalization.
\newblock In \emph{Proceedings of the 40th {International} {Conference} on
  {Machine} {Learning}}, Honolulu, Hawaii, USA, 2023. JMLR.org.

\bibitem[Richter et~al.(2017)Richter, Hayder, and Koltun]{richter_playing_2017}
Stephan~R. Richter, Zeeshan Hayder, and Vladlen Koltun.
\newblock Playing for {Benchmarks}.
\newblock In \emph{{IEEE} {International} {Conference} on {Computer} {Vision},
  {ICCV} 2017, {Venice}, {Italy}, {October} 22-29, 2017}, pages 2232--2241,
  2017.

\bibitem[Roschewitz et~al.(2025)Roschewitz, Mehta, Ribeiro, and
  Glocker]{roschewitz_where_2025}
Mélanie Roschewitz, Raghav Mehta, Fabio De~Sousa Ribeiro, and Ben Glocker.
\newblock Where are we with calibration under dataset shift in image
  classification?
\newblock \emph{Transactions on Machine Learning Research}, 2025.

\bibitem[Saha et~al.(2023)Saha, Hoyer, Obukhov, Dai, and
  Van~Gool]{saha_edaps_2023}
Suman Saha, Lukas Hoyer, Anton Obukhov, Dengxin Dai, and Luc Van~Gool.
\newblock {EDAPS}: {Enhanced} {Domain}-{Adaptive} {Panoptic} {Segmentation}.
\newblock In \emph{Proceedings of the {IEEE}/{CVF} {International} {Conference}
  on {Computer} {Vision} ({ICCV})}, pages 19234--19245, 2023.

\bibitem[Schmidt et~al.(2025)Schmidt, Koerner, Fuchsgruber, Gasperini, Tombari,
  and Günnemann]{schmidt_prior2former_2025}
Sebastian Schmidt, Julius Koerner, Dominik Fuchsgruber, Stefano Gasperini,
  Federico Tombari, and Stephan Günnemann.
\newblock {Prior2Former} - {Evidential} {Modeling} of {Mask} {Transformers} for
  {Assumption}-{Free} {Open}-{World} {Panoptic} {Segmentation}.
\newblock In \emph{Proceedings of the {IEEE}/{CVF} {International} {Conference}
  on {Computer} {Vision} ({ICCV})}, pages 23646--23656, 2025.

\bibitem[Schrüfer et~al.(2024)Schrüfer, Milling, Burkhardt, Eyben, and
  Schuller]{schrufer_are_2024}
Oliver Schrüfer, Manuel Milling, Felix Burkhardt, Florian Eyben, and Björn
  Schuller.
\newblock Are you sure? {Analysing} {Uncertainty} {Quantification} {Approaches}
  for {Real}-world {Speech} {Emotion} {Recognition}.
\newblock In \emph{Interspeech 2024}, pages 3210--3214, 2024.
\newblock ISSN: 2958-1796.

\bibitem[Sensoy et~al.(2018)Sensoy, Kaplan, and
  Kandemir]{sensoy_evidential_2018}
Murat Sensoy, Lance Kaplan, and Melih Kandemir.
\newblock Evidential {Deep} {Learning} to {Quantify} {Classification}
  {Uncertainty}.
\newblock In \emph{Advances in {Neural} {Information} {Processing} {Systems}}.
  Curran Associates, Inc., 2018.

\bibitem[Shannon(1948)]{shannon_mathematical_1948}
C.~E. Shannon.
\newblock A mathematical theory of communication.
\newblock \emph{The Bell System Technical Journal}, 27\penalty0 (3):\penalty0
  379--423, 1948.
\newblock Conference Name: The Bell System Technical Journal.

\bibitem[Shelhamer et~al.(2017)Shelhamer, Long, and
  Darrell]{shelhamer_fully_2017}
Evan Shelhamer, Jonathan Long, and Trevor Darrell.
\newblock Fully {Convolutional} {Networks} for {Semantic} {Segmentation}.
\newblock \emph{IEEE Transactions on Pattern Analysis and Machine
  Intelligence}, 39\penalty0 (4):\penalty0 640--651, 2017.

\bibitem[Sirohi et~al.(2023)Sirohi, Marvi, Büscher, and
  Burgard]{sirohi_uncertainty-aware_2023}
Kshitij Sirohi, Sajad Marvi, Daniel Büscher, and Wolfram Burgard.
\newblock Uncertainty-{Aware} {Panoptic} {Segmentation}.
\newblock \emph{IEEE Robotics and Automation Letters}, 8\penalty0 (5):\penalty0
  2629--2636, 2023.

\bibitem[Smith et~al.(2025)Smith, Kossen, Trollope, Wilk, Foster, and
  Rainforth]{smith_rethinking_2025}
Freddie~Bickford Smith, Jannik Kossen, Eleanor Trollope, Mark van~der Wilk,
  Adam Foster, and Tom Rainforth.
\newblock Rethinking {Aleatoric} and {Epistemic} {Uncertainty}.
\newblock In \emph{Forty-second {International} {Conference} on {Machine}
  {Learning}}, 2025.

\bibitem[Smith and Ferrie(2023)]{smith_uncertainty_2024}
Michael Smith and Frank Ferrie.
\newblock Uncertainty estimation in deep learning for panoptic segmentation.
\newblock \emph{arXiv preprint arXiv:2304.02098}, 2023.

\bibitem[Teed and Deng(2020)]{teed_raft_2020}
Zachary Teed and Jia Deng.
\newblock {RAFT}: {Recurrent} {All}-{Pairs} {Field} {Transforms} for {Optical}
  {Flow}.
\newblock In \emph{Computer {Vision} – {ECCV} 2020}, pages 402--419, Cham,
  2020. Springer International Publishing.

\bibitem[Thisanke et~al.(2023)Thisanke, Deshan, Chamith, Seneviratne,
  Vidanaarachchi, and Herath]{thisanke_semantic_2023}
Hans Thisanke, Chamli Deshan, Kavindu Chamith, Sachith Seneviratne, Rajith
  Vidanaarachchi, and Damayanthi Herath.
\newblock Semantic segmentation using {Vision} {Transformers}: {A} survey.
\newblock \emph{Engineering Applications of Artificial Intelligence},
  126:\penalty0 106669, 2023.

\bibitem[Thoma et~al.(2025)Thoma, Preintner, Aghajanzadeh, Sampath, Mori,
  Fasfous, Vemparala, Frickenstein, Mueller-Gritschneder, and
  Schlichtmann]{thoma_uncertainty_2025}
Moritz Thoma, Tobias Preintner, Emad Aghajanzadeh, Shambhavi~Balamuthu Sampath,
  Pierpaolo Mori, Nael Fasfous, Manoj-Rohit Vemparala, Alexander Frickenstein,
  Daniel Mueller-Gritschneder, and Ulf Schlichtmann.
\newblock Uncertainty {Aware} {Training} to {Improve} {Uncertainty} {Active}
  {Learning} for {Semantic} {Segmentation}.
\newblock In \emph{Proceedings of the {IEEE}/{CVF} {Conference} on {Computer}
  {Vision} and {Pattern} {Recognition} ({CVPR}) {Workshops}}, pages 4410--4420,
  2025.

\bibitem[Urbinati et~al.(2025)Urbinati, Lai, Frenda, and
  Stranisci]{urbinati_are_2025}
Alessandra Urbinati, Mirko Lai, Simona Frenda, and Marco Stranisci.
\newblock Are you sure? {Measuring} models bias in content moderation through
  uncertainty.
\newblock In \emph{Findings of the {Association} for {Computational}
  {Linguistics}: {EMNLP} 2025}, pages 18061--18076, Suzhou, China, 2025.
  Association for Computational Linguistics.

\bibitem[Valiuddin et~al.(2024)Valiuddin, van Sloun, Viviers, de~With, and
  van~der Sommen]{valiuddin_review_2024}
MMA Valiuddin, RJG van Sloun, CGA Viviers, PHN de With, and F van~der Sommen.
\newblock A {Review} of {Bayesian} {Uncertainty} {Quantification} in {Deep}
  {Probabilistic} {Image} {Segmentation}.
\newblock \emph{arXiv preprint arXiv:2411.16370}, 2024.

\bibitem[Vu et~al.(2025)Vu, Valle, Bursuc, Kerssies, de~Geus, Dubbelman, Qian,
  Zhu, Chen, Tang, Wang, Vojíř, Šochman, Matas, Smith, Ferrie, Basu,
  Sakaridis, and Van~Gool]{vu_bravo_2025}
Tuan-Hung Vu, Eduardo Valle, Andrei Bursuc, Tommie Kerssies, Daan de Geus, Gijs
  Dubbelman, Long Qian, Bingke Zhu, Yingying Chen, Ming Tang, Jinqiao Wang,
  Tomáš Vojíř, Jan Šochman, Jiří Matas, Michael Smith, Frank Ferrie,
  Shamik Basu, Christos Sakaridis, and Luc Van~Gool.
\newblock The {BRAVO} {Semantic} {Segmentation} {Challenge} {Results} in
  {UNCV2024}.
\newblock In \emph{Computer {Vision} – {ECCV} 2024 {Workshops}}, pages
  290--306, Cham, 2025. Springer Nature Switzerland.

\bibitem[Wei et~al.(2024)Wei, Chen, Jin, Ma, Liu, Ling, Wang, Chen, and
  Zheng]{wei_stronger_2024}
Zhixiang Wei, Lin Chen, Yi Jin, Xiaoxiao Ma, Tianle Liu, Pengyang Ling, Ben
  Wang, Huaian Chen, and Jinjin Zheng.
\newblock Stronger {Fewer} \& {Superior}: {Harnessing} {Vision} {Foundation}
  {Models} for {Domain} {Generalized} {Semantic} {Segmentation}.
\newblock In \emph{Proceedings of the {IEEE}/{CVF} {Conference} on {Computer}
  {Vision} and {Pattern} {Recognition} ({CVPR})}, pages 28619--28630, 2024.

\bibitem[Wen et~al.(2021)Wen, Jerfel, Muller, Dusenberry, Snoek,
  Lakshminarayanan, and Tran]{wen_combining_2021}
Yeming Wen, Ghassen Jerfel, Rafael Muller, Michael~W. Dusenberry, Jasper Snoek,
  Balaji Lakshminarayanan, and Dustin Tran.
\newblock Combining {Ensembles} and {Data} {Augmentation} {Can} {Harm} {Your}
  {Calibration}.
\newblock In \emph{International {Conference} on {Learning} {Representations}},
  2021.

\bibitem[Wilson et~al.(2022)Wilson, Izmailov, Hoffman, Gal, Li, Pradier,
  Vikram, Foong, Lotfi, and Farquhar]{wilson_evaluating_2022}
Andrew~Gordon Wilson, Pavel Izmailov, Matthew~D. Hoffman, Yarin Gal, Yingzhen
  Li, Melanie~F. Pradier, Sharad Vikram, Andrew Foong, Sanae Lotfi, and
  Sebastian Farquhar.
\newblock Evaluating {Approximate} {Inference} in {Bayesian} {Deep} {Learning}.
\newblock In \emph{Proceedings of the {NeurIPS} 2021 {Competitions} and
  {Demonstrations} {Track}}, pages 113--124. PMLR, 2022.
\newblock ISSN: 2640-3498.

\bibitem[Xu et~al.(2025)Xu, Zhang, Li, Wang, Guan, Yan, Zhang, and
  Song]{xu_dual-temporal_2025}
Xiaolong Xu, Lei Zhang, Jiayi Li, Lituan Wang, Yifan Guan, Yu Yan, Leyi Zhang,
  and Hao Song.
\newblock Dual-{Temporal} {Exemplar} {Representation} {Network} for {Video}
  {Semantic} {Segmentation}.
\newblock In \emph{Proceedings of the {IEEE}/{CVF} {International} {Conference}
  on {Computer} {Vision} ({ICCV})}, pages 10775--10785, 2025.

\bibitem[Zhang et~al.(2018)Zhang, Cisse, Dauphin, and
  Lopez-Paz]{zhang_mixup_2018}
Hongyi Zhang, Moustapha Cisse, Yann~N. Dauphin, and David Lopez-Paz.
\newblock mixup: {Beyond} {Empirical} {Risk} {Minimization}.
\newblock In \emph{International {Conference} on {Learning} {Representations}},
  2018.

\bibitem[Zhou(2025)]{zhou_ensemble_2025}
Zhi-Hua Zhou.
\newblock \emph{Ensemble {Methods}: {Foundations} and {Algorithms}}.
\newblock Chapman and Hall/CRC, New York, 2 edition, 2025.

\end{thebibliography}


\begin{thebibliography}{21}
\providecommand{\natexlab}[1]{#1}
\providecommand{\url}[1]{\texttt{#1}}
\expandafter\ifx\csname urlstyle\endcsname\relax
  \providecommand{\doi}[1]{doi: #1}\else
  \providecommand{\doi}{doi: \begingroup \urlstyle{rm}\Url}\fi

\bibitem[Cheng et~al.(2022)Cheng, Misra, Schwing, Kirillov, and
  Girdhar]{cheng_masked-attention_2022}
Bowen Cheng, Ishan Misra, Alexander~G. Schwing, Alexander Kirillov, and Rohit
  Girdhar.
\newblock Masked-attention {Mask} {Transformer} for {Universal} {Image}
  {Segmentation}.
\newblock In \emph{2022 {IEEE}/{CVF} {Conference} on {Computer} {Vision} and
  {Pattern} {Recognition} ({CVPR})}, pages 1280--1289, 2022.

\bibitem[Cordts et~al.(2016)Cordts, Omran, Ramos, Rehfeld, Enzweiler, Benenson,
  Franke, Roth, and Schiele]{cordts_cityscapes_2016}
Marius Cordts, Mohamed Omran, Sebastian Ramos, Timo Rehfeld, Markus Enzweiler,
  Rodrigo Benenson, Uwe Franke, Stefan Roth, and Bernt Schiele.
\newblock The {Cityscapes} {Dataset} for {Semantic} {Urban} {Scene}
  {Understanding}.
\newblock In \emph{Proc. of the {IEEE} {Conference} on {Computer} {Vision} and
  {Pattern} {Recognition} ({CVPR})}, 2016.

\bibitem[Deery et~al.(2023)Deery, Lee, and Waslander]{deery_propandl_2023}
Jacob Deery, Chang~Won Lee, and Steven~L. Waslander.
\newblock {ProPanDL}: {A} {Modular} {Architecture} for {Uncertainty}-{Aware}
  {Panoptic} {Segmentation}.
\newblock In \emph{2023 20th {Conference} on {Robots} and {Vision} ({CRV})},
  pages 137--144, 2023.

\bibitem[Ghiasi et~al.(2021)Ghiasi, Cui, Srinivas, Qian, Lin, Cubuk, Le, and
  Zoph]{ghiasi_simple_2021}
Golnaz Ghiasi, Yin Cui, Aravind Srinivas, Rui Qian, Tsung-Yi Lin, Ekin~D.
  Cubuk, Quoc~V. Le, and Barret Zoph.
\newblock Simple {Copy}-{Paste} {Is} a {Strong} {Data} {Augmentation} {Method}
  for {Instance} {Segmentation}.
\newblock In \emph{Proceedings of the {IEEE}/{CVF} {Conference} on {Computer}
  {Vision} and {Pattern} {Recognition} ({CVPR})}, pages 2918--2928, 2021.

\bibitem[Guo et~al.(2017)Guo, Pleiss, Sun, and
  Weinberger]{guo_calibration_2017}
Chuan Guo, Geoff Pleiss, Yu Sun, and Kilian~Q. Weinberger.
\newblock On {Calibration} of {Modern} {Neural} {Networks}.
\newblock In \emph{Proceedings of the 34th {International} {Conference} on
  {Machine} {Learning}}, pages 1321--1330. PMLR, 2017.

\bibitem[Huang et~al.(2018)Huang, Hsu, Chiu, Wu, and Sun]{huang_efficient_2018}
Po-Yu Huang, Wan-Ting Hsu, Chun-Yueh Chiu, Ting-Fan Wu, and Min Sun.
\newblock Efficient {Uncertainty} {Estimation} for {Semantic} {Segmentation} in
  {Videos}.
\newblock In \emph{Proceedings of the {European} {Conference} on {Computer}
  {Vision} ({ECCV})}, 2018.

\bibitem[Jaeger et~al.(2023)Jaeger, Lüth, Klein, and
  Bungert]{jaeger_call_2023}
Paul~F. Jaeger, Carsten~Tim Lüth, Lukas Klein, and Till~J. Bungert.
\newblock A {Call} to {Reflect} on {Evaluation} {Practices} for {Failure}
  {Detection} in {Image} {Classification}.
\newblock In \emph{The {Eleventh} {International} {Conference} on {Learning}
  {Representations}}, 2023.

\bibitem[Kahl et~al.(2024)Kahl, Lüth, Zenk, Maier-Hein, and
  Jaeger]{kahl_values_2024}
Kim-Celine Kahl, Carsten~T. Lüth, Maximilian Zenk, Klaus Maier-Hein, and
  Paul~F. Jaeger.
\newblock {ValUES}: {A} {Framework} for {Systematic} {Validation} of
  {Uncertainty} {Estimation} in {Semantic} {Segmentation}.
\newblock In \emph{The {Twelfth} {International} {Conference} on {Learning}
  {Representations}}, 2024.

\bibitem[Kirillov et~al.(2019)Kirillov, He, Girshick, Rother, and
  Dollar]{kirillov_panoptic_2019}
Alexander Kirillov, Kaiming He, Ross Girshick, Carsten Rother, and Piotr
  Dollar.
\newblock Panoptic {Segmentation}.
\newblock In \emph{Proceedings of the {IEEE}/{CVF} {Conference} on {Computer}
  {Vision} and {Pattern} {Recognition} ({CVPR})}, 2019.

\bibitem[Lin et~al.(2014)Lin, Maire, Belongie, Hays, Perona, Ramanan, Dollár,
  and Zitnick]{lin_microsoft_2014}
Tsung-Yi Lin, Michael Maire, Serge Belongie, James Hays, Pietro Perona, Deva
  Ramanan, Piotr Dollár, and C.~Lawrence Zitnick.
\newblock Microsoft {COCO}: {Common} {Objects} in {Context}.
\newblock In \emph{Computer {Vision} – {ECCV} 2014}, pages 740--755, Cham,
  2014. Springer International Publishing.

\bibitem[Minderer et~al.(2021)Minderer, Djolonga, Romijnders, Hubis, Zhai,
  Houlsby, Tran, and Lucic]{minderer_revisiting_2021}
Matthias Minderer, Josip Djolonga, Rob Romijnders, Frances~Ann Hubis, Xiaohua
  Zhai, Neil Houlsby, Dustin Tran, and Mario Lucic.
\newblock Revisiting the {Calibration} of {Modern} {Neural} {Networks}.
\newblock In \emph{Advances in {Neural} {Information} {Processing} {Systems}},
  2021.

\bibitem[Morrison et~al.(2019)Morrison, Milan, and
  Antonakos]{morrison_estimating_2019}
Doug Morrison, Anton Milan, and Nontas Antonakos.
\newblock Estimating uncertainty in instance segmentation using dropout
  sampling.
\newblock In \emph{{CVPR} 2019 {Robotic} {Vision} {Probabilistic} {Object}
  {Detection} {Challenge}}, 2019.

\bibitem[Mukhoti et~al.(2021)Mukhoti, Kirsch, van Amersfoort, Torr, and
  Gal]{mukhoti_deep_2021}
Jishnu Mukhoti, Andreas Kirsch, Joost van Amersfoort, Philip~HS Torr, and Yarin
  Gal.
\newblock Deep deterministic uncertainty: {A} simple baseline.
\newblock \emph{arXiv preprint arXiv:2102.11582}, 2021.

\bibitem[Oquab et~al.(2024)Oquab, Darcet, Moutakanni, Vo, Szafraniec, Khalidov,
  Fernandez, HAZIZA, Massa, El-Nouby, Assran, Ballas, Galuba, Howes, Huang, Li,
  Misra, Rabbat, Sharma, Synnaeve, Xu, Jegou, Mairal, Labatut, Joulin, and
  Bojanowski]{oquab_dinov2_2024}
Maxime Oquab, Timothée Darcet, Théo Moutakanni, Huy~V. Vo, Marc Szafraniec,
  Vasil Khalidov, Pierre Fernandez, Daniel HAZIZA, Francisco Massa, Alaaeldin
  El-Nouby, Mido Assran, Nicolas Ballas, Wojciech Galuba, Russell Howes, Po-Yao
  Huang, Shang-Wen Li, Ishan Misra, Michael Rabbat, Vasu Sharma, Gabriel
  Synnaeve, Hu Xu, Herve Jegou, Julien Mairal, Patrick Labatut, Armand Joulin,
  and Piotr Bojanowski.
\newblock {DINOv2}: {Learning} {Robust} {Visual} {Features} without
  {Supervision}.
\newblock \emph{Transactions on Machine Learning Research}, 2024.

\bibitem[Pedregosa et~al.(2011)Pedregosa, Varoquaux, Gramfort, Michel, Thirion,
  Grisel, Blondel, Prettenhofer, Weiss, Dubourg, Vanderplas, Passos,
  Cournapeau, Brucher, Perrot, and Duchesnay]{pedregosa_scikit-learn_2011}
F. Pedregosa, G. Varoquaux, A. Gramfort, V. Michel, B. Thirion, O. Grisel, M.
  Blondel, P. Prettenhofer, R. Weiss, V. Dubourg, J. Vanderplas, A. Passos, D.
  Cournapeau, M. Brucher, M. Perrot, and E. Duchesnay.
\newblock Scikit-learn: {Machine} {Learning} in {Python}.
\newblock \emph{Journal of Machine Learning Research}, 12:\penalty0 2825--2830,
  2011.

\bibitem[Richter et~al.(2017)Richter, Hayder, and Koltun]{richter_playing_2017}
Stephan~R. Richter, Zeeshan Hayder, and Vladlen Koltun.
\newblock Playing for {Benchmarks}.
\newblock In \emph{{IEEE} {International} {Conference} on {Computer} {Vision},
  {ICCV} 2017, {Venice}, {Italy}, {October} 22-29, 2017}, pages 2232--2241,
  2017.

\bibitem[Sirohi et~al.(2023)Sirohi, Marvi, Büscher, and
  Burgard]{sirohi_uncertainty-aware_2023}
Kshitij Sirohi, Sajad Marvi, Daniel Büscher, and Wolfram Burgard.
\newblock Uncertainty-{Aware} {Panoptic} {Segmentation}.
\newblock \emph{IEEE Robotics and Automation Letters}, 8\penalty0 (5):\penalty0
  2629--2636, 2023.

\bibitem[Smith and Ferrie(2023)]{smith_uncertainty_2024}
Michael Smith and Frank Ferrie.
\newblock Uncertainty estimation in deep learning for panoptic segmentation.
\newblock \emph{arXiv preprint arXiv:2304.02098}, 2023.

\bibitem[Teed and Deng(2020)]{teed_raft_2020}
Zachary Teed and Jia Deng.
\newblock {RAFT}: {Recurrent} {All}-{Pairs} {Field} {Transforms} for {Optical}
  {Flow}.
\newblock In \emph{Computer {Vision} – {ECCV} 2020}, pages 402--419, Cham,
  2020. Springer International Publishing.

\bibitem[Vu et~al.(2025)Vu, Valle, Bursuc, Kerssies, de~Geus, Dubbelman, Qian,
  Zhu, Chen, Tang, Wang, Vojíř, Šochman, Matas, Smith, Ferrie, Basu,
  Sakaridis, and Van~Gool]{vu_bravo_2025}
Tuan-Hung Vu, Eduardo Valle, Andrei Bursuc, Tommie Kerssies, Daan de Geus, Gijs
  Dubbelman, Long Qian, Bingke Zhu, Yingying Chen, Ming Tang, Jinqiao Wang,
  Tomáš Vojíř, Jan Šochman, Jiří Matas, Michael Smith, Frank Ferrie,
  Shamik Basu, Christos Sakaridis, and Luc Van~Gool.
\newblock The {BRAVO} {Semantic} {Segmentation} {Challenge} {Results} in
  {UNCV2024}.
\newblock In \emph{Computer {Vision} – {ECCV} 2024 {Workshops}}, pages
  290--306, Cham, 2025. Springer Nature Switzerland.

\bibitem[Zhu et~al.(2021)Zhu, Su, Lu, Li, Wang, and Dai]{zhu_deformable_2021}
Xizhou Zhu, Weijie Su, Lewei Lu, Bin Li, Xiaogang Wang, and Jifeng Dai.
\newblock Deformable {DETR}: {Deformable} {Transformers} for {End}-to-{End}
  {Object} {Detection}.
\newblock In \emph{International {Conference} on {Learning} {Representations}},
  2021.

\end{thebibliography}
}

% To split the supplementary pages from the main paper, you can use \href{https://support.apple.com/en-ca/guide/preview/prvw11793/mac#:~:text=Delete%20a%20page%20from%20a,or%20choose%20Edit%20%3E%20Delete).}{Preview (on macOS)}, \href{https://www.adobe.com/acrobat/how-to/delete-pages-from-pdf.html#:~:text=Choose%20%E2%80%9CTools%E2%80%9D%20%3E%20%E2%80%9COrganize,or%20pages%20from%20the%20file.}{Adobe Acrobat} (on all OSs), as well as \href{https://superuser.com/questions/517986/is-it-possible-to-delete-some-pages-of-a-pdf-document}{command line tools}.

\end{document}